\documentclass[10pt,twocolumn,letterpaper]{article}

\usepackage[pagenumbers]{cvpr} % To force page numbers, e.g. for an arXiv version
\usepackage{graphicx} % Required for inserting images
\usepackage{multirow}   % For multi-row cells in tables
\usepackage[table]{xcolor} % For coloring table cells/rows
\usepackage{siunitx} % For aligning numbers in tables
\usepackage[percent]{overpic} % For overlaying image 
\usepackage{pifont} 
\newcommand{\cmark}{\ding{51}} % ✓
\newcommand{\xmark}{\ding{55}} % ✗
\usepackage{tabularx}
\usepackage{booktabs}
\usepackage{epigraph}
\usepackage{makecell}
\usepackage{listings}
\usepackage{longtable}
\usepackage{catchfile}
\usepackage{caption}
\usepackage{capt-of}
\usepackage{subcaption}
\usepackage[most]{tcolorbox}
\usepackage{xcolor}

\usepackage{hhline}
\usepackage[most]{tcolorbox}

\newcolumntype{Y}{>{\centering\arraybackslash}X}

\definecolor{color_1}{RGB}{255,0,128}
\definecolor{color_2}{RGB}{0,128,128}
\definecolor{color_3}{RGB}{0,128,0}
\definecolor{color_4}{RGB}{128,0,0}
\definecolor{color_5}{RGB}{128,0,128}
\definecolor{cadetgrey}{RGB}{0.57, 0.64, 0.69}

\definecolor{Gray}{gray}{0.9}

\newcommand{\correctmark}{{\color{correctgreen}\fontsize{8}{8}\selectfont\bfseries\cmark} }
\newcommand{\wrongmark}{{\color{wrongred}\fontsize{8}{8}\selectfont\bfseries\xmark} }

\newcommand{\Benchmark}{{AVS}}
\newcommand{\Method}{AVS}
\newcommand{\Qry}{\text{qry}}
\newcommand{\Tgt}{\text{tgt}}

\definecolor{correctgreen}{RGB}{26,187,46}
\definecolor{wrongred}{RGB}{243,1,3}

\newcommand{\receptacle}[1]{%
  \begingroup
  \setlength{\fboxsep}{2pt}
  \colorbox{olive!15}{#1}%
  \endgroup
}
\newcommand{\targetobj}[1]{%
  \begingroup
  \setlength{\fboxsep}{2pt}
  \colorbox{blue!15}{#1}%
  \endgroup
}

\newcolumntype{Z}{>{\centering\arraybackslash}m{0.22\textwidth}}
\newcolumntype{L}{>{\centering\arraybackslash}m{0.44\textwidth}}
\newcolumntype{R}{>{\centering\arraybackslash}m{0.44\textwidth}}

\newcommand{\Image}[1]{
    \multicolumn{\LT@cols}{l}{\includegraphics[width=\textwidth]{#1}}\\
    \midrule
}

\newcommand{\ImagePair}[2]{%
  \multicolumn{2}{L|}{\includegraphics[width=\linewidth]{#1}} &
  \multicolumn{2}{R}{\includegraphics[width=\linewidth]{#2}} \\
  \midrule
}

\newcommand{\ImagePairLast}[2]{%
  \multicolumn{2}{L|}{\includegraphics[width=\linewidth]{#1}} &
  \multicolumn{2}{R}{\includegraphics[width=\linewidth]{#2}} \\
}

\definecolor{cvprblue}{rgb}{0.21,0.49,0.74}
\usepackage[pagebackref,breaklinks,colorlinks,allcolors=cvprblue]{hyperref}

\newif\ifsuppl
\supplfalse
\newif\ifarxiv
\arxivtrue

\ifarxiv
    \newcommand{\Suppl}{{Appendix}}
\else
    \newcommand{\Suppl}{\textbf{supplementary maeterial}}
\fi

\def\paperID{41266} % *** Enter the Paper ID here
\def\confName{CVPR}
\def\confYear{2026}

\title{Toward Ambulatory Vision: Learning Visually-Grounded Active View Selection}

\author{
Juil Koo\textsuperscript{$\ast$} $\quad$
Daehyeon Choi\textsuperscript{$\ast$} $\quad$
Sangwoo Youn\textsuperscript{$\ast$}$\quad$
Phillip Y. Lee $\quad$
Minhyuk Sung \\
KAIST\\
{\tt\small \{63days,daehyeonchoi,andy2884,phillip0701,mhsung\}@kaist.ac.kr}\\
}

\begin{document}
% \maketitle
\twocolumn[{%
\renewcommand\twocolumn[1][]{#1}%
\maketitle
\ifarxiv
    \begin{center}
    \vspace{-1.5\baselineskip}
    Project Page: \url{https://active-view-selection.github.io}
    \end{center}
\else
\fi
\begin{center}
\centering
\captionsetup{type=figure}
% \dummyfig{1\linewidth}{0.5\linewidth}{Teaser.}
\includegraphics[width=\textwidth]{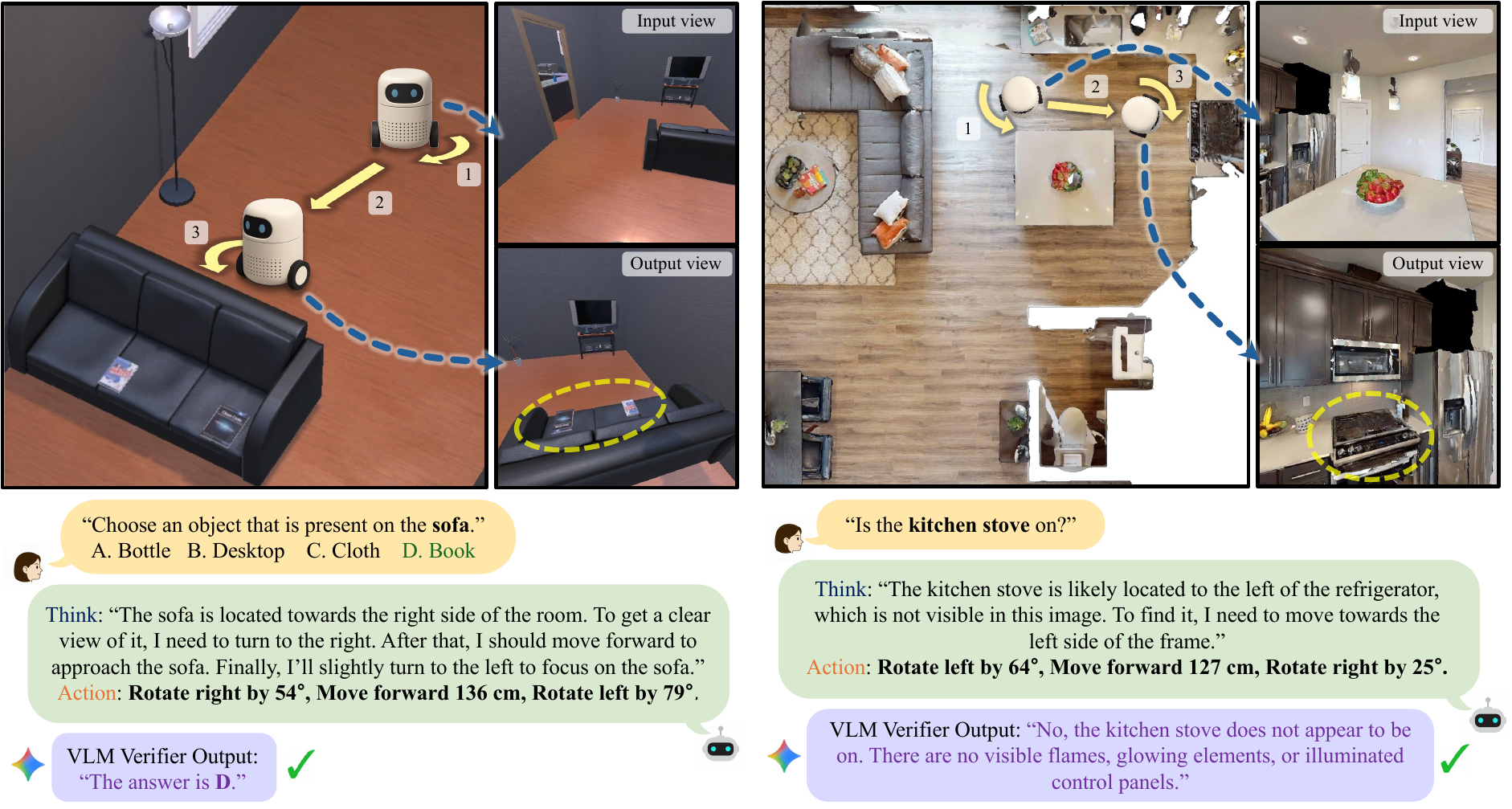}
\vspace{-\baselineskip}
\caption{\textbf{An overview of our proposed Visually-Grounded Active View Selection (VG-AVS).} Given a 3D environment from synthetic (left) to real scenes (right) and a question, our learning-based active view selection (AVS) framework predicts continuous actions to refine the agent's viewpoint. The refined view is then fed into a VLM verifier, which answers the question based on the improved observation.}
\label{fig:teaser}
\end{center}
}]
\def\thefootnote{*}\footnotetext{Equal contribution.}\def\thefootnote{\arabic{footnote}}

\begin{abstract}
Vision Language Models (VLMs) excel at visual question answering (VQA) but remain limited to snapshot vision, reasoning from static images. In contrast, embodied agents require ambulatory vision, actively moving to obtain more informative views. We introduce Visually Grounded Active View Selection (VG-AVS), a task that selects the most informative next viewpoint using only the visual information in the current image, without relying on scene memory or external knowledge. To support this task, we construct a synthetic dataset with automatically generated paired query–target views and question–answer prompts. We also propose a framework that fine-tunes pretrained VLMs through supervised fine-tuning (SFT) followed by RL-based policy optimization. Our approach achieves strong question answering performance based on viewpoint selection and generalizes robustly to unseen synthetic and real scenes. Furthermore, incorporating our learned VG-AVS framework into existing scene-exploration-based EQA systems improves downstream question-answering accuracy.
\end{abstract}

% Prior Embodied Question Answering (EQA) frameworks include active perception only as one component within a complex pipeline involving exploration, scene memorization, and commonsense reasoning, which obscures its core role. 

% Recent Visual Language Models (VLMs) have shown strong capabilities in joint text and image understanding. However, when applied to embodied agents that must perceive and reason about 3D environments rather than static images, a key limitation emerges: the inability to identify and capture missing visual information. We propose a framework that learns camera pose control for visual question answering (VQA), enabling agents to predict camera motions that acquire views necessary to answer questions. Unlike prior scene understanding approaches that rely on semantic priors or pre-built scene representations, our method focuses on actively discovering missing information, which leads to stronger generalization to unseen environments. Using GRPO-based fine-tuning of VLMs for camera control, we demonstrate that object-centric pose parameterization, defined by a target point and rotation around it, aligns better with human spatial reasoning and produces smoother view transitions. Experiments on synthetic datasets (CLEVR, ProcTHOR) and realistic simulations (Habitat) show that our method significantly improves VQA performance and generalization through active perception.
\ifarxiv
\vspace{-1.5\baselineskip}
\else
\fi 
\section{Introduction}
\label{sec:intro}
\ifarxiv
    \vspace{-0.25\baselineskip}
\else
\fi
%%%%%%%%%%%%%%%%%%%%%%%%%%%%%%%%%%%%%%%%%%%%%%%%%%%%%%%%%%%%%%%%
\begin{flushright}
\begin{minipage}{0.91\linewidth}
\vspace{0.5em}
{
\itshape
``We look around, walk up to something interesting and move around it so as to see it from all sides, and go from one vista to another. That is natural vision.''
}
\vspace{0.5em}
\hrule height 0.4pt
\vspace{0.5em}
\hfill \textsc{James J. Gibson}~\cite{Gibson:1979Ecological}
\end{minipage}
\end{flushright}
%%%%%%%%%%%%%%%%%%%%%%%%%%%%%%%%%%%%%%%%%%%%%%%%%%%%%%%%%%%%%%%%

% \begin{center}
% \begin{minipage}{0.9\linewidth}
% \vspace{0.5em}
% % \hrule height 0.6pt
% % \vspace{0.75em}
% \raggedright\itshape
% ``We look around, walk up to something interesting and move around it so as to see it from all sides, and go from one vista to another. That is natural vision.''

% \vspace{0.25em}
% \raggedleft\upshape---\textsc{James J. Gibson}
% \vspace{0.75em}
% \hrule height 0.6pt
% \vspace{0.5em}
% \end{minipage}
% \end{center}

% Over the past decade of research in Visual Question Answering (VQA), we have now witnessed the rise of Visual–Language Models (VLMs) that have essentially mastered the task, seemingly closing the chapter on this line of inquiry. What, then, should be the next step? We approach this question by revisiting the fundamental purpose of the visual system. James J. Gibson, one of the most influential psychologists of the 20th century, offered crucial insights through his \emph{ecological} approach to vision~\cite{Gibson:1979Ecological}.

We are now witnessing the rise of vision language models (VLMs) that have effectively mastered visual question answering (VQA), seemingly closing the chapter on this line of research. What, then, should come next?

To address this question, we revisit the fundamental purpose of the visual perception system. James J. Gibson, one of the most influential perceptual psychologists of the twentieth century, offered foundational insights through his \emph{ecological} approach to perception~\cite{Gibson:1979Ecological}. According to Gibson, current VLMs operate within what he described as \emph{snapshot vision}, the ability to interpret a single static retinal image. Much of early computer vision research similarly focused on this static image interpretation. Yet Gibson emphasized that the real problem of vision in animals extends far beyond a fixed view. As he wrote, ``natural vision depends on the eyes in the head on a body supported by the ground''. Vision in the real world requires looking around and moving toward objects of interest, an ability he referred to as \emph{ambulatory vision}.

% To address this question, we revisit the fundamental purpose of vision itself. James J. Gibson, one of the most influential psychology researchers of the twentieth century, offered crucial insights through his \emph{ecological} approach to perception \cite{Gibson:1979Ecological}. Current VLMs, despite their impressive capabilities, are confined to interpreting a single static image—a capacity Gibson termed \emph{snapshot vision}. The early history of computer vision can similarly be viewed as the pursuit of this ability: understanding a retinal image. However, the true problem of vision in humans and animals lies beyond this static interpretation. As Gibson wrote, “natural vision depends on the eyes in the head on a body supported by the ground.” In other words, vision in the real world involves looking around, walking toward objects of interest, and observing them from multiple viewpoints—an ability he called \emph{ambulatory vision}.
%Current VLMs, despite their impressive capabilities, remain confined to interpreting static images—a capacity Gibson called \emph{snapshot vision}. The early history of computer vision also aims to this: understanding a retinal image rather than perceiving an environment. In contrast, natural vision, as Gibson noted, ``depends on the eyes in the head on a body supported by the ground''. Vision in the real world therefore requires looking around, moving toward objects of interest, and observing them from multiple viewpoints—an ability he termed \emph{ambulatory vision}.

In this work, we study active perception for VQA by assuming an embodied agent situated in a scene and training it to select the most informative viewpoint for answering a question. Unlike prior approaches, which rely on memorized scene representations~\cite{Yang:20253D-Mem,Saxena:2025GraphEQA} or external commonsense knowledge~\cite{Ren:2024ExploreEQA,Liu:2024EXPRESSBench}, we develop a system that operates solely on the visual information available in the current observation. We refer to this task as Visually-Grounded Active View Selection (VG-AVS).

Active perception has previously been studied under the Embodied Question Answering (EQA) framework~\cite{Das:2018EQA}. EQA typically involves four components: (1) scene exploration, (2) scene memorization, (3) commonsense reasoning, and (4) localization and perception. Most recent work~\cite{Ren:2024ExploreEQA,Liu:2024EXPRESSBench,Yang:20253D-Mem,Saxena:2025GraphEQA} focuses on exploration, memorization, and reasoning, corresponding respectively to planning, representation learning, and language understanding. The fourth component—localization and perception—remains less explored, despite being the core computer-vision challenge. Our study focuses precisely on this aspect and provides a concrete setup for learning active visual perception.

Compared with previous efforts, our formulation advances active perception in three key aspects.
(1) We aim to learn the optimal viewpoint rather than only the agent’s position in the scene \cite{Ren:2024ExploreEQA,Liu:2024EXPRESSBench}. To achieve this, we consider the full set of control parameters for a mobile agent: heading rotation, forward translation, and view rotation.
(2) We pursue fine-grained, continuous control of these parameters instead of coarsely discretizing them into actions such as turn-left, turn-right or move-forward.~\cite{Yang:2025Mindjourney}. Our model predicts continuous rotation angles and translation distances, enabling precise viewpoint adjustment. This formulation avoids the complex multi-turn navigation settings used in prior EQA methods~\cite{Ren:2024ExploreEQA,Liu:2024EXPRESSBench,Yang:20253D-Mem,Saxena:2025GraphEQA}, which makes learning continuous, high-resolution control policies feasible.
(3) Most importantly, unlike previous zero-shot EQA methods~\cite{Ren:2024ExploreEQA,Liu:2024EXPRESSBench,Yang:20253D-Mem,Saxena:2025GraphEQA}, we propose a fine-tuning-based active perception framework and demonstrate its strong generalization to unseen environments and diverse question types.

For training, as one of our key contributions, we introduce a curated synthetic dataset, called ~\Benchmark{} dataset. Built on ProcTHOR~\cite{Deitke:2022ProcTHOR}, each sample consists of a rendered query view and target view. The query view partially includes objects visible in the target view while omitting others, simulating incomplete visual observations. A corresponding question prompt asks about the missing information, requiring the agent to infer the appropriate mobility parameters needed to reach the target viewpoint.

We fine-tune pretrained VLMs using two complementary strategies: supervised fine-tuning (SFT) to learn ground-truth transformations, and reinforcement learning (RL)-based unsupervised fine-tuning, where the model predicts mobility parameters at the end of its reasoning process without explicit supervision. Combining the two in a sequential SFT-then-RL scheme first grounds the model through supervision and then refines it via unsupervised policy optimization, improving both performance and generalization.

% Our experiments show that training-based methods significantly outperform zero-shot approaches in VG-AVS, confirming that learning active visual perception beyond pretrained VLM capabilities leads to more robust and generalizable spatial reasoning. \jk{}
Our experiments show that our training-based methods significantly outperform zero-shot approaches in VG-AVS, confirming the benefit of learning active visual perception beyond pretrained VLM capabilities. Despite being trained only on a small-scale synthetic dataset, our method generalizes to real scenes with diverse question types and can serve as a plug-and-play component that improves existing EQA frameworks.

%In contrast, many previous EQA works focus only on locomotion, where the agent learns to reach a target location but not to choose the most informative view.
%In our experiments, we demonstrate that incorporating our view adaptation module into existing EQA pipelines consistently improves task performance.
%partially observable Markov decision process (POMDP) problems
%As expected, SFT achieves higher training accuracy but tends to overfit the training data, whereas GRPO yields better generalization at the cost of slightly lower overall performance. 
\ifarxiv 
    \vspace{-0.25\baselineskip}
\else 
\fi
\section{Related Work}
\label{sec:related_work}
\ifarxiv 
    \vspace{-0.25\baselineskip}
\else 
\fi

\paragraph{Active Visual Question Answering.}
Beyond solving VQA from static images, several works explore active visual question answering (Active VQA), where the model interacts with visual inputs before answering. PixelReasoner~\cite{Su:2025:PixelReasoner}, ToA~\cite{Liang:2023ToA}, and Directional Guidance~\cite{Liu:2024Rightthisway} all operate in the 2D image space, performing actions such as cropping, selecting key frames, or predicting coarse directions toward regions of interest to obtain more informative views. However, these methods operate strictly in the 2D image space with limited action types, remaining confined to the given frame rather than exploring new viewpoints in the underlying 3D scene.

MindJourney~\cite{Yang:2025Mindjourney} aims to enhance the spatial reasoning of VLMs by generating new observations with a generative model, but its action space is restricted to discrete primitives and actions are selected via beam search rather than by learning a policy. In contrast, we directly learn a policy for an active perception system in physically grounded 3D scenes, operating in a fine-grained, continuous action space.

\ifarxiv 
    \vspace{-1.8\baselineskip}
\else 
    \vspace{-\baselineskip}
\fi
\paragraph{Embodied Question Answering.}
There has been substantial effort to tackle embodied question answering (EQA), both in constructing datasets~\cite{Das:2018EQA,Majumdar:2024OpenEQA,Liu:2024EXPRESSBench} and in designing sophisticated methods~\cite{Ren:2024ExploreEQA,Liu:2024EXPRESSBench,Yang:20253D-Mem,Saxena:2025GraphEQA}. Since EQA requires multiple abilities, such as large-scale scene exploration, scene memorization, common sense reasoning, and active perception, prior work has mainly focused either on efficient scene exploration~\cite{Ren:2024ExploreEQA,Liu:2024EXPRESSBench} or on better 3D scene representations for memory construction~\cite{Yang:20253D-Mem,Saxena:2025GraphEQA}. In contrast, the crucial link between partial visual observations and fine-grained view selection, that is, how an agent should refine its viewpoint based on existing visual clues, remains underexplored. Our work explicitly targets this computer vision challenge and can be adopted as a plug-in module that fills this missing component in EQA, leading to improved overall performance.
\ifarxiv 
    \vspace{-1\baselineskip}
\else 
    \vspace{-\baselineskip}
\fi
\paragraph{Post-Training in VLMs.}
Motivated by the strong success of post-training in the LLM literature~\cite{Wei:2022FLAN, Ouyang:2022InstructGPT,Wang:2023SelfInstruct, Chung:2024ScalingInstruction, Ouyang:2022GPTRLHF, Shao:2024GRPO}, many works have focused on post-training vision-language models (VLMs)~\cite{Wang:2024Qwen2VL,Bai:2025Qwen2.5VL,Li:2024LLaVAOneVision,Team:2025Kimi,Hong:2025GLM,Shen:2025VLM-R1} to enhance their capabilities, including spatial understanding~\cite{Ma:2025SpatialReasoner,Chen:2025SpatialMLLM,Ma:2024SpatialLLM,Chen:2024SpatialVLM,Cheng:2024SpatialRGPT,Junfei:2025VILASR} and multi-view or 3D understanding~\cite{Zhu:2024LLaVA3D,deng:20253D-LLAVA, Gholami:2025EGO3D-VLM}. In particular, GRPO-based reinforcement learning approaches~\cite{Shao:2024GRPO, Yu:2025DAPO, zheng:2025GSPO} have demonstrated that post-training can improve model performance by optimizing over the model’s own reasoning process without requiring additional human annotations or explicit supervision. Another line of work fine-tunes VLMs to endow them with action-oriented decision-making abilities~\cite{zhai:2024VLMDecision-Making,Liu:2025RL-VLA,molmoact2025,Huang:2025ThinkAct,Kim2025Robot-R1}. However, none of these works seeks to improve active perception, the ability to compensate for missing visual evidence in the current view.

\ifarxiv
\vspace{-0.25\baselineskip}
\else
\fi
%%%%%%%%%%%%%%%%%%%%%%%%%%%%%%%%
\begin{figure}
    \centering
    \includegraphics[width=\linewidth]{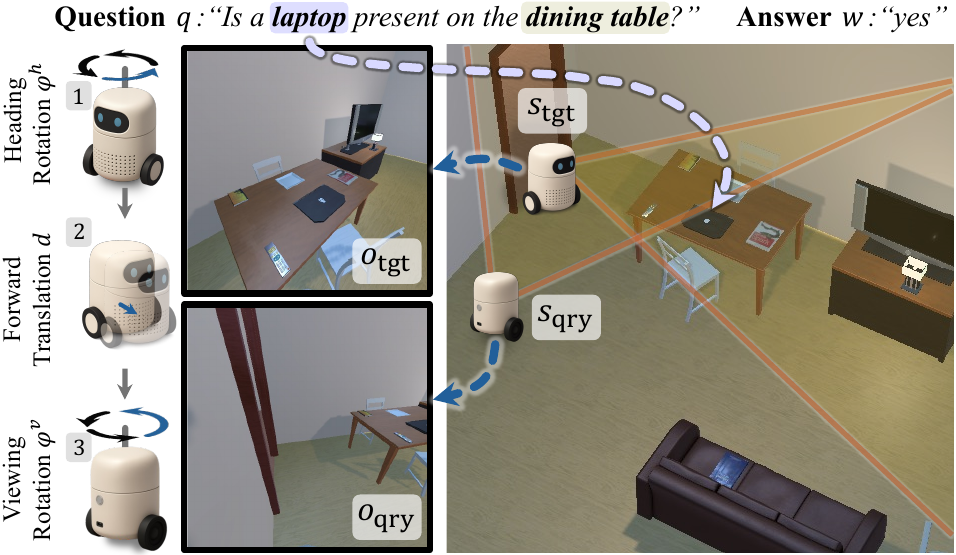}
    \ifarxiv 
        \vspace{-1\baselineskip}
    \else 
        \vspace{-0.5\baselineskip}
    \fi
        \caption{\textbf{Our action sequence (left) and \Benchmark{} dataset sample (right).} The action controls how the agent adjusts its viewpoint. On the right, the target object (\textit{laptop}) is visible only in the target view, while the query view shows only the supporting object (\textit{dining table}), which may serve as a visual clue that motivates active view selection to gather sufficient visual evidence for answering.}
    \label{fig:dataset_action}
    \vspace{-1.5\baselineskip}
\end{figure}
%%%%%%%%%%%%%%%%%%%%%%%%%%%%%%%%
\vspace{-0.5\baselineskip}
\section{Visually-Grounded Active View Selection}
\label{sec:method_overview}
\vspace{-0.25\baselineskip}
We study the problem of learning an \emph{active perception system} that adjusts its egocentric viewpoint to acquire more informative observations for answering a language query. 

Beyond passive answering from static images, our objective is to cultivate an active perception ability, namely the capacity to recognize when the current view lacks sufficient visual evidence, identify informative cues from the current visual-linguistic context, and actively adjust the viewpoint to acquire the missing information. Through this process, the model learns to perform question-driven view selection in a learning-based manner, ultimately enhancing its visual grounding and improving VQA performance.

Unlike prior works on Active VQA~\cite{Liang:2023ToA, Ma:2025SpatialReasoner, Liu:2024Rightthisway}, which typically define the action space in the 2D image domain and apply simple visual operators such as cropping or zooming to focus on salient regions, we consider more spatially grounded scenarios where the agent is situated in a 3D environment. Our formulation enables the agent to refine its viewpoint through explicit locomotion and viewpoint rotation, demanding fine-grained control within a substantially larger action space.

To realize this active perception capability, we introduce three key components: a dedicated training dataset in Section~\ref{sec:data_curation}, a systematically designed framework for continuous active view selection in Section~\ref{sec:problem_formulation}, and an effective two-stage training strategy in Section~\ref{sec:two_stage_strategy}. 
\vspace{-0.25\baselineskip}
\subsection{\Benchmark{} Dataset}
\label{sec:data_curation}
\vspace{-0.25\baselineskip}
To enable such visually-grounded active view selection, it is essential to construct training data that explicitly supports this capability. To this end, we curate a dataset, called~\Benchmark{}, designed to train models that actively adjust their viewpoints to gather sufficient visual evidence for answering a given question.

Each sample in the dataset is represented as a tuple $(q, w, s_{\text{tgt}}, s_{\text{qry}}, o_{\text{tgt}}, o_{\text{qry}})$, where $q$ denotes the language query, $w$ its corresponding answer, $s_{\text{tgt}}$ and $s_{\text{qry}}$ indicate the target and query camera poses,
and $o_{\text{tgt}}$ and $o_{\text{qry}}$ are their corresponding rendered observations. 

The target view $o_{\text{tgt}}$ contains sufficient visual information to answer the query, while the query view $o_{\text{qry}}$ provides only partial visual evidence, encouraging the model to actively adjust its viewpoint to reveal the missing information. To capture this contextual difference between the two views within each training sample, we build the~\Benchmark{} dataset using the ProcTHOR~\cite{Deitke:2022ProcTHOR} environment, which provides richly annotated indoor 3D scenes with instance-level object labels and explicit surface–object relations, where certain objects are designated as \emph{supporting} objects that serve as surfaces for other assets, such as countertops holding cups or beds supporting pillows, and these supporting objects could offer visual cues about the unseen \emph{target} object referred to by the question.

Below is an overview of our automatic data curation pipeline. We first select an object of interest and generate a corresponding question–answer pair $(q, w)$ referring to that object.
We then sample an answerable target view $o_{\text{tgt}}$ and a contextual (query) view $o_{\text{qry}}$.
To determine these views, we leverage the instance labels to render each viewpoint together with its instance segmentation mask, which provides pixel-level visibility information for every object in the scene.

Let $x_{\text{tgt}}$ and $x_{\text{sup}}$ denote the target object of interest and its supporting object, respectively. We use $N_p(o,x)$ to represent the number of pixels in view $o$ that belong to object $x$, and $c(o, x)$ to denote the normalized distance between the projected centroid of $x$ and the image center. We define three thresholds $\epsilon^{\text{sup}}_{\text{vis}} > \epsilon^{\text{obj}}_{\text{vis}} > \epsilon^{\text{obj}}_{\text{inv}}$, 
corresponding to the visibility thresholds for the supporting and target objects, and the invisibility threshold for the target object, respectively. Then, the target and query views are defined as:
\begin{align}
\label{eq:o_tgt_sample}
o_{\text{tgt}} &\sim \{ o \mid N_p(o, x_{\text{tgt}}) > \epsilon^{\text{obj}}_{\text{vis}},\, c(o, x_{\text{tgt}}) < \delta_{\text{center}} \} \\
\label{eq:o_qry_sample}
o_{\text{qry}} &\sim \{ o \mid N_p(o, x_{\text{tgt}}) < \epsilon^{\text{obj}}_{\text{inv}}, \, N_p(o, x_{\text{sup}}) > \epsilon^{\text{sup}}_{\text{vis}} \}
\end{align}
where $\delta_{\text{center}}$ specifies the maximum allowed distance for the target to be regarded as centered. Given the selected target and supporting objects, we construct the corresponding question–answer pair by instantiating a predefined question template. The illustration of the dataset example is in the right of Figure \ref{fig:dataset_action}, and further implementation details are provided in the \Suppl{}. 

The resulting training set consists of 1,320 scenes with 1,867 tuple samples focused solely on binary \emph{existence} questions. 

For evaluation, we construct two additional benchmark datasets that cover more diverse question types and real indoor scenes.
\Benchmark{}-ProcTHOR extends beyond the binary object existence questions used for training to include three question types: \emph{existence}, \emph{counting}, and \emph{state}. Furthermore, \Benchmark{}-HM3D is built from real-world indoor scenes in the Habitat–Matterport 3D dataset~\cite{Ramakrishnan:2021HM3D}.
Details of these benchmarks are presented in Section~\ref{sec:experiment_results}.

% \begin{figure}
%     \centering
%     \includegraphics[width=1\linewidth]{figures/action.pdf}
%     \vspace{-0.5\baselineskip}
%     \caption{\textbf{Illustration of our action space design.} In every turn, each action is executed sequentially: (1) a heading rotation $\varphi^h$ that determines the movement direction relative to the current azimuth $\varphi$. (2) a forward translation $d$ along the rotated $\varphi^v$ that adjusts the camera’s azimuth after the translation. These three components compactly span all physically executable motions of the embodied agent under our formulation.}
%     \label{fig:action_space}
%     \vspace{-\baselineskip}
% \end{figure}

\subsection{\Method{} Framework}
\label{sec:problem_formulation}
\vspace{-0.25\baselineskip}
% As discussed in Section~\ref{sec:method_overview}, we formalize the active view selection problem as learning a policy that samples an action moving the agent from a contextual viewpoint toward the \emph{answerable observation space}.

% Formally, we formulate the view selection problem as a continuous decision-making problem, 

As discussed in Section~\ref{sec:method_overview}, we cast active view selection as learning a policy that moves the agent from a contextual viewpoint toward the \emph{answerable observation space}.

Formally, we model this as a continuous decision-making problem,
where the agent predicts a real-valued action that adjusts its egocentric viewpoint to acquire sufficient visual evidence for answering a given question.
Let the state $s \in \mathcal{S}$ represent the agent's 3D position and orientation in the 3D environment, and the corresponding observation $o=\Omega(s)$ be the egocentric RGB view obtained from $s$ by the observation function $\Omega(\cdot)$. The policy $\pi_\theta(\cdot \vert o, q)$ outputs an action $a \in \mathcal{A}$, such as rotation angles and translation distances, that determines how the agent should move to acquire sufficient visual evidence for answering the language query $q$. Executing $a$ in the environment updates the state via $s'=\mathcal{T}(s,a)$ and produces a new observation $o'=\Omega(s')$. This continuous formulation enables the agent to directly reach an answerable viewpoint with a single fine-grained action, rather than relying on multiple discrete steps, thereby simplifying training from a multi-step process to a single-step policy optimization. We next detail the design of our action space below.

% into a one-step control objective.

% Formally, we formulate the viewpoint selection problem as a continuous partially observable Markov decision process (POMDP)
% $\mathcal{E} = (\mathcal{S}, \mathcal{A}, \mathcal{O}, \mathcal{T}, \Omega, \rho, \gamma)$,
% where the state $s \in \mathcal{S}$ represents the agent’s 3D position and orientation, and the observation $o \in \mathcal{O}$ is the egocentric RGB view rendered by the observation model $\Omega(o \mid s)$.
% Both the state $s$ and action $a \in \mathcal{A}$ are continuous (e.g., rotation angles and movement distances), enabling the agent to directly reach an answerable viewpoint with a single action instead of performing multiple discrete turns.
% This formulation simplifies training by reducing the multi-step decision process to a single-step policy optimization.
% The transition function $s' = \mathcal{T}(s, a)$ defines how the state evolves given an action. We next detail the design of our action space below.

\vspace{-\baselineskip}
\paragraph{Action Space Design.}
We represent the state $s$ as a triplet $s = (x, y, \varphi)$, where $(x, y)$ denotes the agent’s 2D position and $\varphi$ is the azimuth orientation. We fix the height and elevation angle for simplicity.
The action space is parameterized to compactly span all physically executable movements of the embodied agent using a minimal set of continuous components, resulting in a triplet $a = (\varphi^h, d, \varphi^v)$:
\begin{itemize}
\item \textbf{Heading rotation} $\varphi^h \in (-180^\circ, 180^\circ]$:
azimuthal turning angle that determines the moving direction relative to the agent’s current orientation $\varphi$.
\item \textbf{Forward translation} $d \ge 0$:
distance to move forward along the rotated heading.
\item \textbf{View rotation} $\varphi^v \in (-180^\circ, 180^\circ]$:
final azimuthal offset of the agent relative to the moving direction, modeling a fine-grained head turning.
\end{itemize}
All azimuthal angles follow the same convention, where positive angles indicate right turns.

The action is performed sequentially: the agent first determines its heading direction by $\varphi^h$, then moves forward by $d$ along that heading, and finally adjusts its viewing direction by applying the view rotation $\varphi^v$ to the current azimuth orientation. Accordingly, the state transition is given by:
%%%%%%%%%%%%%%%%%%%%%%%%
\begin{align}
    s' = \begin{pmatrix}
        x' \\
        y' \\
        \varphi'
    \end{pmatrix}=
    \mathcal{T}(s, a) =
    \begin{pmatrix}
        x + d \sin(\varphi + \varphi^h) \\
        y + d \cos(\varphi + \varphi^h) \\
        \varphi + \varphi^h + \varphi^v
    \end{pmatrix}
    .
\end{align}
%%%%%%%%%%%%%%%%%%%%%%%%
Given two states, specifically $s’ = s_\Tgt$ and $s = s_\Qry$, the ground-truth action $a_\Tgt$ that moves the agent from the query state $s_\Qry$ to the target state $s_\Tgt$ can be analytically computed as follows:
% we first define the positional differences as $\Delta x = x' - x$ and $\Delta y = y' - y$.
% Then, the action parameters are given by
%%%%%%%%%%%%%%%%%%%%%%%%
\begin{align}
\label{eq:action_computation}
a_{\text{tgt}} =
\begin{pmatrix}
    \varphi^h_{\text{tgt}} \\
    d_{\text{tgt}} \\
    \varphi^v_{\text{tgt}}
\end{pmatrix}=
\begin{pmatrix}
\operatorname{atan2}(\Delta x, \Delta y) - \varphi \\
\sqrt{(\Delta x)^2 + (\Delta y)^2} \\
\varphi' - (\varphi + \varphi^h_{\text{tgt}})
\end{pmatrix},
% d &= \sqrt{(\Delta x)^2 + (\Delta y)^2}, \\
% \varphi^h &= \operatorname{atan2}(\Delta y, \Delta x) - \varphi, \\
% \varphi^v &= \varphi' - (\varphi + \varphi^h).
\end{align}
%%%%%%%%%%%%%%%%%%%%%%%%
where $\Delta x = x' - x$ and $\Delta y = y' - y$ denote the positional differences. See the left of Figure~\ref{fig:dataset_action} for an illustration of action space design.\\
% \dc{x: sin, y: cos, $\varphi^h_{tgt} = \operatorname{atan2}(\Delta x, \Delta y)$, right? \\}

\begin{figure*}
    \centering
    \includegraphics[width=1\linewidth]{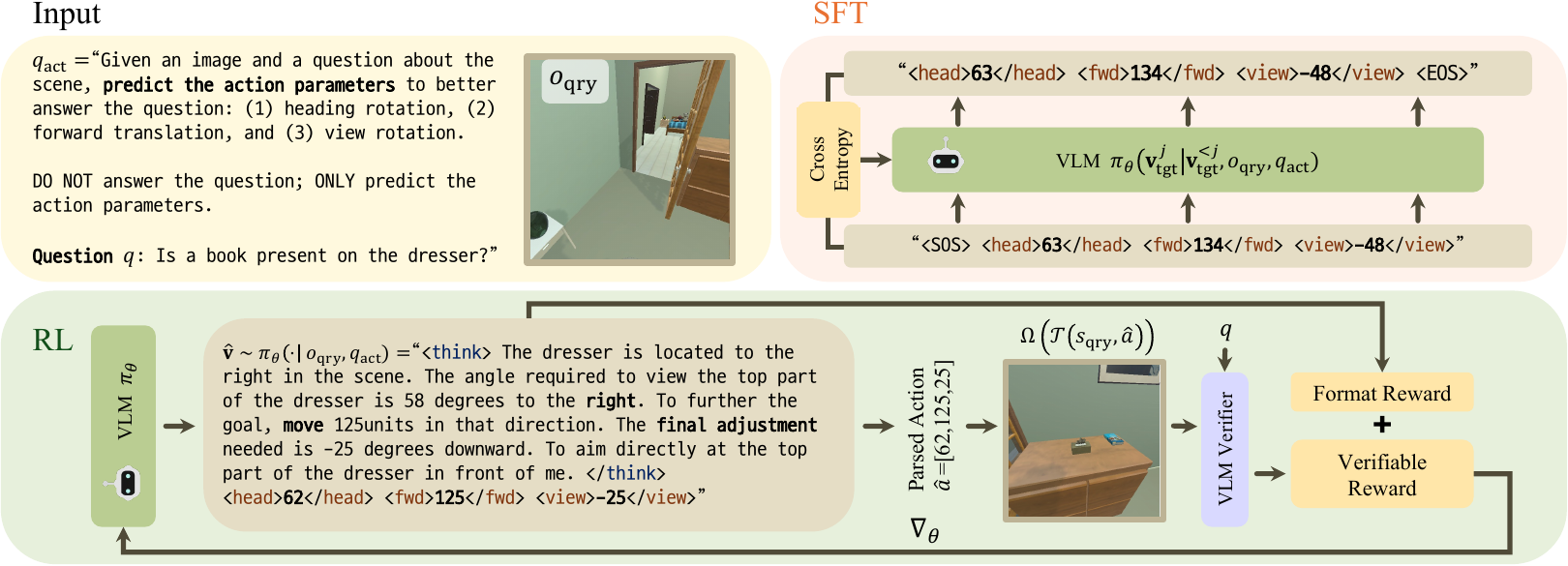}
    \vspace{-\baselineskip}
    \caption{\textbf{Overview of training strategies.} For a given query image $o_{\text{qry}}$ and question $q_{\text{act}}$, the action model is trained to predict action to obtain next view. In SFT, the model is trained to directly predict ground-truth actions from the input. In contrast, under RL, the model first generates a token sequence that includes its own reasoning process, followed by the final action prediction, and the model is optimized so that outputs leading to higher rewards become more likely.}
    \label{fig:placeholder}
    \vspace{-\baselineskip}
\end{figure*}
\vspace{-\baselineskip}
\subsubsection{Supervised Fine-Tuning}
\label{sec:sft}
Under the problem formulation discussed in Section~\ref{sec:problem_formulation}, the most straightforward approach is to adopt a teacher-forcing objective for learning the policy, where the model is supervised to predict the ground-truth action given the input observation and language query. 

Note that such supervised fine-tuning is feasible in our setup since our synthetic dataset provides access to ground-truth target views, enabling an analytic computation of the ground-truth action between paired query and target viewpoints.
In contrast, existing EQA setups lack such view-level supervision, as ground-truth actions are difficult to define or annotate due to long-horizon navigation trajectories.

Specifically, we parameterize the policy with a VLM. To enable the VLM to predict real-valued actions, we introduce a function $\texttt{str}(\cdot)$ that maps real-valued numbers to their string representations.
Given an action $a = (\varphi^h, d, \varphi^v)$, we convert it into the following formatted string representation: $\mathbf{v}=$\texttt{<H>}\texttt{str}$(\varphi^h)$\texttt{</H>} \texttt{<D>}\texttt{str}$(d)$\texttt{</D>} \texttt{<V>}\texttt{str}$(\varphi^v)$\texttt{</V>}, where \texttt{<H>}, \texttt{<D>}, and \texttt{<V>} are special tags to explicitly delimit each scalar component.

Given a tuple $(q,w,s_\Tgt, s_\Qry, o_\Tgt, o_\Qry)$ sampled from our ~\Benchmark{} dataset, we first compute the ground-truth target action $a_\Tgt$ following Equation~\ref{eq:action_computation}, which specifies how the agent should move from the query viewpoint to the target one. The teacher-forcing objective for SFT is expressed as follows:
%%%%%%%%%%%%%%%%%%%%%%%%%%%%%%%%
\begin{align}
\mathcal{L}_{\text{SFT}}=
\sum_{j=1}^{L}
\log \pi_\theta(\mathbf{v}_{\text{tgt}}^j \mid \mathbf{v}_{\text{tgt}}^{<j}, o_{\text{qry}}, q_{\text{act}}),
\end{align}
%%%%%%%%%%%%%%%%%%%%%%%%%%%%%%%%
where $\mathbf{v}_{\text{tgt}}$ denotes the string representation of $a_\Tgt$, and 
$q_{\text{act}}$ is an action-specific instruction expanded from $q$, which explicitly prompts the model to predict the action rather than directly answering the question.

% We also expand $q$ into an action-specific instruction $q_{\text{act}}$, which explicitly prompts the model to predict the action rather than directly answering the question.
% We denote the string representation of $a_{\text{tgt}}$ as $\mathbf{v}_{\text{tgt}}^{1:L}$. The SFT objective is defined as:

While SFT enables the model to learn which action to predict from a multimodal context, it confines the VLM to human-annotated actions.

%%%%%%%%%%%%%%%%%%%%%%%%%%%%%%%%%%%%%%%%
% \definecolor{opensourcecolor}{HTML}{cdb4db}
% \definecolor{spatialcolor}{HTML}{ffc8dd}
% \definecolor{eqacolor}{HTML}{ffafcc}
% \definecolor{procolor}{HTML}{bde0fe}
% \definecolor{ourscolor}{HTML}{a2d2ff}

% \definecolor{opensourcecolor}{HTML}{f19c79}
% \definecolor{spatialcolor}{HTML}{ffe1a8} 
% \definecolor{eqacolor}{HTML}{f6f4d2}
% \definecolor{procolor}{HTML}{cbdfbd}
% \definecolor{ourscolor}{HTML}{d4e09b}

\definecolor{opensourcecolor}{HTML}{0072B2} 
\definecolor{spatialcolor}{HTML}{E69F00}   
\definecolor{eqacolor}{HTML}{009E73}      
\definecolor{procolor}{HTML}{D55E00}     
\definecolor{ourscolor}{HTML}{cbdfbd}     

\begin{table*}[h!]
    \centering
    \footnotesize
    \caption{\textbf{Quantitative comparison on our \Benchmark{} benchmark.} We report VQA accuracy on \Benchmark{}-ProcTHOR and LLM-Match scores on \Benchmark{}-HM3D, normalized to a percentage scale. The best except for \textit{`No Action'} in each column is in \textbf{bold} and second best is \underline{underlined}.}
    \label{tab:vgavs_main}
    \vspace{-0.5\baselineskip}
    \setlength{\tabcolsep}{2pt}
    % Define a light gray color for highlighting 'Ours'
    % Setup for siunitx columns, assuming scores are like XX.X
    % \sisetup{detect-weight, mode=text}
    \begin{tabularx}{\linewidth}{
        >{\raggedright\arraybackslash}p{2.4cm} >{\raggedright\arraybackslash}p{2.5cm} |
        Y Y Y | Y |  % ProcTHOR: Ex, Cnt, State | Average
        Y Y Y Y Y | Y  % HM3D: Attr, Ex, Cnt, Obj, State | Average
    }
    \toprule
    \multicolumn{2}{c|}{\multirow{2}{*}{\textbf{Action Model}}}  & \multicolumn{4}{c|}{\textbf{AVS-ProcTHOR}~\cite{Deitke:2022ProcTHOR}} & \multicolumn{6}{c}{\textbf{AVS-HM3D}~\cite{Ramakrishnan:2021HM3D}} \\
    \cmidrule(lr){3-6}\cmidrule(lr){7-12}
    & & \scriptsize{\textbf{Existence}} & \scriptsize{\textbf{Counting}} & \scriptsize{\textbf{State}} & \scriptsize{\textbf{Average}} 
    & \scriptsize{\textbf{Existence}} & \scriptsize{\textbf{Counting}} & \scriptsize{\textbf{State}} & \scriptsize{\textbf{Attribute}} & \scriptsize{\textbf{Object}} & \scriptsize{\textbf{Average}} \\
    \midrule
    \rowcolor{Gray}& Query view        & 49.22 & 16.36 & 61.57 & 42.38 & 67.50 & 56.15 & 54.59 & 66.67 & 48.33 &  58.65\\
    \rowcolor{Gray}\multirow{-2}{*}{\textbf{No Action}}& Target view       & 93.02 & 69.14 & 92.58 & 84.91 & 86.67 &  77.50 & 80.00 & 74.48 & 71.79  & 78.09 \\
    \midrule
    \textbf{Backbone Model} & \cellcolor{opensourcecolor!18} Qwen2.5-VL-7B~\cite{Bai:2025Qwen2.5VL}            & 64.34 & 29.74 & 56.55 & 50.21 & 66.25 & 46.15 & 66.49 & 49.17 & 46.86 & 54.98 \\
    \midrule
    \addlinespace[0.5ex]
    \multirow{2}{*}{\textbf{Spatial VLMs}} & \cellcolor{spatialcolor!18}ViLaSR~\cite{Junfei:2025VILASR}                   & 57.95 & 25.46 & 52.84 & 45.42 & 68.33 & 48.46 & 52.70 & 50.00 & 50.00 & 53.90 \\
    & \cellcolor{spatialcolor!18}SpatialReasoner~\cite{Ma:2025SpatialReasoner}          & 54.65 & 22.68 & 52.62 & 43.32 & 70.42 & 50.00 & 47.30 & 37.50 & 40.00 & 49.04 \\
    % \addlinespace[1ex]
    \midrule
    \addlinespace[0.5ex]
    \textbf{EQA Framework} & \cellcolor{eqacolor!18}Fine-EQA~\cite{Liu:2024EXPRESSBench}                   & 63.57 & 31.97 & 64.41 & 53.32 & 70.00 & 52.31 & 52.70 & 54.17 & 44.44 & 54.72 \\
    \midrule
    \addlinespace[0.5ex]
    \multirow{2}{*}{\textbf{Proprietary Models}} & \cellcolor{procolor!18}GPT-5~\cite{OpenAI:2025GPT5}                    & 81.01 & 55.58 & 79.69 & 72.09 & \underline{76.67} & 54.62 & 65.95 & 65.83 & 60.00 & 64.91 \\
    & \cellcolor{procolor!18}Gemini-2.5-Pro~\cite{Google:2025Gemini25}         & 82.95 & 52.79 & 81.00 & 72.25 & 74.17 & 59.23 & 60.81 & 64.17 & 60.00 & 63.67 \\
    \midrule
    \addlinespace[0.5ex]
    \multirow{3}{*}{\makecell[l]{\textbf{AVS Framework}\\(Backbone:\\Qwen2.5-VL-7B~\cite{Bai:2025Qwen2.5VL})}} & \cellcolor{ourscolor}SFT               & \underline{91.28} & 57.06 & \underline{83.84} & 77.39 & 67.50 & \underline{70.77} & 62.16 & 66.67 & 55.56 & 64.53 \\
    & \cellcolor{ourscolor}RL             & 86.82 & \underline{65.24} & 83.41 & \underline{78.49} & \textbf{81.25} & 70.00 & \underline{72.97} & \underline{69.17} & \underline{60.00}  & \underline{70.68} \\
    & \cellcolor{ourscolor}\textbf{SFT+RL (Ours)}              & \textbf{91.47} & \textbf{69.52} & \textbf{90.17} & \textbf{83.72} & 74.58 & \textbf{71.54} & \textbf{73.78} & \textbf{70.83} & \textbf{62.78} & \textbf{70.70} \\
    \bottomrule
    \end{tabularx}
    \vspace{-\baselineskip}
\end{table*}

%%%%%%%%%%%%%%%%%%%%%%%%%%%%%%%%%%%%%%%%

% While SFT effectively enables the model to grasp what action should be predicted from a multimodal context, it primarily fits the VLM to human-annotated actions.
% In contrast, our subsequent reinforcement learning stage encourages the VLM to enhance its performance and generalizability through its own chain-of-thought reasoning and active exploration, rather than merely imitating annotated examples.

% \subsection{Two-Stage Training Strategy}
% Under the formulation discussed in Section~\ref{sec:problem_formulation}, the objective is to learn a policy $\pi_\theta(a \mid o_{\text{qry}}, q)$ that samples actions $a$, guiding the agent from the contextual viewpoint toward the answerable observation space given the current observation and language query.
% We parameterize the policy with a VLM, and train the VLM with a simple yet effective two-stage training strategy. 

% We train it in two stages:
% (1) Supervised Fine-Tuning (SFT) to establish an initial understanding of action semantics, and
% (2) Reinforcement Learning (RL) to explore the continuous action space, enabling the model to discover more optimal viewpoints beyond human annotations and to generalize effectively to unseen environments (see Section~\ref{}).

\subsubsection{Reinforcement Learning}
\label{sec:rl}
Another direction for learning the policy is to employ reinforcement learning (RL), which leverages the model’s internal chain-of-thought reasoning process to arrive at the final action prediction without explicit supervision. 
Once the model generates a token sequence through its reasoning process, denoted as $\hat{\mathbf{v}}=$\texttt{<think>...</think>} \texttt{<H>h}\texttt{</H>} \texttt{<D>d}\texttt{</D>} \texttt{<V>v}\texttt{</V>}, we denote the executable real-valued action parsed from this output as $\hat{a}$. The policy is optimized to maximize rewards computed from the generated sequence. We employ a verifiable reward $r^{\text{ver}}$ that converts each predicted view into a binary feedback signal, together with a format reward $r^{\text{fmt}}$ that encourages the model to produce a correctly formatted action string. 

Concretely, the verifiable reward $r^{\text{ver}}$ is measured using a frozen pre-trained VLM $v_\phi$ that acts as an external verifier and checks whether the question can be correctly answered from the predicted view, defined as:
\begin{align}
r^{\text{ver}}(o, q, w) =
\begin{cases}
1, & \text{if } v_\phi(o, q) = w, \\
0, & \text{otherwise},
\end{cases}
\end{align}
where $(q,w)$ is a question-answer pair and $o$ denotes the input observation.

Formally, the RL objective maximizes the expected reward over sampled tokens:
\begin{align}
\max_\theta
\mathbb{E}_{\hat{\mathbf{v}}\sim\pi_\theta(\cdot \vert o_\Qry, q_{\text{act}})}
\left[
r^\text{fmt}(\hat{\mathbf{v}}) + r^{\text{ver}}\big(\hat{o}, q, w\big)
\right],
\end{align}
where $\hat{o}=\Omega(\mathcal{T}(s_{\text{qry}}, \hat{a}))$ with $\hat{a}$ denoting the real-valued action parsed from the sampled token sequence $\hat{\mathbf{v}}$. We use Group Relative Policy Optimization (GRPO)~\cite{Shao:2024GRPO} for policy optimization.

\subsubsection{Bridging SFT and RL}
\label{sec:two_stage_strategy}
While SFT enables the model to efficiently learn plausible actions from visual–linguistic inputs through explicit supervision from paired query–target viewpoints in our dataset, we empirically observe that SFT alone quickly saturates and provides limited further gains once the basic mapping from observations to actions is learned.

On the other hand, when the model is trained with RL from scratch, the resulting policy typically underperforms the SFT-trained model and is less stable in the continuous action space, even with carefully designed rewards.

Empirically, we find that combining the two objectives in a staged manner is crucial: warming up the policy with SFT provides a good initialization that captures plausible action magnitudes and directions, and subsequent fine-tuning with RL brings additional improvements by refining these actions under task-specific rewards through the model's own reasoning process. This two-stage training strategy yields better performance than using either SFT or RL alone.

\vspace{-0.25\baselineskip}
\section{Experiments}
\label{sec:experiment_results}
\vspace{-0.1\baselineskip}
See the \Suppl{} for training details, extended comparisons, and cross-dataset generalization.
% To demonstrate the effectiveness of our learning-based active perception system, we conduct a series of experiments covering multiple aspects.
% In Section~\ref{sec:act2answer_results}, we introduce the setup of the proposed \Benchmark{} benchmark and evaluate our system’s ability to generate actions that guide the agent toward answerable viewpoints based on partial observations and language queries. These experiments are conducted both in synthetic environments and real-world indoor 3D scenes to assess generalizability.

% In Section~\ref{}, we further show that our learned active perception system for ambulatory vision can be integrated into existing EQA frameworks to improve their answer accuracy, which typically lacks fine-grained control over the agent’s final viewpoint.
% Finally, we demonstrate that the VLM trained on our \Benchmark{} dataset improves general spatial understanding even in out-of-task scenarios not seen during training, highlighting its broader generalization ability.

\subsection{Experiments on VG-AVS}
\label{sec:vgavs_results}
\vspace{-0.1\baselineskip}
%%%%%%%%%%%%%%%%%%%%%%%%%%%%%%%%%%%%%%%%
% Preamble (once)
% \usepackage{graphicx,booktabs,tabularx,array,xcolor}
% \newcolumntype{Y}{>{\centering\arraybackslash}X}
% \setlength{\tabcolsep}{1pt}
% preamble

%\definecolor{wrongblue}{RGB}{38,82,170}
%\definecolor{correctred}{RGB}{192,57,43}

% \definecolor{correctgreen}{RGB}{105,182,28}
% \definecolor{wrongred}{RGB}{243,43,12}
% \definecolor{correctgreen}{RGB}{132,153,79}
% \definecolor{wrongred}{RGB}{243,1,3}

\newcommand{\imgframe}[2]{
\begingroup \setlength{\fboxsep}{0pt}
\setlength{\fboxrule}{1.3pt}% 테두리 두께 (원하면 조절) 
\fcolorbox{#1}{white}{\includegraphics[width=1.0\linewidth,keepaspectratio]{#2}}% 
\endgroup }

\newcommand{\hcell}[1]{\makecell{\scriptsize #1}}
\begin{figure*}[t]
\scriptsize
\centering
\renewcommand{\arraystretch}{1.0}
\setlength{\tabcolsep}{1pt}
\arrayrulecolor{black}
\setlength{\arrayrulewidth}{0.2pt}

\renewcommand{\receptacle}[1]{%
  \begingroup
  \setlength{\fboxsep}{2pt}% 기본값(3pt)보다 훨씬 작게
  \colorbox{olive!15}{#1}%
  \endgroup
}
\renewcommand{\targetobj}[1]{%
  \begingroup
  \setlength{\fboxsep}{2pt}% 기본값(3pt)보다 훨씬 작게
  \colorbox{blue!15}{#1}%
  \endgroup
}

\newcommand{\wrongimg}[2][]{%
  \begin{overpic}[#1]{#2}
    \put(3,75){\color{wrongred}\fontsize{15}{15}\selectfont\bfseries\xmark}
  \end{overpic}
}

\newcommand{\correctimg}[2][]{%
  \begin{overpic}[#1]{#2}
    \put(3,75){\color{correctgreen}\fontsize{15}{15}\selectfont\bfseries\cmark}
  \end{overpic}
}

\begin{tabularx}{\linewidth}{YYYYYYYYY}
\toprule
% ===== Category header row =====
& \multicolumn{1}{c|}{} & \hcell{EQA\\Framework} & \hcell{Backbone\\Model} & \hcell{Proprietary\\Model} & \hcell{Spatial\\VLM} & \multicolumn{3}{|c}{\hcell{AVS Framework}} \\
\midrule
% ===== Model-name header row (no bold, smaller font) =====
Query View & \multicolumn{1}{c|}{Target View} & \hcell{Fine-EQA~\cite{Liu:2024EXPRESSBench}} & \hcell{Qwen-2.5-VL~\cite{Bai:2025Qwen2.5VL}} & \hcell{Gemini-2.5\\-Pro~\cite{Google:2025Gemini25}} & \hcell{ViLaSR~\cite{Junfei:2025VILASR}} & \multicolumn{1}{|c}{\hcell{SFT}} & \hcell{RL} & \hcell{\textbf{SFT+RL (Ours)}} \\
\midrule
%\multicolumn{9}{c}{\Benchmark{}-ProcTHOR}
%\midrule
% ---- Rows (GPT / VILA-SR set) ----
%\RowNine{figures/quali_cropped/procthor/counting_gpt_vilasr_0063}{(Counting) ``\textit{ How many \textbf{cups} on the \textbf{sofa}?}''}
\multicolumn{9}{c}{\footnotesize (Counting) ``\textit{How many \targetobj{\textbf{mugs}} near the \receptacle{\textbf{diningtable}}?}''}\\[4pt]
\includegraphics[width=\linewidth,keepaspectratio]{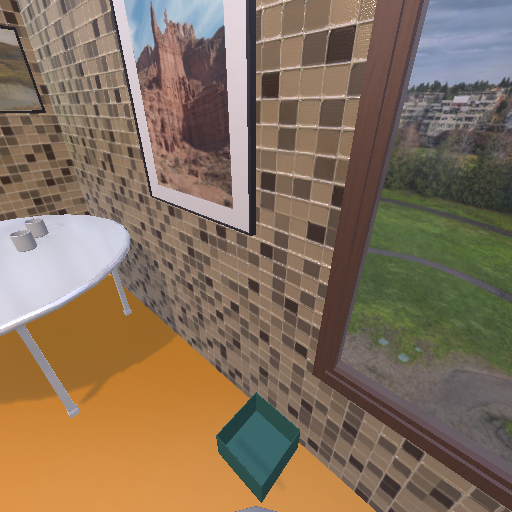} &
\includegraphics[width=\linewidth,keepaspectratio]{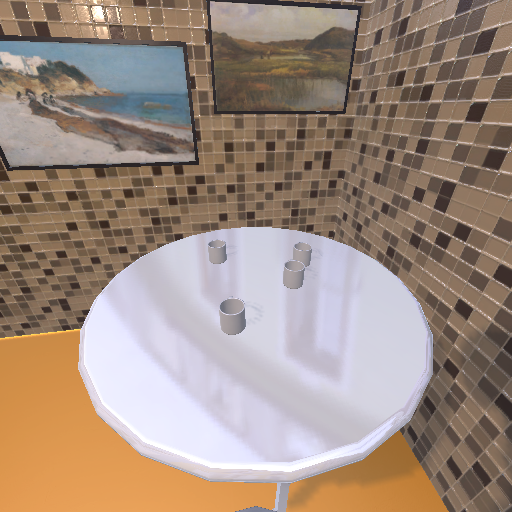} &
\wrongimg[width=\linewidth,keepaspectratio]{{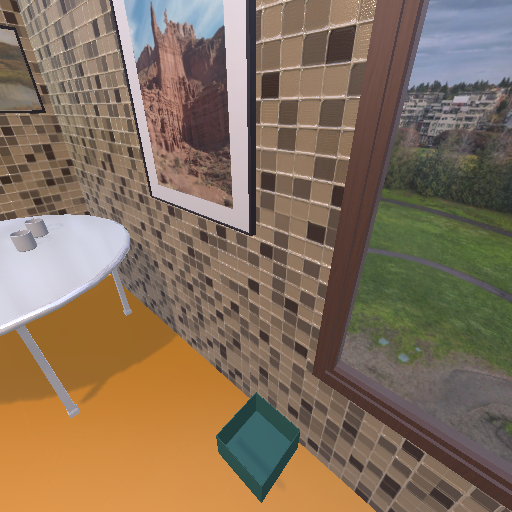}} &
\wrongimg[width=\linewidth,keepaspectratio]{{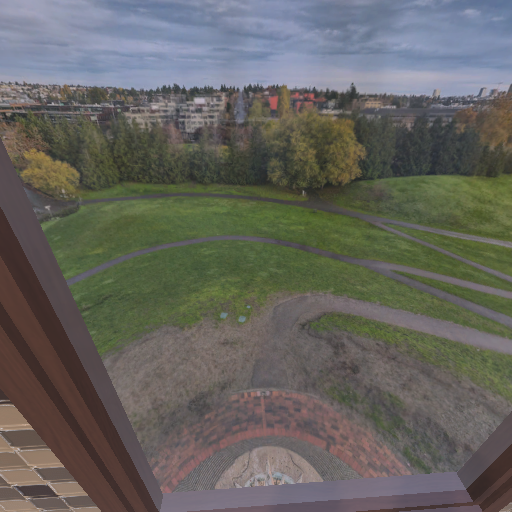}} &
\wrongimg[width=\linewidth,keepaspectratio]{{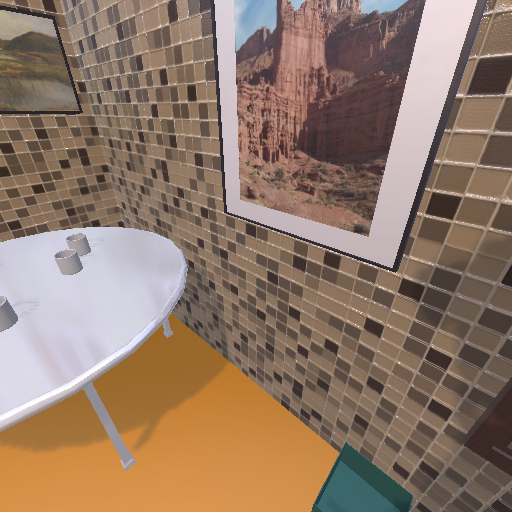}} &
\wrongimg[width=\linewidth,keepaspectratio]{{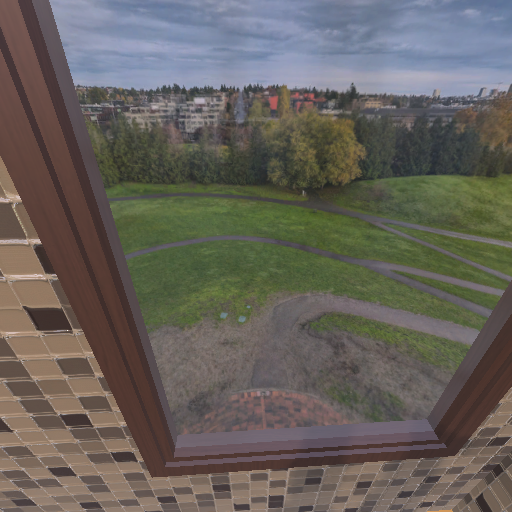}} &
\correctimg[width=\linewidth,keepaspectratio]{{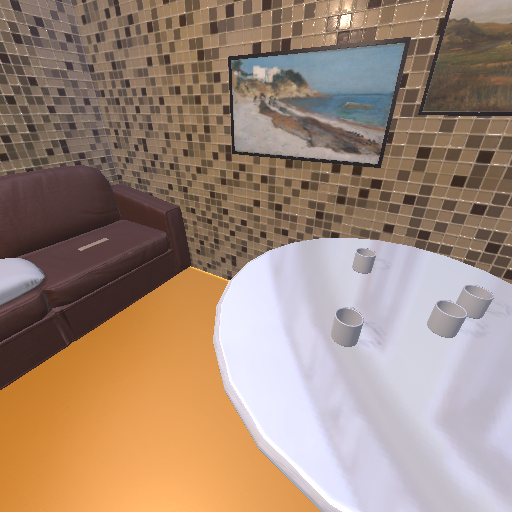}} &
\correctimg[width=\linewidth,keepaspectratio]{{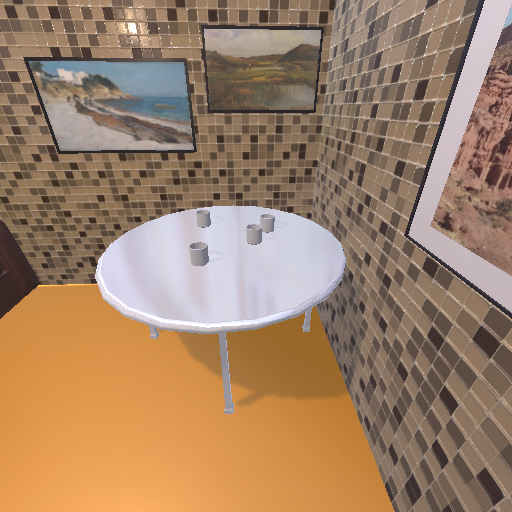}} &
\correctimg[width=\linewidth,keepaspectratio]{{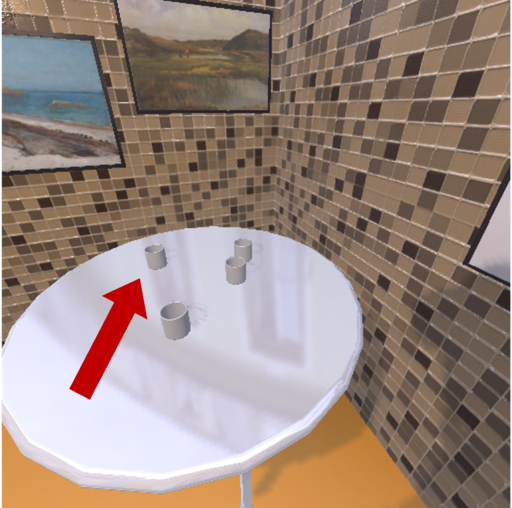}} \\

\midrule

\multicolumn{9}{c}{\footnotesize (State) ``\textit{Choose a \targetobj{\textbf{laptop}}'s state on the \receptacle{\textbf{desk}}. A: opened. B: closed.}''}\\[4pt]
\includegraphics[width=\linewidth,keepaspectratio]{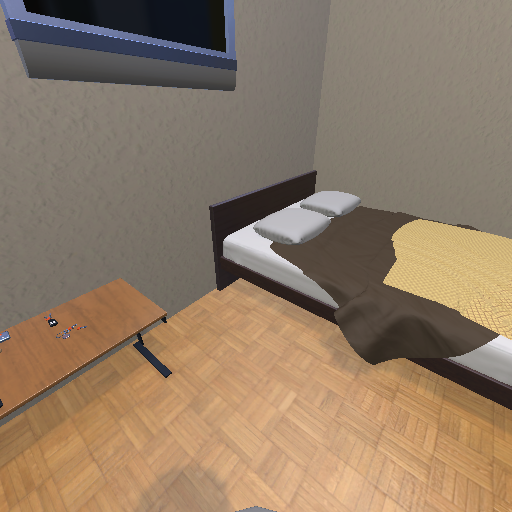} &
\includegraphics[width=\linewidth,keepaspectratio]{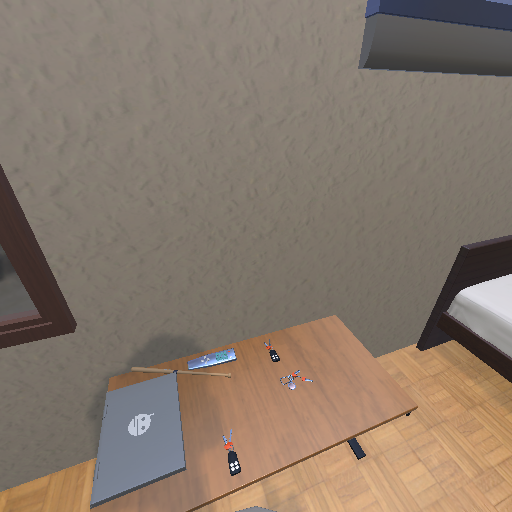} &
\wrongimg[width=\linewidth,keepaspectratio]{{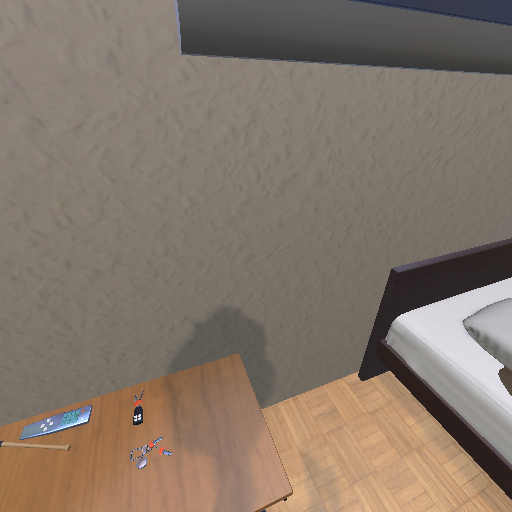}} &
\wrongimg[width=\linewidth,keepaspectratio]{{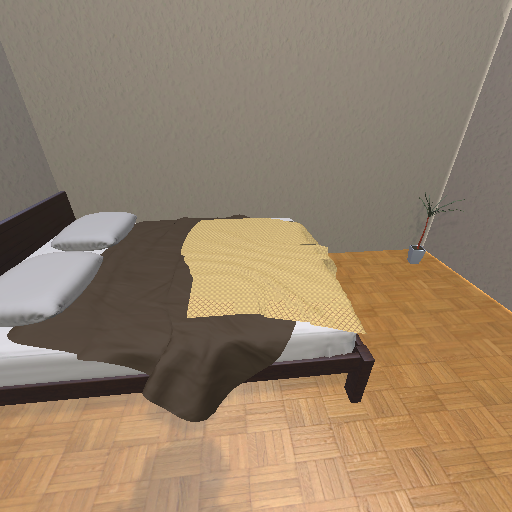}} &
\wrongimg[width=\linewidth,keepaspectratio]{{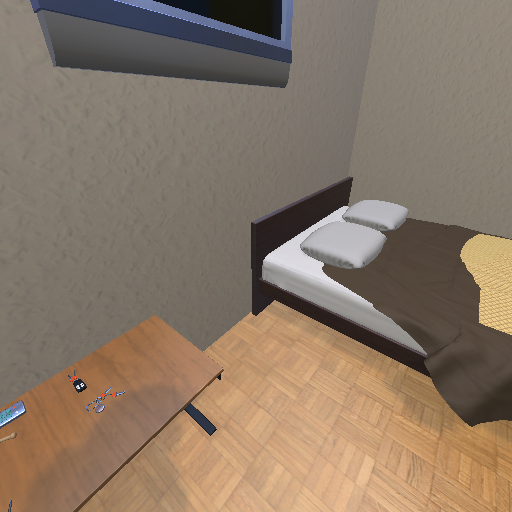}} &
\wrongimg[width=\linewidth,keepaspectratio]{{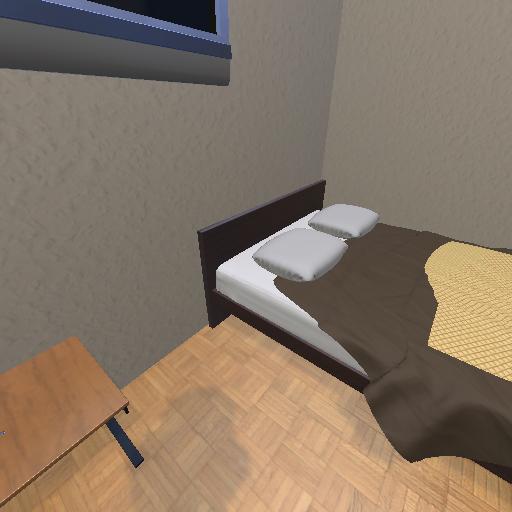}} &
\correctimg[width=\linewidth,keepaspectratio]{{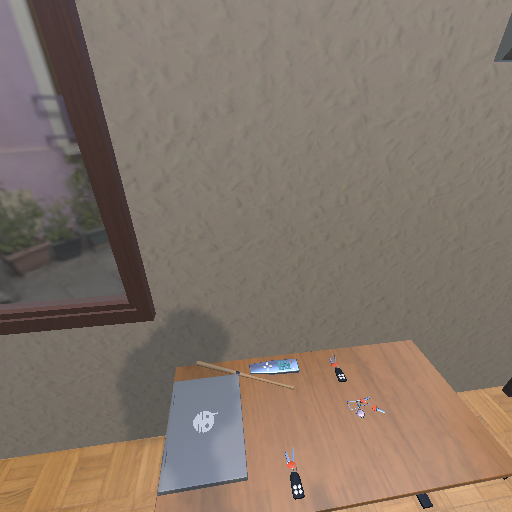}} &
\correctimg[width=\linewidth,keepaspectratio]{{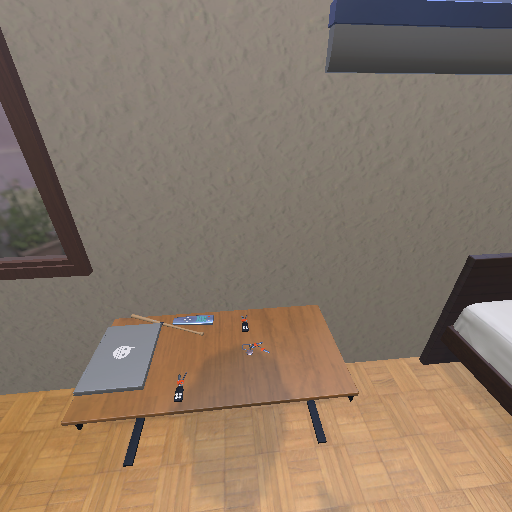}} & 
\correctimg[width=\linewidth,keepaspectratio]{{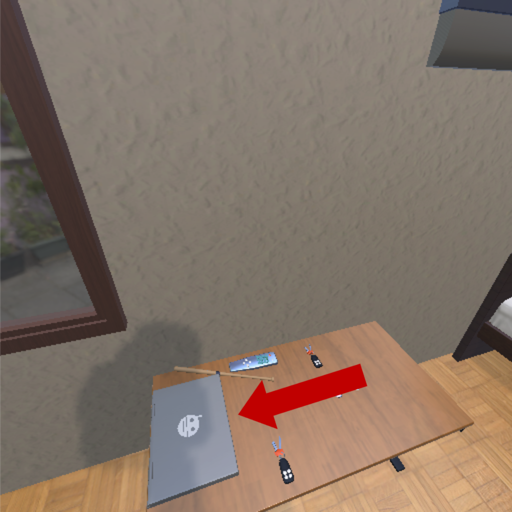}} \\

\midrule
%\multicolumn{9}{c}{\Benchmark{}-HM3D}
%\midrule
%\RowNine{figures/quali_cropped/hm3d/gpt_vilasr_0165}{(State) ``\textit{Is the \textbf{lamp} in the bedroom next to the window \textbf{turned on}?}''}

% ===== Category header row for second block (Gemini / SpatialReasoner) =====
% ===== Model-name header row variant (Gemini / SpatialReasoner) =====
%\hcell{Query View} & \hcell{Target View} & \hcell{Fine-EQA~\cite{Liu:2024EXPRESSBench}} & \hcell{Qwen-2.5-VL~\cite{Bai:2025Qwen2.5VL}} & \hcell{Gemini-2.5\\Pro~\cite{Google:2025Gemini25}} & \hcell{Spatial\\ Reasoner~\cite{Ma:2025SpatialReasoner}} & \hcell{SFT} & \hcell{RL} & \hcell{\textbf{Ours}} \\
%\midrule 
% ---- Rows (Gemini / SpatialReasoner set) ----
%\RowNine{figures/quali_cropped/procthor/counting_gemini_spatialreasoner_0478}{(Counting) ``\textit{How many \textbf{mugs} near the \textbf{diningtable}?}''}

\multicolumn{9}{c}{\footnotesize (Counting) ``\textit{How many \targetobj{\textbf{paintings}} are hanging on the wall near the \receptacle{\textbf{sofa}} in the living room?}''}\\[4pt]
\includegraphics[width=\linewidth,keepaspectratio]{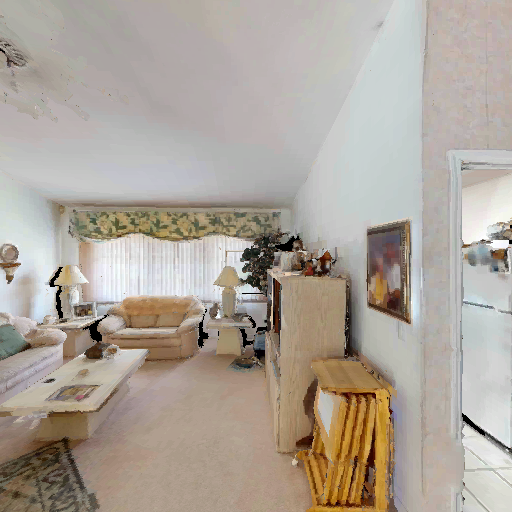} &
\includegraphics[width=\linewidth,keepaspectratio]{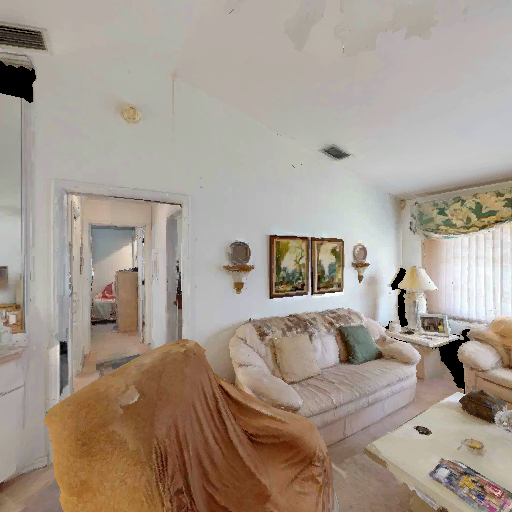} &
%\fcolorbox{blue}{white}
\wrongimg[width=\linewidth,keepaspectratio]{{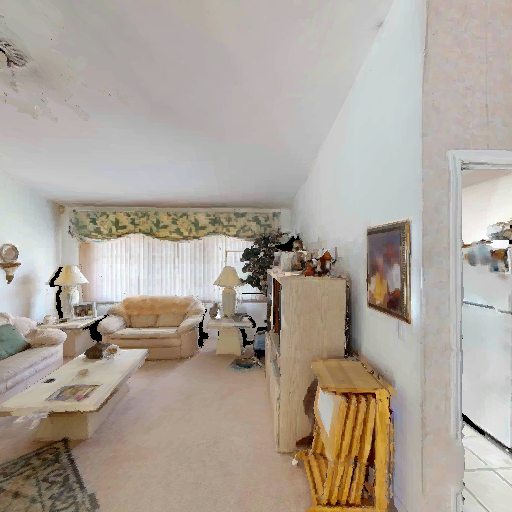}} &
\wrongimg[width=\linewidth,keepaspectratio]{{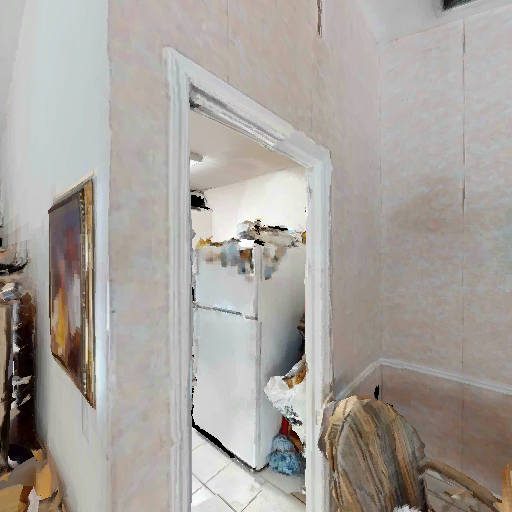}} &
\correctimg[width=\linewidth,keepaspectratio]{{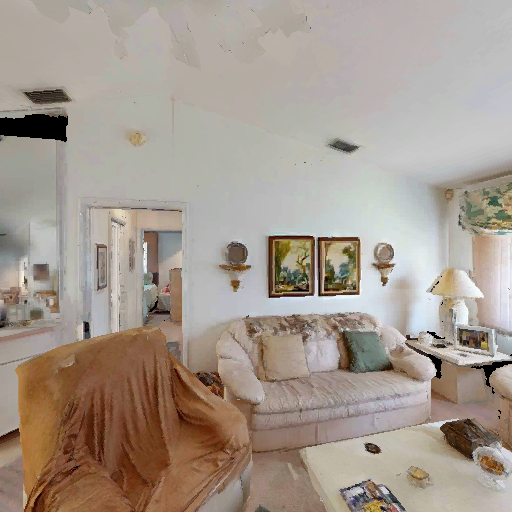}} &
\wrongimg[width=\linewidth,keepaspectratio]{{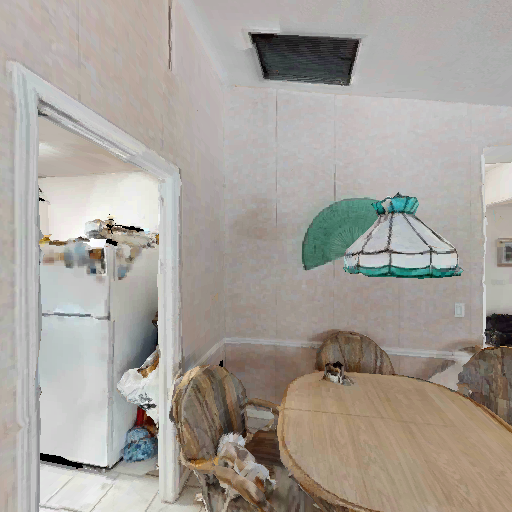}} &
\correctimg[width=\linewidth,keepaspectratio]{{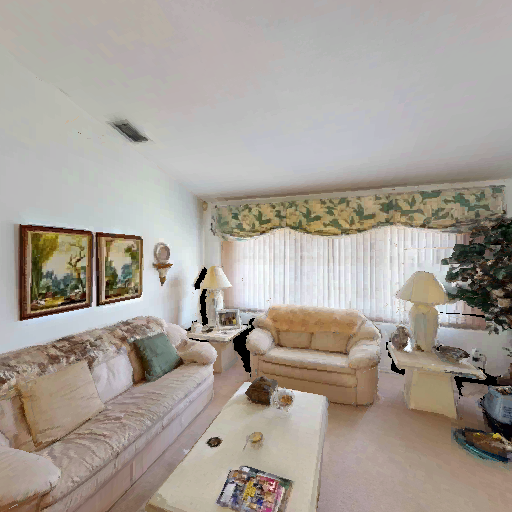}} &
\correctimg[width=\linewidth,keepaspectratio]{{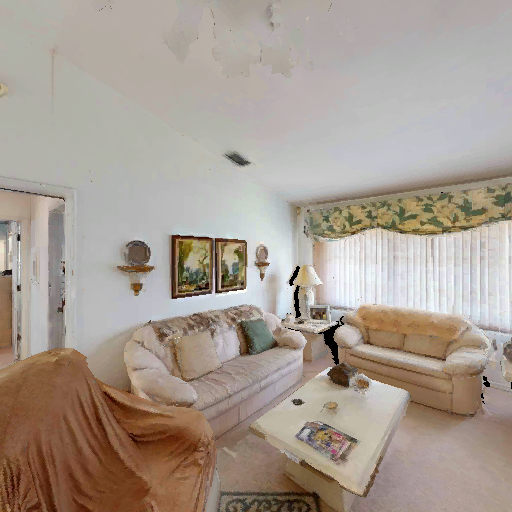}} &
\correctimg[width=\linewidth,keepaspectratio]{{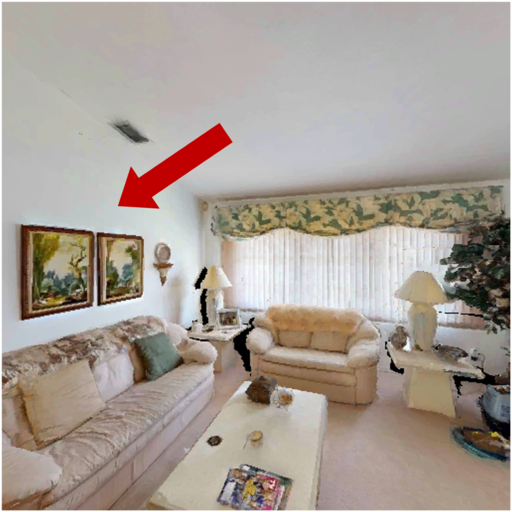}} \\

\midrule

\multicolumn{9}{c}{\footnotesize (Existence) ``\textit{Is there a \targetobj{\textbf{space}} for my winter coat in the \receptacle{\textbf{closet}} in the cloakroom?}''}\\[4pt]
\includegraphics[width=\linewidth,keepaspectratio]{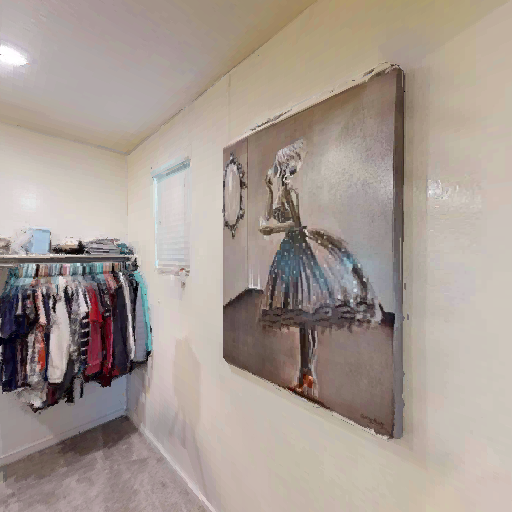} &
\includegraphics[width=\linewidth,keepaspectratio]{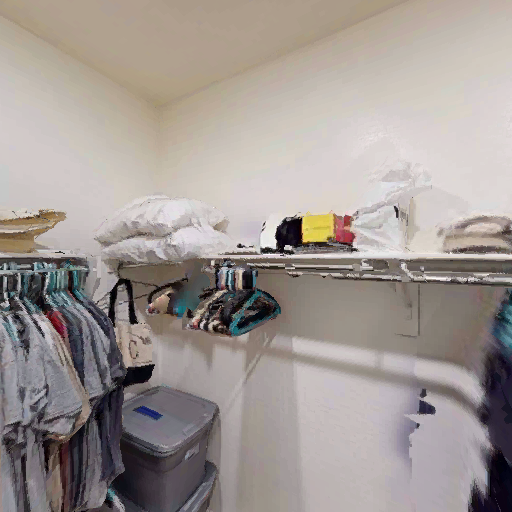} &
\correctimg[width=\linewidth,keepaspectratio]{{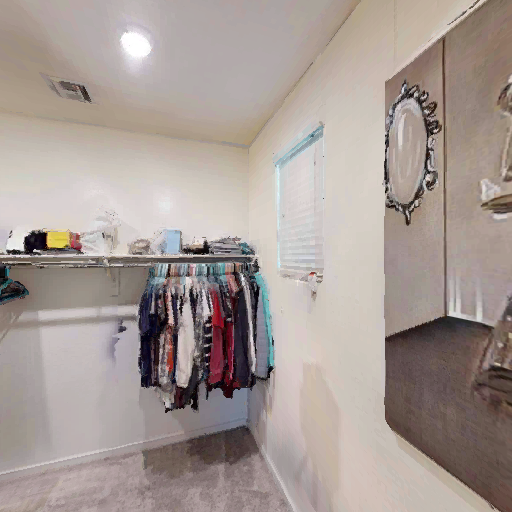}} &
\wrongimg[width=\linewidth,keepaspectratio]{{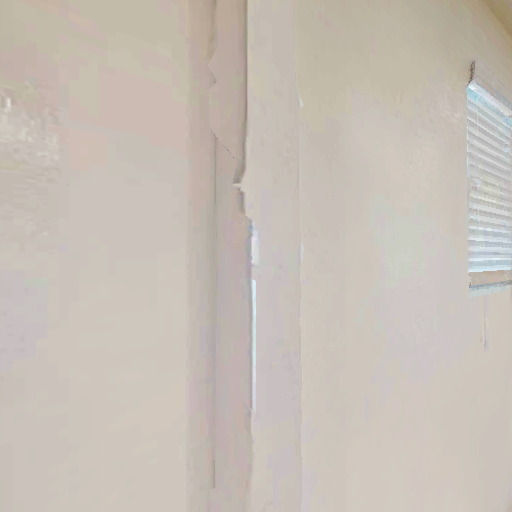}} &
\wrongimg[width=\linewidth,keepaspectratio]{{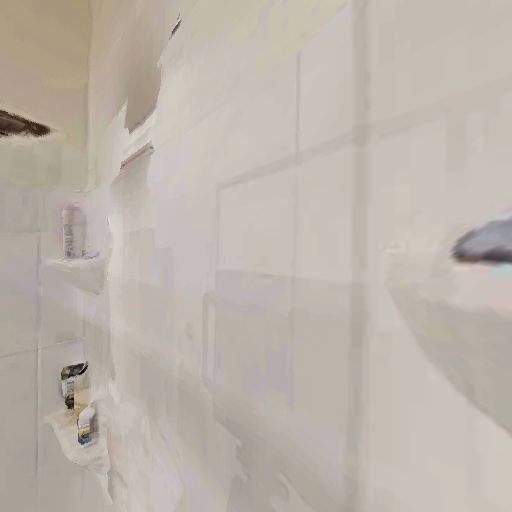}} &
\wrongimg[width=\linewidth,keepaspectratio]{{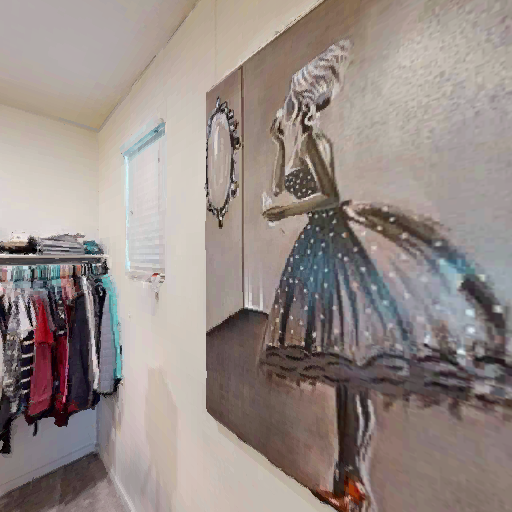}} &
\correctimg[width=\linewidth,keepaspectratio]{{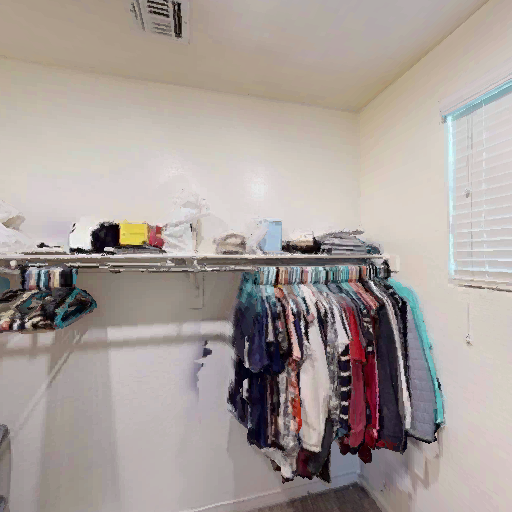}} &
\correctimg[width=\linewidth,keepaspectratio]{{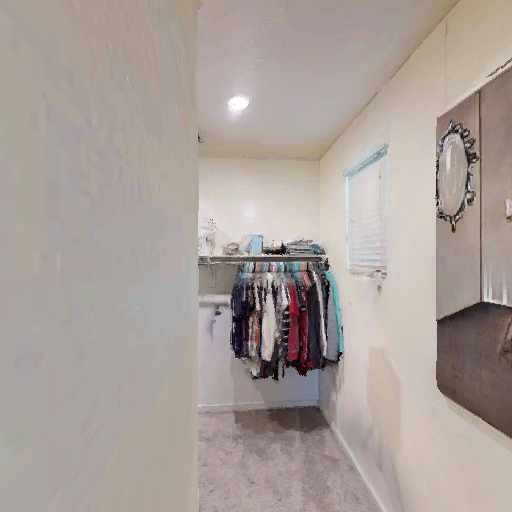}} &
\correctimg[width=\linewidth,keepaspectratio]{{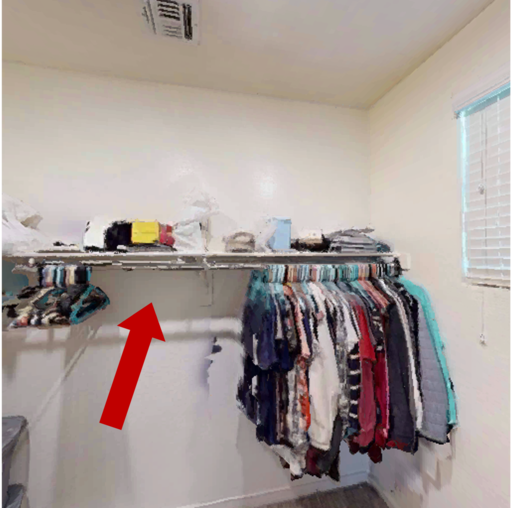}}
\\

% \multicolumn{9}{c}{\footnotesize (State) ``\textit{Did I hang up the \targetobj{\textbf{paintings}} in the \receptacle{\textbf{hallway}}?}''}\\[4pt]
% \includegraphics[width=\linewidth,keepaspectratio]{figures/quali_cropped/hm3d/gemini_vilasr_0201/input.png} &
% \includegraphics[width=\linewidth,keepaspectratio]{figures/quali_cropped/hm3d/gemini_vilasr_0201/gt.png} &
% \wrongimg[width=\linewidth,keepaspectratio]{{figures/quali_cropped/hm3d/gemini_vilasr_0201/eqa.png}} &
% \wrongimg[width=\linewidth,keepaspectratio]{{figures/quali_cropped/hm3d/gemini_vilasr_0201/qwen.png}} &
% \wrongimg[width=\linewidth,keepaspectratio]{{figures/quali_cropped/hm3d/gemini_vilasr_0201/pro.png}} &
% \wrongimg[width=\linewidth,keepaspectratio]{{figures/quali_cropped/hm3d/gemini_vilasr_0201/spatial.png}} &
% \correctimg[width=\linewidth,keepaspectratio]{{figures/quali_cropped/hm3d/gemini_vilasr_0201/sft.png}} &
% \correctimg[width=\linewidth,keepaspectratio]{{figures/quali_cropped/hm3d/gemini_vilasr_0201/grpo.png}} &
% \correctimg[width=\linewidth,keepaspectratio]{{figures/quali_cropped/hm3d/gemini_vilasr_0201/ours_marked.png}}\\
\midrule

\multicolumn{9}{c}{\footnotesize (State) ``\textit{Is the \targetobj{\textbf{lamp}} in the bedroom next to the \receptacle{\textbf{window}} turned on?}''}\\[4pt]
\includegraphics[width=\linewidth,keepaspectratio]{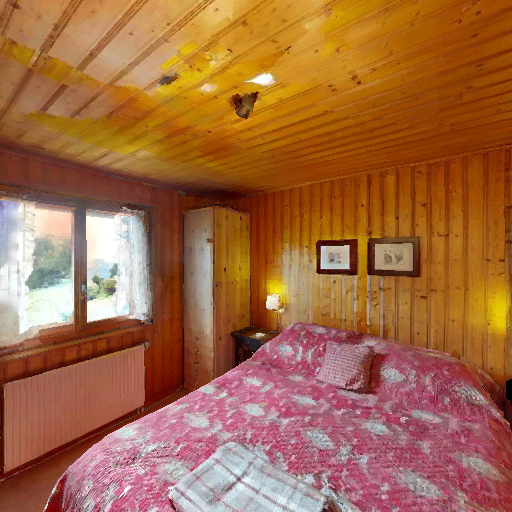} &
\includegraphics[width=\linewidth,keepaspectratio]{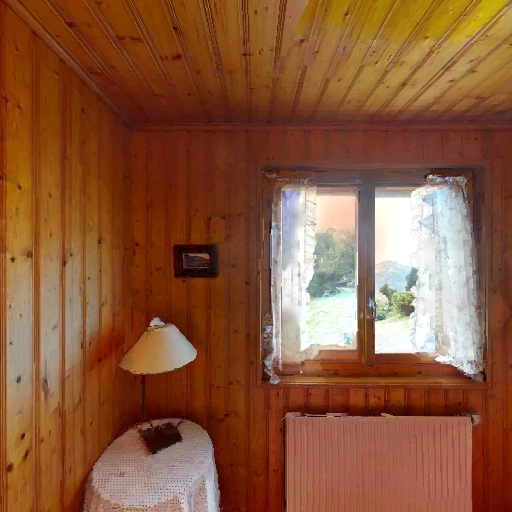} &
\wrongimg[width=\linewidth,keepaspectratio]{{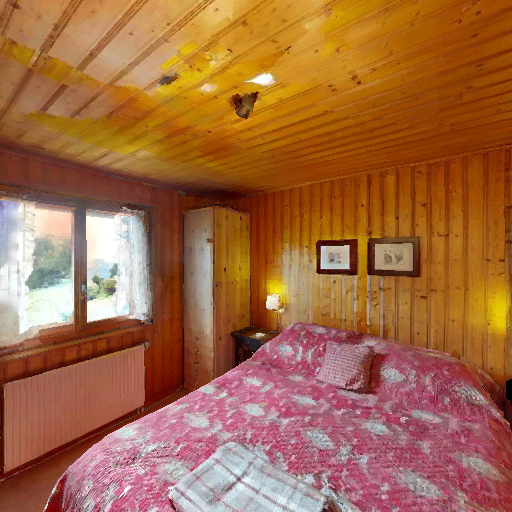}} &
\wrongimg[width=\linewidth,keepaspectratio]{{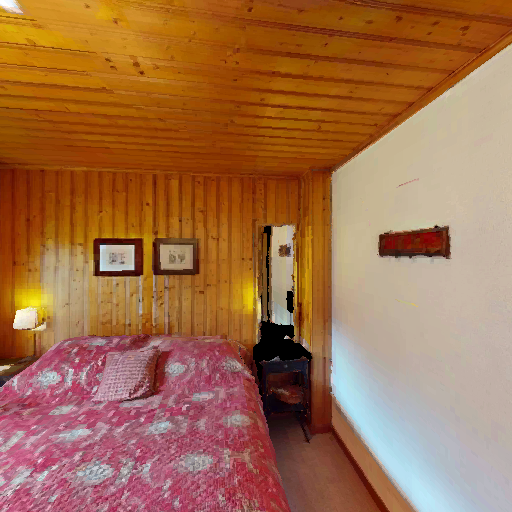}} &
\wrongimg[width=\linewidth,keepaspectratio]{{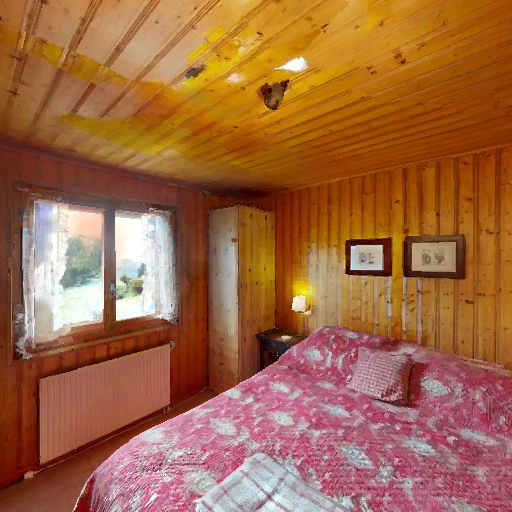}} &
\wrongimg[width=\linewidth,keepaspectratio]{{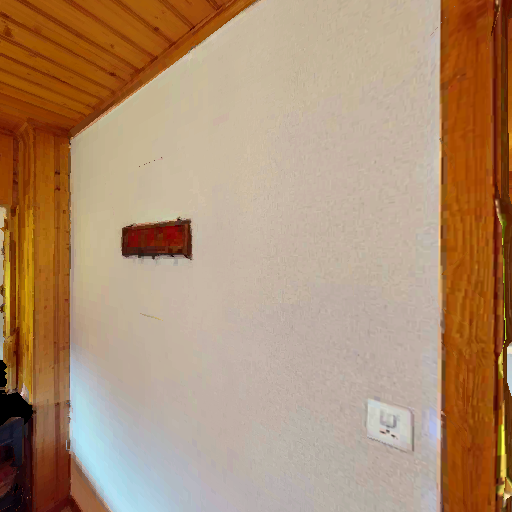}} &
\correctimg[width=\linewidth,keepaspectratio]{{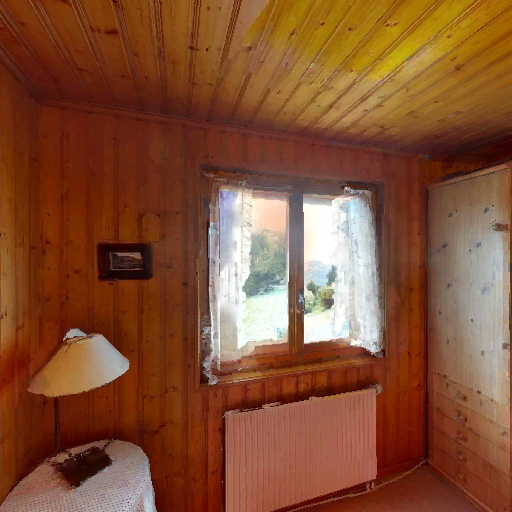}} &
\correctimg[width=\linewidth,keepaspectratio]{{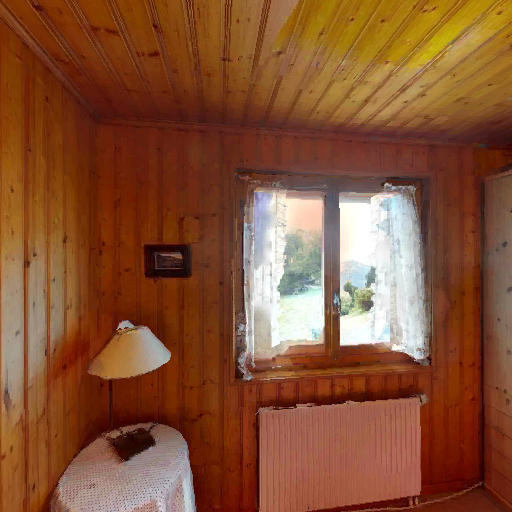}} &
\correctimg[width=\linewidth,keepaspectratio]{{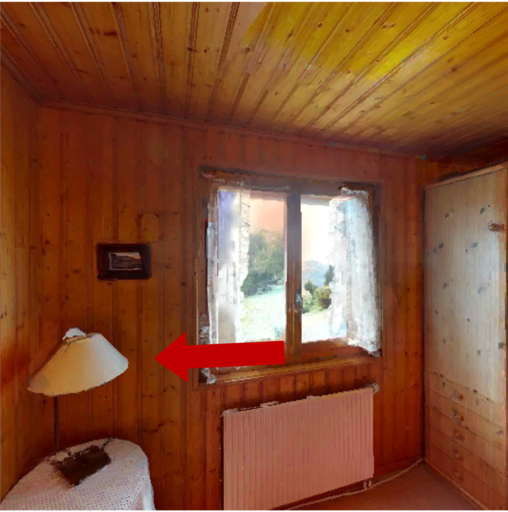}} \\
% \midrule
\bottomrule
\end{tabularx}
\vspace{-\baselineskip}
\caption{\textbf{Qualitative results in \Benchmark{}-ProcTHOR (top two rows) and \Benchmark{}-HM3D (bottom three rows).} \targetobj{Blue} denotes the object of interest, and \receptacle{gray} denotes surrounding objects that may serve as visual clues for reasoning. \correctmark corresponds to correct answers (or LLM-Match scores $=5$), \wrongmark corresponds to wrong answers (or LLM-Match scores $\leq 2$). The red arrow in the final column marks the region of interest.}
\label{fig:quali_benchmark}
\vspace{-1.5\baselineskip}
\end{figure*}

%%%%%%%%%%%%%%%%%%%%%%%%%%%%%%%%%%%%%%%%
% We evaluate diverse approaches on the proposed \Benchmark{} benchmark, which requires precise spatial reasoning and visually grounded action rather than reliance on world knowledge alone.
% The benchmark intentionally provides partial and ambiguous initial views, making it challenging to infer the most informative next viewpoint. \\

\paragraph{Experiment Setups.}
For evaluation on the Visually-Grounded Active View Selection (VG-AVS) task, we use two benchmark datasets: \Benchmark{}-ProcTHOR and \Benchmark{}-HM3D. \Benchmark{}-ProcTHOR is automatically generated following the data curation pipeline introduced in Section~\ref{sec:data_curation}, but includes more diverse question types than the training set, covering 516, 538, and 458 samples for \emph{existence}, \emph{counting}, and \emph{state}, respectively. 

\Benchmark{}-HM3D consists of real indoor scenes constructed from triplets of (question, answer, ground-truth view) in the validation split of Fine-EQA~\cite{Liu:2024EXPRESSBench}. Due to the absence of object visibility checks in real scenes, we first generate query views by randomly perturbing the ground-truth viewpoints so that the target object becomes invisible, and then manually collect 208 samples where the query view still provides sufficient contextual visual clues. \Benchmark{}-HM3D spans five question types: \emph{existence}, \emph{counting}, \emph{state}, \emph{attribute}, and \emph{object}. See the \Suppl{} for more details on the data construction.

% For evaluation on the Visually-Grounded Active View Selection (VG-AVS) task, we construct two benchmarks that assess both in-domain and real-world generalization, named \Benchmark{}-ProcTHOR and \Benchmark{}-HM3D.

% Beyond synthetic indoor scenes, \Benchmark{}-HM3D is comprised of real indoor environments in the Habitat-Matterport 3D (HM3D) dataset~\cite{Ramakrishnan:2021HM3D}. It is constructed from question–answer and ground-truth view pairs derived from the validation split of EXPRESS-Bench~\cite{Liu:2024EXPRESSBench}. Since automatic curation is infeasible in real scenes due to the lack of object-level visibility information, we first generate query views by randomly perturbing the ground-truth viewpoints so that the target objects become invisible. We then manually select samples where the query view provides sufficient contextual visual clues to infer the target view. The resulting benchmark contains 208 examples spanning \emph{attribute}, \emph{counting}, \emph{existence}, \emph{object}, and \emph{state} question categories. 

As evaluation metrics, we report VQA accuracy on \Benchmark{}-ProcTHOR, where all questions are in multiple-choice format so accuracy can be computed directly. For \Benchmark{}-HM3D, which consists of open-ended questions, we use LLM-Match~\cite{Majumdar:2024OpenEQA}, an LLM-based correctness metric where an LLM assigns a discrete score from 1 to 5 by comparing the predicted answer with the annotated ground-truth answer. We use Gemini-2.5-Flash~\cite{Google:2025Gemini25} for VQA accuracy and GPT-4~\cite{OpenAI:2023GPT4} for LLM-Match following prior work~\cite{Majumdar:2024OpenEQA}.

\vspace{-\baselineskip}
\paragraph{Quantitative Results.} 
We present a quantitative comparison in Table~\ref{tab:vgavs_main}. 
To provide loose reference upper and lower bounds that reflect the difficulty of the task, we report the VQA accuracy and LLM-Match scores when the VLM verifier is directly given the query view and the target view, respectively. For all action models, we instead first predict actions and feed the resulting predicted views to the VLM verifier to compute the metrics.

As shown, open-source models, including Qwen2.5-VL-7B~\cite{Bai:2025Qwen2.5VL} and recent spatial VLMs such as ViLaSR~\cite{Junfei:2025VILASR} and SpatialReasoner~\cite{Ma:2025SpatialReasoner}, all fail to achieve meaningful active view selection. In contrast, our AVS framework, which also uses Qwen2.5-VL-7B~\cite{Bai:2025Qwen2.5VL} as its backbone, consistently outperforms these models by large margins across different training strategies. This highlights that off-the-shelf open VLMs are insufficient to perform informative viewpoint changes from partial observations, and that training on our curated dataset is crucial for acquiring such active perception ability.

In our AVS framework, across different training strategies, all variants outperform other approaches, while the two-stage training scheme (SFT+RL) yields a further improvement of more than 5\% in VQA accuracy compared to either method alone. Moreover, they show strong generalizability to diverse question types that are not seen during training, including \emph{counting} and \emph{state} questions. Additional training strategy variants and ablations are provided in the \Suppl{}. 

Fine-EQA~\cite{Liu:2024EXPRESSBench}, an EQA framework proposed for a task closely related to ours, performs significantly worse on our benchmark. This indicates that EQA frameworks aiming for long-horizon navigation lack fine-grained viewpoint control, as their policies rely on coarse, discrete actions instead of continuous, precise adjustments.

When comparing against proprietary models, the advantage of our learned active perception system becomes even more pronounced. Despite using only a 7B-parameter backbone~\cite{Bai:2025Qwen2.5VL}, our model surpasses substantially larger proprietary models, including GPT-5~\cite{OpenAI:2025GPT5} and Gemini-2.5-Pro~\cite{Google:2025Gemini25}. This clearly highlights that active perception for ambulatory vision requires an explicit training procedure for precise view refinement, rather than relying solely on large-scale pretraining.

Beyond the synthetic scenes from ProcTHOR~\cite{Deitke:2022ProcTHOR}, the results on \Benchmark{}-HM3D reported in Table~\ref{tab:vgavs_main} clearly demonstrate that our AVS framework generalizes well to real indoor environments with diverse question types. Despite being trained only on a relatively small-scale synthetic 3D dataset with binary object existence questions, our approach outperforms all other baselines by clear margins, achieving 70.70 with SFT+RL compared to 64.91 with GPT-5~\cite{OpenAI:2025GPT5} in average score. This shows that a well-curated dataset that compactly supervises active view selection is sufficient to enable strong transfer to real-world scenes, even under shifts in both scene distribution and question format. Moreover, since our AVS module is orthogonal to the choice of backbone, these gains suggest that even stronger VLMs could further benefit from being equipped with our learned active perception system.

\vspace{-\baselineskip}
\paragraph{Qualitative Results.}
As shown in Figure~\ref{fig:quali_benchmark}, when the target object in the query view is only partially visible or appears too small, our AVS framework successfully refines the viewpoint to make the object fully observable and properly scaled. In the first row, for example, when the dining table is only partially visible in the query view, our framework moves the agent closer to the table so that all mug cups become fully observable. In the fourth row, we show a real indoor scene where the closet is only partially visible in the query view. Our method adjusts the viewpoint to fully reveal the closet, enabling the VLM verifier to answer correctly, demonstrating the generalizability of ours to real-world environments.

\subsection{Experiments on EQA}
\label{sec:eqa_results}
\vspace{-0.2\baselineskip}
\begin{table}[t!]
\centering
\setlength{\tabcolsep}{6pt}
\renewcommand{\arraystretch}{1.15}

\caption{\textbf{Comparison on Fine-EQA benchmark.}
Each column reports normalized LLM-Match score for different question types. Plugging our AVS framework into Fine-EQA improves performance. In each coolumn, best in \textbf{bold}, second best \underline{underlined}.}
\label{tab:existing_eqa}
\vspace{-0.5\baselineskip}
{\scriptsize
\begin{tabularx}{1.0\columnwidth}{l *{6}{Y} | Y}
\toprule
\textbf{Method} &
\textbf{Attr.} & \textbf{Count.} & \textbf{Exist.} &
\textbf{Obj.} & \textbf{State} & \textbf{Loc.} & \textbf{Avg.} \\
\midrule

Fine-EQA~\cite{Liu:2024EXPRESSBench} &
49.23 & 38.75 & 67.92 & \textbf{57.41} & \underline{63.20} & 41.14 & 52.94 \\
\midrule
w/ SFT &
\textbf{59.08} & \underline{45.00} & 67.29 & \underline{56.30} & 58.40 & \underline{45.71} & \underline{55.30} \\

w/ RL &
52.00 & 44.38 & \textbf{69.17} & 47.41 & 62.00 & 42.29 & 52.87 \\

\textbf{w/ Ours} &
\underline{53.23} & \textbf{52.50} & \underline{68.13} & 55.93 &
\textbf{65.40} & \textbf{50.86} & \textbf{57.67} \\
\bottomrule
\end{tabularx}}
\vspace{-\baselineskip}
\end{table}
% \begin{figure*}
% \dummyfig{1\linewidth}{0.3\linewidth}{Qualitative results in Active-EQA.}
% \caption{Qualitative results in Active-EQA.}    
% \end{figure*}
% Preamble (once)
% \usepackage{graphicx,booktabs,tabularx,array,xcolor}
% \newcolumntype{Y}{>{\centering\arraybackslash}X}
% \setlength{\tabcolsep}{1pt}

% ---- 9-column row macro ----
% Images expected at:
%  <base>/input.png, gt.png, eqa.png, qwen.png, pro.png, spatial.png, sft.png, grpo.png, ours.png

\renewcommand{\receptacle}[1]{%
  \begingroup
  \setlength{\fboxsep}{2pt} 
  \colorbox{gray!15}{#1}%
  \endgroup
}
\renewcommand{\targetobj}[1]{%
  \begingroup
  \setlength{\fboxsep}{2pt} 
  \colorbox{blue!15}{#1}%
  \endgroup
}

\newcommand{\wrongimg}[2][]{%
  \begin{overpic}[#1]{#2}
    \put(3,75){\color{wrongred}\fontsize{15}{15}\selectfont\bfseries\xmark}
  \end{overpic}
}

\newcommand{\correctimg}[2][]{%
  \begin{overpic}[#1]{#2}
    \put(3,75){\color{correctgreen}\fontsize{15}{15}\selectfont\bfseries\cmark}
  \end{overpic}
}

\newcommand{\RowFour}[2]{%
  \multicolumn{4}{c}{ #2}\\[4pt]
  \wrongimg[width=\linewidth,keepaspectratio]{#1/eqa.png} &
  \correctimg[width=\linewidth,keepaspectratio]{#1/sft.png}    &
  \correctimg[width=\linewidth,keepaspectratio]{#1/grpo.png}   &
  \correctimg[width=\linewidth,keepaspectratio]{#1/ours_marked.png}  \\
}

\begin{figure}[t]
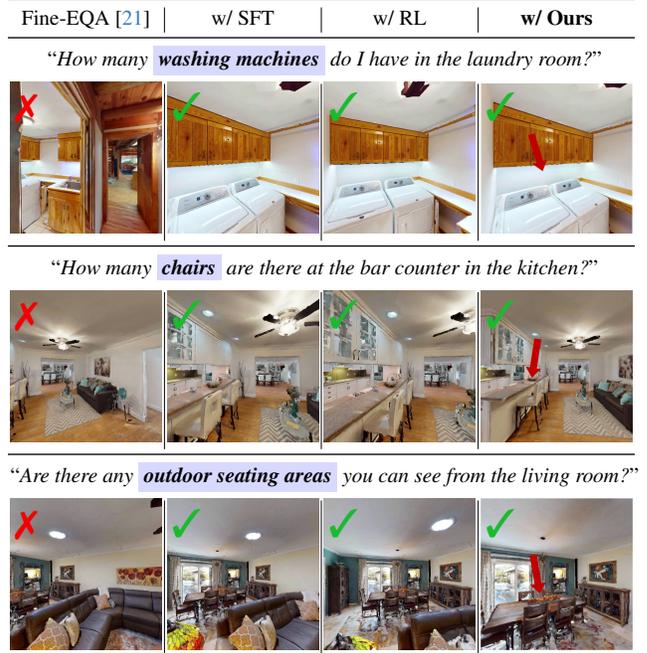

\footnotesize
\centering
\renewcommand{\arraystretch}{1.0}
\setlength{\tabcolsep}{1pt}
\arrayrulecolor{black}
\setlength{\arrayrulewidth}{0.2pt}
\begin{tabularx}{\columnwidth}{Y|Y|Y|Y}
\toprule
Fine-EQA~\cite{Liu:2024EXPRESSBench} & w/ SFT & w/ RL & \textbf{w/ Ours} \\
\midrule
\RowFour{figures/quali_cropped/eqa/sample_0096}{
``\textit{How many\targetobj{\textbf{washing machines}} do I have in the laundry room?}''}
\midrule
\RowFour{figures/quali_cropped/eqa/sample_0135}{``\textit{How many\targetobj{\textbf{chairs}} are there at the bar counter in the kitchen?}''}
\midrule
\RowFour{figures/quali_cropped/eqa/sample_0282}{``\textit{Are there any\targetobj{\textbf{outdoor seating areas}} you can see from the living room?}''}
\bottomrule
\end{tabularx}
%, booktabs, tabularx, array, xcolor 등)
% \newcolumntype{Y}{>{\centering\arraybackslash}X}
% \setlength{\tabcolsep}{1pt}
\vspace{-\baselineskip}
\caption{\textbf{Qualitative results in the Fine-EQA benchmark~\cite{Liu:2024EXPRESSBench}.} \targetobj{Blue} highlights denote object-of-interest. \correctmark corresponds to high scores ($=5$) and \wrongmark corresponds to low LLM-Match scores ($\leq2$). The red arrow in the final column marks the region of interest.}
\label{fig:quali_eqa}
\vspace{-1.5\baselineskip}
\end{figure}

Our AVS framework can naturally serve as a plug-and-play component to enhance existing EQA pipelines. Since EQA inherently requires multi-task abilities, prior work~\cite{Majumdar:2024OpenEQA, Liu:2024EXPRESSBench, Das:2018EQA, Ren:2024ExploreEQA} has largely underexplored the final step of fine-grained view refinement once the agent stops near the target, which is crucial for gathering the visual evidence needed to answer the question. Our AVS framework directly complements this stage: it takes the terminal view produced by the EQA policy, refines the viewpoint, and feeds the refined observation into the VLM verifier for answering.

As shown in Table~\ref{tab:existing_eqa}, this simple plug-in yields clear LLM-Match improvements over the base Fine-EQA~\cite{Liu:2024EXPRESSBench} framework, increasing the average score from 52.94 to 57.67 when combined with our AVS framework trained using SFT+RL, denoted as (w/ Ours) in the table.

% Beyond demonstrating generalization to real scenes in our active view selection task, this result also shows that our AVS module transfers well to EQA setups with real-world environments. This suggests that our method can serve as a complementary component that strengthens existing EQA frameworks.
Beyond demonstrating generalization to real scenes in VG-AVS task, this result also shows that our AVS module transfers well to EQA setups with real-world environments. This suggests that our method can serve as a complementary component that strengthens existing EQA frameworks.

As shown in Figure~\ref{fig:quali_eqa}, when the final view produced by the base EQA framework is insufficient for answering a question, our module successfully identifies the region of interest and adjusts the viewpoint to obtain a more informative observation. For example, in the first row, our model moves the agent inside the room to better capture the laundry room mentioned in the question. Comparing different training variants, in the second row, SFT and RL produce views where the left side of the bar counter is only partially visible, making the chair count less clear, whereas our two-stage training method moves to a viewpoint where the entire counter is clearly captured.

\vspace{-0.25\baselineskip}
\section{Conclusion}
\label{sec:conclusion}
\vspace{-0.25\baselineskip}
We advance toward ambulatory vision by reframing VQA as an active perception problem and introducing Visually Grounded Active View Selection (VG-AVS). By focusing on viewpoint selection from a single image and enabling VLM fine-tuning through a novel synthetic dataset and an SFT+RL fine-tuning framework, our approach achieves significant gains in view-selection-based question answering. It also generalizes well to unseen scenes and provides meaningful improvements when incorporated into existing scene-exploration-based EQA pipelines.

\clearpage
{
    \small
    \bibliographystyle{ieeenat_fullname}
    \bibliography{main}
}

\ifarxiv
    \clearpage
    \newpage
    \section*{Appendix}
    \label{sec:appendix}
    \renewcommand{\thesection}{A}
    \renewcommand{\thetable}{A\arabic{table}}
    \renewcommand{\thefigure}{A\arabic{figure}}
    \ifsuppl
    \makeatletter
    \newcommand{\manuallabel}[2]{\def\@currentlabel{#2}\label{#1}}
    \makeatother
    \manuallabel{sec:related_work}{2}
    \manuallabel{sec:method_overview}{3}
    \manuallabel{sec:data_curation}{3.1}
    \manuallabel{sec:problem_formulation}{3.2}
    \manuallabel{sec:sft}{3.2.1}
    \manuallabel{sec:rl}{3.2.2}
    \manuallabel{sec:two_stage_strategy}{3.2.3}
    \manuallabel{sec:experiment_results}{4}
    \manuallabel{sec:vgavs_results}{4.1}
    \manuallabel{sec:eqa_results}{4.2}
    \manuallabel{eq:o_tgt_sample}{1}
    \manuallabel{eq:o_qry_sample}{2}
    \manuallabel{tab:vgavs_main}{1}
    
    \newcommand{\refofpaper}[1]{of the main paper}
    \newcommand{\refinpaper}[1]{in the main paper}
\else
    \newcommand{\refofpaper}[1]{\unskip}
    \newcommand{\refinpaper}[1]{\unskip}
\fi

\ifsuppl
\noindent 
% In this supplementary material, we first present additional experiments following the experiments discussed~\refinpaper{}, including additional comparisons with additional different training strategies in Section~\ref{sec:supp_training_variants_results}, more EQA experiments with Open-EQA dataset in Section~\ref{sec:supp_open_eqa_results}, cross-dataset generalization in Section~\ref{sec:supp_cross_dataset_results}, additional experiemnt results with multi-turn actions in Section~\ref{sec:supp_multi_turn_results}. Next, we provide details on training our models and experiemnt setups in Section~\ref{sec:supp_implementation_details}. Then, we provide more details on our dataset curation in Section~\ref{sec:supp_data_implementation_details} and the input prompts we use in Section~\ref{sec:supp_prompts}. Lastly, we present more qualitative comparisons results in Section~\ref{sec:supp_more_qualitative_comparisons} and qualitative examples of our AVS framework in Section~\ref{sec:supp_qualitative_examples}.
In this supplementary material, we first present additional experiments that extend those discussed in Section~\ref{sec:experiment_results}~\refofpaper{}, including comparisons of different training strategies in Section~\ref{sec:supp_training_variants_results}, further EQA experiments on the Open-EQA~\cite{Majumdar:2024OpenEQA} dataset in Section~\ref{sec:supp_open_eqa_results}, cross-dataset generalization in Section~\ref{sec:supp_cross_dataset_results}, and additional results with multi-turn actions in Section~\ref{sec:supp_multi_turn_results}. 
Next, we provide details on model training and experimental setups in Section~\ref{sec:supp_implementation_details}. 
We then detail our dataset curation procedure in Section~\ref{sec:supp_data_implementation_details} and the input prompts used in Section~\ref{sec:supp_prompts}. 
Finally, we present additional qualitative comparison results in Section~\ref{sec:supp_more_qualitative_comparisons} and further qualitative examples of our AVS framework in Section~\ref{sec:supp_qualitative_examples}.
\else
\fi

\newcommand{\biggerwrongimg}[2][]{%
  \begin{overpic}[#1]{#2}
    \put(3,75){\color{wrongred}\fontsize{25}{25}\selectfont\bfseries\xmark}
  \end{overpic}
}

\newcommand{\biggercorrectimg}[2][]{%
  \begin{overpic}[#1]{#2}
    \put(3,75){\color{correctgreen}\fontsize{25}{25}\selectfont\bfseries\cmark}
  \end{overpic}
}

\begin{figure*}[t!]
\centering
\footnotesize
\rule{\textwidth}{0.8pt}\\[4pt]
{Question: \textit{``Is there a \targetobj{\textbf{painting}} on the right wall near the hallway?''}}\\
\rule{\textwidth}{0.2pt}\\
% 위아래 가는 선 대신 쓰고 싶으면 주석 해제
% \rule{\linewidth}{0.4pt}\\[2pt]

% ---------- 왼쪽: Exploration (1 row) ----------
\begin{minipage}[t]{0.6\linewidth}
  \centering
  Exploration\\
  \includegraphics[width=\linewidth]{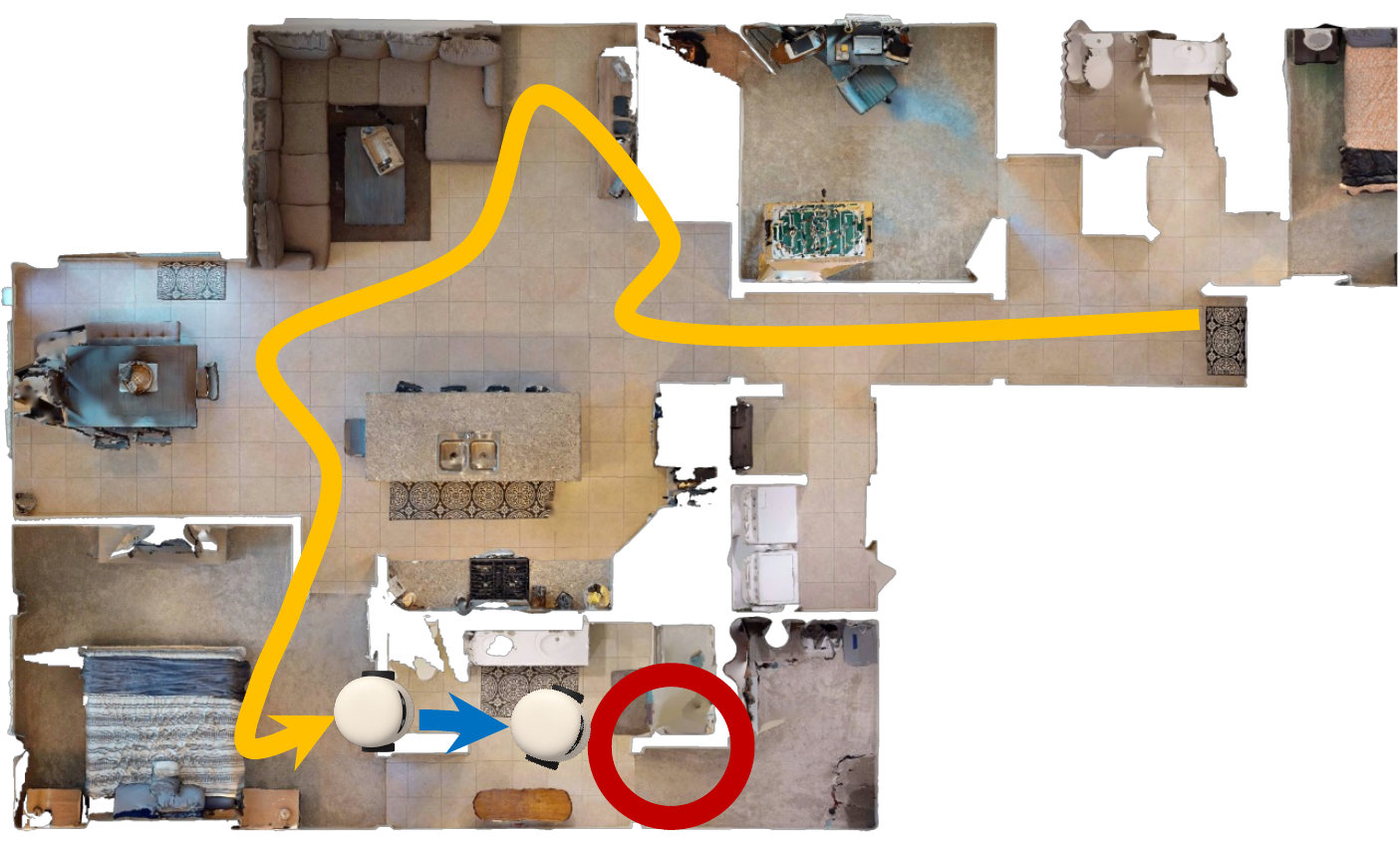}
\end{minipage}%
% ---------- 오른쪽: 2x2 (Fine-EQA, SFT, RL, Ours) ----------
\begin{minipage}[t]{0.32\linewidth}
  \centering
  \begin{tabular}[t]{cc}
    \footnotesize Fine-EQA~\cite{Liu:2024EXPRESSBench} &
    \footnotesize w/ SFT\\[3pt]
    \biggerwrongimg[width=0.48\linewidth]{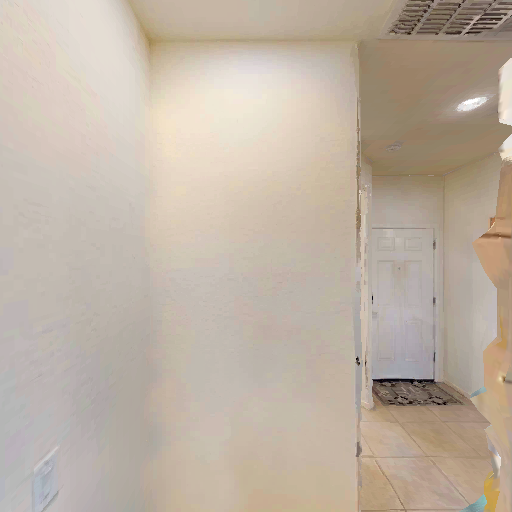} &
    \biggercorrectimg[width=0.48\linewidth]{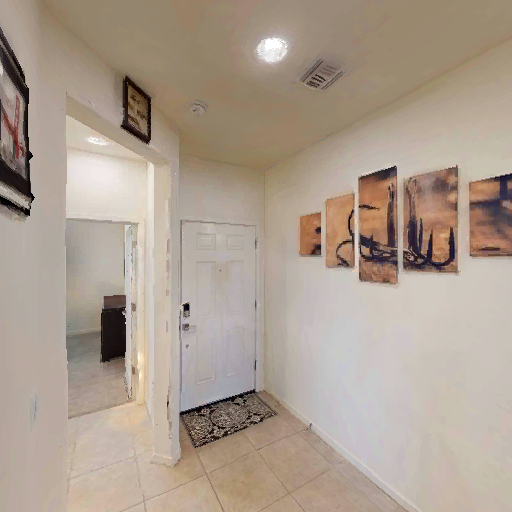} \\[3pt]
    \midrule
    \footnotesize w/ RL &
    \footnotesize \textbf{w/ SFT + RL (Ours)} \\[3pt]
    \biggercorrectimg[width=0.48\linewidth]{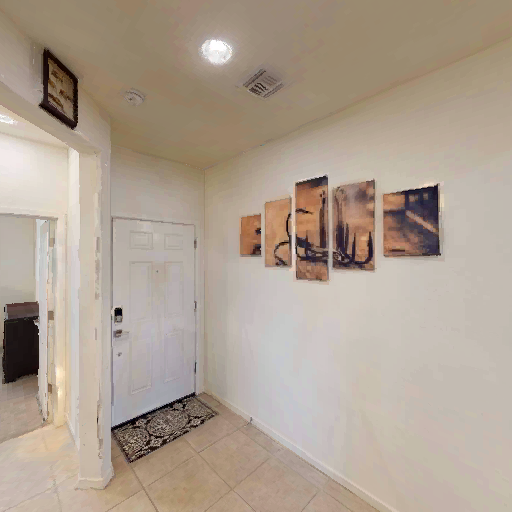} &
    \biggercorrectimg[width=0.48\linewidth]{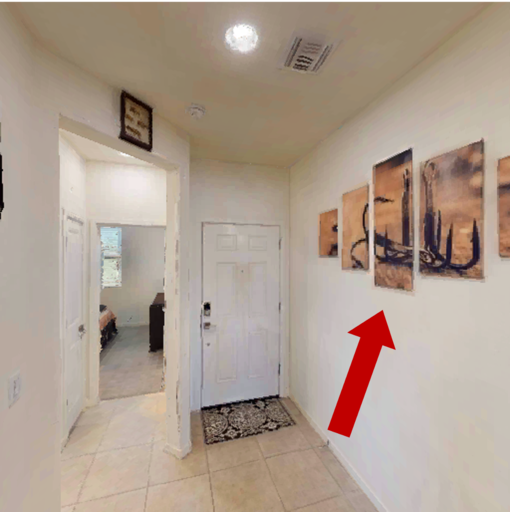} \\
  \end{tabular}
\end{minipage}
% \bottomrule
\\[4pt]
\rule{\textwidth}{0.8pt}
\caption{\textbf{Illustration of integrating our AVS framework into an EQA pipeline.} On the left, the yellow line represents the exploration path of the EQA pipeline, the blue arrow the final viewpoint refinement by our model, and the red circle the region of interest given the question. On the right, we show the views obtained by each method, where \correctmark denotes a correct view with a high LLM-Match score ($=5$) and \wrongmark an incorrect view with a low score ($\leq 2$). Given the question about the \targetobj{\emph{painting}}, our AVS framework effectively refines the viewpoint such that the final predicted view reveals the painting on the wall. This demonstrates that the training-free EQA pipeline lacks query-conditioned final viewpoint refinement, whereas our learning-based method can minimally adjust the viewpoint to obtain an answerable view.}
\label{fig:quali_eqa_concept}
\end{figure*}

\subsection{Extended Experiments}
\subsubsection{Comparison with More Training Variants}
\label{sec:supp_training_variants_results}
% \definecolor{opensourcecolor}{HTML}{0072B2} 
% \definecolor{spatialcolor}{HTML}{E69F00}   
% \definecolor{eqacolor}{HTML}{009E73}      
% \definecolor{procolor}{HTML}{D55E00}     
% \definecolor{ourscolor}{HTML}{cbdfbd}     

\begin{table*}[t!]
    \centering
    \footnotesize
    \caption{\textbf{Quantitative comparison on internal baselines.} We report VQA accuracy on \Benchmark{}-ProcTHOR and LLM-Match scores on \Benchmark{}-HM3D, normalized to a percentage scale. The best in each column is in \textbf{bold} and second best is \underline{underlined}.}
    \label{tab:supp_internal_comparison}
    \setlength{\tabcolsep}{2pt}
    % Define a light gray color for highlighting 'Ours'
    \definecolor{Gray}{gray}{0.9}
    % Setup for siunitx columns, assuming scores are like XX.X
    % \sisetup{detect-weight, mode=text}
    \begin{tabularx}{\linewidth}{
        >{\raggedright\arraybackslash}p{3.0cm} |
        Y Y Y | Y |  % ProcTHOR: Ex, Cnt, State | Average
        Y Y Y Y Y | Y  % HM3D: Attr, Ex, Cnt, Obj, State | Average
    }
    
    \toprule
    \multirow{2}{*}{\textbf{AVS Framework}}  & \multicolumn{4}{c|}{\textbf{AVS-ProcTHOR}} & \multicolumn{6}{c}{\textbf{AVS-HM3D}} \\
    \cmidrule(lr){2-5}\cmidrule(lr){6-11}
    & \scriptsize{\textbf{Existence}} & \scriptsize{\textbf{Counting}} & \scriptsize{\textbf{State}} & \scriptsize{\textbf{Average}} 
    & \scriptsize{\textbf{Existence}} & \scriptsize{\textbf{Counting}} & \scriptsize{\textbf{State}} & \scriptsize{\textbf{Attribute}} & \scriptsize{\textbf{Object}} & \scriptsize{\textbf{Average}} \\
    \midrule
    SFT               & \underline{91.28} & 57.06 & 83.84 & 77.39 & 67.50 & \underline{70.77} & 62.16 & 66.67 & 55.56 & 64.53 \\
    SFT (Extended) & \underline{91.28} & 61.52 & \underline{85.59} & \underline{79.46} & 76.67 & 67.69 & 68.92 & 73.33 & 55.56 & 68.43 \\
    SFT w/ NT Loss~\cite{Zausinger:2025NTLoss}           & 81.59 & 60.41 & 77.51 & 73.17 & 77.08 &  70.00  & 63.24 & 60.83 & \textbf{63.89} & 67.01 \\
    RL             & 86.82 & \underline{65.24} & 83.41 & 78.49 & \textbf{81.25} & 70.00 & \underline{72.97} & \underline{69.17} & 60.00  & \underline{70.68} \\
    RL w/ $r^\text{pose}$ & 86.05 & 61.52 & 83.41 & 76.99 & \underline{78.33} & 66.92 & 72.43 & \textbf{70.83} & 58.89 & 69.48 \\
    \textbf{SFT+RL (Ours)}              & \textbf{91.47} & \textbf{69.52} & \textbf{90.17} & \textbf{83.72} & 74.58 & \textbf{71.54} & \textbf{73.78} & \textbf{70.83} & \underline{62.78} & \textbf{70.70} \\
    \bottomrule
    \end{tabularx}
    \vspace{-\baselineskip}
\end{table*}

As discussed in Section~\ref{sec:vgavs_results}~\refofpaper{}, we further evaluate three internal variants for a comprehensive comparison of training strategies. The quantitative results are summarized in Table~\ref{tab:supp_internal_comparison}. Below, we detail each variant and its training setup:
\begin{itemize}
    \item \textbf{SFT w/ NT Loss}~\cite{Zausinger:2025NTLoss}:
    this variant is trained with an auxiliary regression-aware loss, called Number Token Loss (NT Loss)~\cite{Zausinger:2025NTLoss}, on action magnitudes
    instead of a cross-entropy loss. In our setting, this change brings no benefit over vanilla SFT training.
    
    \item \textbf{SFT (extended)}:
    to fairly compare against our two-stage training, we extend SFT training until convergence (from 7 to 20 epochs) so that the only difference lies in the training objective.
    While longer SFT training yields modest gains, SFT alone still underperforms
    our two-stage training (SFT+RL).

    \item \textbf{RL w/} $r^{\text{pose}}$:
    this variant incorporates an additional distance-based reward inversely proportional
    to the discrepancy between the predicted and target camera poses
    (position and orientation), thereby injecting weak supervision of the
    ground-truth action into RL. Nevertheless, it still fails to learn
    meaningful actions without the SFT warm-up stage, highlighting that
    a supervised initialization followed by RL is more effective than
    relying on more sophisticated reward designs for RL alone.
\end{itemize}

\subsubsection{More Experiments on EQA}
\label{sec:supp_open_eqa_results}
\begin{table*}[t!]
\centering
\setlength{\tabcolsep}{6pt}
\renewcommand{\arraystretch}{1.15}

\caption{\textbf{Comparison on Open-EQA benchmark.}
Each column reports normalized LLM-Match scores for different question types. 
Plugging our AVS framework into an exiting EQA pipeline~\cite{Liu:2024EXPRESSBench} improves performance. 
Best in \textbf{bold}, second best \underline{underlined}.}
\label{tab:supp_open_eqa}

\vspace{-0.5\baselineskip}
{\scriptsize
\begin{tabularx}{\textwidth}{l *{6}{Y} | Y}
\toprule
\textbf{Method} &
\textbf{Object Recognition} & \textbf{Spatial Understanding} & \textbf{Object State Recognition} &
\textbf{Attribute Recognition} & \textbf{Object Localization} & \textbf{Functional Reasoning} & \textbf{Average} \\
\midrule

Fine-EQA~\cite{Liu:2024EXPRESSBench} &
40.00 & 45.26 & 51.11 & \underline{52.73} & \textbf{41.71} & \textbf{48.24} & \underline{46.51} \\
\midrule

w/ SFT &
\underline{44.80} & 42.11 & \underline{60.00} & 43.64 & 35.43 & 45.88 & 45.31 \\

w/ RL &
41.60 & \underline{46.32} & 55.56 & 43.03 & 35.43 & \textbf{48.24} & 45.03 \\

\textbf{w/ SFT+RL (Ours)} &
\textbf{52.80} & \textbf{65.20} & \textbf{61.40} & \textbf{58.20} & \underline{36.60} & \underline{46.20} & \textbf{53.40} \\
\bottomrule
\end{tabularx}}
\end{table*}

We report additional results on EQA setups using the Open-EQA~\cite{Majumdar:2024OpenEQA} dataset in Table~\ref{tab:supp_open_eqa}. We use the same Fine-EQA~\cite{Liu:2024EXPRESSBench} configuration as the base EQA pipeline as in Section~\ref{sec:eqa_results} \refofpaper{}, and then incorporate our AVS framework on top. As shown, plugging our AVS framework into this pipeline improves performance, yielding a 6.8\%pt gain in average LLM-Match score compared to the Fine-EQA baseline.

We also illustrate the integration of our AVS framework into an existing EQA pipeline in Figure~\ref{fig:quali_eqa_concept}.

\subsubsection{Cross-Dataset Generalization}
\label{sec:supp_cross_dataset_results}
% \begin{table}[t]
% \centering
% \caption{\textbf{Results on SAT benchmark.}
%   We report performance both for synthetic and real splits.
%   Best is in \textbf{bold}, second best is \underline{underlined}.}
% \label{tab:spatial_reasoning}
% \definecolor{Gray}{gray}{0.9}
% % \sisetup{detect-weight, mode=text}

% \setlength{\tabcolsep}{4pt}
% \scriptsize

% \begin{tabularx}{\linewidth}{
%     p{1.3cm} | >{\centering\arraybackslash}p{2.1cm} | Y Y  >{\centering\arraybackslash}p{1.8cm}
% }
% \toprule
%  \multirow{2}{*}{\textbf{Scene Type}} & \textbf{Backbone} & \multicolumn{3}{c}{\textbf{AVS Framework}}\\
%  \cmidrule(lr){2-5}
%  % \cline{2-5} 
%  & {\textbf{Qwen2.5-VL-7B}~\cite{Bai:2025Qwen2.5VL}} &
%    {\textbf{SFT}} &
%    {\textbf{RL}} &
%    {\textbf{SFT + RL (Ours)}} \\
% \midrule
% \textbf{Synthetic} & \underline{59.11} & \textbf{69.33} & 57.31 & \textbf{69.33} \\
% \textbf{Real  } & 60.00 &  \underline{67.33} & 57.33 & \textbf{77.33} \\
% \bottomrule
% \end{tabularx}
% \end{table}

% ====================================================
% =========== TWO TABLES with MINIPAGE ===============
% ====================================================

\begin{table*}
\centering
\scriptsize
% ---------- Left table (about 60%) ----------
    \begin{minipage}[t]{0.49\textwidth}
        \captionof{table}{\textbf{Results on SAT benchmark.}
        We report performance both for synthetic and real splits.
        For each row, the best is in \textbf{bold} and the second best is \underline{underlined}.}
        \label{tab:spatial_reasoning}
        \setlength{\tabcolsep}{4pt}
        \begin{tabularx}{\linewidth}{
                p{1.3cm} | >{\centering\arraybackslash}p{2.1cm} | Y Y  >{\centering\arraybackslash}p{1.8cm}
            }
            \toprule
             \multirow{2}{*}{\textbf{Scene Type}} & \textbf{Backbone} & \multicolumn{3}{c}{\textbf{AVS Framework}}\\
             \cmidrule(lr){2-5}
             % \cline{2-5} 
             & {\textbf{Qwen2.5-VL-7B}~\cite{Bai:2025Qwen2.5VL}} &
               {\textbf{SFT}} &
               {\textbf{RL}} &
               {\textbf{SFT + RL (Ours)}} \\
            \midrule
            \textbf{Synthetic} & \underline{59.11} & \textbf{69.33} & 57.31 & \textbf{69.33} \\
            \textbf{Real  } & 60.00 &  \underline{67.33} & 57.33 & \textbf{77.33} \\
            \bottomrule
        \end{tabularx}
    \end{minipage}
\hfill
% ---------- Right table (about 40%) ----------
    \begin{minipage}[t]{0.49\textwidth}
        \captionof{table}{\textbf{Quantitative comparison with multi-turn actions.} We report VQA accuracy on \Benchmark{}-ProcTHOR. The best in each column is in \textbf{bold} and second best is \underline{underlined}.}
        \label{tab:supp_multiturn}
        \setlength{\tabcolsep}{2pt}
        \begin{tabularx}{\linewidth}{
            >{\raggedright\arraybackslash}p{1.4cm} | >{\centering\arraybackslash}p{1.2cm} |
            Y Y Y | Y  % ProcTHOR: Ex, Cnt, State | Average
        }
            \toprule
            \multirow{2}{*}{{\makecell{\textbf{Training}\\\textbf{Variants}}}} & \multirow{2}{*}{{\makecell{\textbf{Action}\\\textbf{Steps}}}} & \multicolumn{4}{c}{\textbf{AVS-ProcTHOR}} \\
            \cmidrule(lr){3-6}
            & & \scriptsize{\textbf{Existence}} & \scriptsize{\textbf{Counting}} & \scriptsize{\textbf{State}} & \scriptsize{\textbf{Average}} \\
            \midrule
            \multirow{2}{*}{\makecell{SFT+RL\\(Q-only)}} & 1 & \textbf{91.47} & \textbf{69.52} & \underline{90.17} & \textbf{83.72} \\
                                    & 2 & 87.60          & 61.52          & 87.55            & 78.89          \\
            % \midrule
            % \multirow{2}{*}{RL}     & 1 & 86.82          & 65.24          & 83.41            & 78.49          \\
                                    % & 2 & 80.04          & 43.31          & 82.75            & 68.70          \\
            \midrule
            \multirow{2}{*}{\makecell{SFT+RL\\(Q+T)}} & 1 & \underline{90.50} & \underline{68.22} & \textbf{90.39}   & \underline{83.04} \\
                                                          & 2 & \underline{90.50} & \underline{68.22} & \underline{90.17} & 82.96         \\
            \bottomrule
        \end{tabularx}    
    \end{minipage}
\end{table*}
To assess whether training on our relatively small-scale dataset harms the general spatial understanding of the original VLM, Qwen2.5-VL-7B~\cite{Bai:2025Qwen2.5VL}, we additionally evaluate our models on the SAT~\cite{Ray:2024SAT} spatial reasoning benchmark, which comprises several downstream tasks across synthetic and real-image splits. As shown in Table~\ref{tab:spatial_reasoning}, although our framework is not explicitly trained for this benchmark, it generally improves the backbone’s performance, with the RL-only training strategy being the only variant that slightly underperforms. These results indicate that training on our synthetic dataset does not lead to noticeable catastrophic forgetting and can transfer reasonably well to broader spatial reasoning tasks.

% learning visually grounded actions enhances spatial reasoning and metric awareness, leading to gains even on tasks beyond our training objective.

\subsubsection{Results with Multi-Turn Actions}
\label{sec:supp_multi_turn_results}
We provide additional results with multi-turn action sequences in Table~\ref{tab:supp_multiturn} and observe that executing multiple actions does not yield clear performance gains over the single-step setting.

We conduct these multi-turn experiments with two models. The first is our main model used in Section~\ref{sec:experiment_results}~\refofpaper{}, which is trained with query views as input only and denoted as SFT+RL (Q-only). We also train a variant that takes both the query and target views as input, denoted as SFT+RL (Q+T). This SFT+RL (Q+T) model attains accuracy comparable to our original SFT+RL (Q-only) model and exhibits more stable performance between single- and multi-turn rollouts, yet its performance likewise does not improve in the multi-turn setting.

% When we naively roll out this policy for two steps, the average accuracy slightly drops compared to the single-step setting.

% We also train a variant that takes both the query and target views as input, denoted as SFT+RL (Q+T). 

Overall, these results indicate that our continuous action design enables the agent to reach an informative viewpoint within a single turn.

\subsection{Training and Experiment Setup Details}
\label{sec:supp_implementation_details}
\paragraph{Training.} We use Qwen2.5-VL-7B~\cite{Bai:2025Qwen2.5VL} as the backbone VLM, keeping the vision encoder frozen during training.
We train the model with a batch size of 32 for SFT and 128 for RL, running 7 epochs of SFT followed by 5 additional epochs of GRPO-based reinforcement learning.
Following prior work~\cite{Zausinger:2025NTLoss}, we set the weight of the Number Token Loss to 0.3 for the SFT model trained with NT Loss (see Table~\ref{tab:supp_internal_comparison}).
The total reward in GRPO is a weighted sum of a format reward and a verifier reward, with weights 0.3 and 1.0, respectively. We set the KL penalty coefficient $\beta$ to 0.04.
During GRPO training, we use a separate, frozen Qwen2.5-VL-7B~\cite{Bai:2025Qwen2.5VL} as the VLM verifier for computing verifiable reward $r^{\text{ver}}$, with a group size of 16. We use learning rates of $2\times10^{-5}$ for SFT and $10^{-6}$ for GRPO.

\paragraph{EQA Baseline.}
In Table~\ref{tab:vgavs_main}~\refofpaper{}, for a comparison with an EQA framework, we employ the state-of-the-art Fine-EQA approach~\cite{Liu:2024EXPRESSBench}. This framework includes both a frontier-based exploration strategy, which simply aims to visit unseen regions, and a goal-oriented exploration (GOE) strategy, which guides the agent toward semantically informative areas given the current observation and language query. We adopt the GOE strategy so that the framework can actively adjust its viewpoint based on contextual cues, aligning it with our Visually-Grounded Active View Selection (VG-AVS) task. Following Fine-EQA~\cite{Liu:2024EXPRESSBench}, we use \texttt{prism-dinosiglip+7b} introduced in Prismatic VLMs~\cite{Karamcheti:2024Prismatic} as the VLM backbone and replace the older GPT-4 models with the GPT-5 family~\cite{OpenAI:2025GPT5} for region prioritization and exploration termination decisions.

\subsection{More Details on Dataset Curation}
\label{sec:supp_data_implementation_details}
We provide additional details on the data curation procedure briefly introduced in Sections~\ref{sec:data_curation} and~\ref{sec:vgavs_results}~\refofpaper{}. We first describe the AVS dataset and AVS-ProcTHOR, both curated with ProcTHOR~\cite{Deitke:2022ProcTHOR} 3D scenes, and then present AVS-HM3D, which is constructed from real indoor scenes in Habitat-Matterport3D~\cite{Ramakrishnan:2021HM3D}.

\subsubsection{AVS Dataset and AVS-ProcTHOR}
Both of our synthetic datasets are constructed using the same fully automatic curation pipeline on ProcTHOR~\cite{Deitke:2022ProcTHOR} 3D scenes, as introduced in Section~\ref{sec:data_curation}~\refofpaper{}. The AVS dataset is used for training, whereas AVS-ProcTHOR is reserved for evaluation. They share identical scene configurations; the only difference lies in the question types.

The AVS training dataset consists exclusively of binary object–existence questions of the form ``\textit{Is there a target object on the supporting object?}'', where the correct answer is always ``\textit{yes}''. In every sample, the target object mentioned in the question is guaranteed to exist on the specified supporting object, so that selecting a viewpoint that clearly reveals the object consistently yields a positive reward from the frozen VLM verifier, while uninformative viewpoints yield zero reward. This design aligns the verifier’s feedback directly with the quality of view selection during RL training.

In contrast, AVS-ProcTHOR employs more diverse question types for a more comprehensive evaluation: \emph{Existence}, \emph{Counting}, and \emph{State}. Existence questions ask whether the target object is on the supporting object, counting questions ask how many instances of the target object are on that supporting object, and state questions query the state of the target object. For the \emph{Existence} category, we additionally cast questions into a multiple-choice format to make random guessing less likely to succeed.

Building upon the notations introduced in Section~\ref{sec:data_curation}~\refofpaper{}, the overall curation pipeline consists of three stages:

\paragraph{Stage 1: Scene Modification by Question Type.}
Given a data sample, we first select a pair of a supporting object and a target object placed on it using ProcTHOR scene metadata, and modify the scene if needed to make it suitable for the specified question type. 

The scene modification rules for each question type are as follows:
\begin{itemize}[itemsep=2pt]
\item \textit{\textbf{Existence.}} No scene modification done.
\item \textit{\textbf{Counting.}} Place between two and five instances of the target object on or near the supporting object.
\item \textit{\textbf{State.}} Choose a target object with a controllable state (e.g., Faucet with on/off, Book with open/closed, Mug with filled/empty) and randomly assign one of its possible states.
\end{itemize}

\paragraph{Stage 2: Viewpoint Sampling.}
To obtain a target camera pose $s_\Tgt$, we leverage the built-in function in ProcTHOR, which returns a set of agent poses (positions and orientations) from which a specified object is both visible and interactable. Specifically, we call the \texttt{GetInteractablePoses()} built-in function on the supporting object to retrieve a list of nearby camera poses that can effectively observe it. We then sort these candidate poses by their distance to the supporting object, select the top 10 closest poses. We then randomly sample one pose from this subset and use it as the target camera pose $s_\Tgt$, provided that it satisfies the rule described in Equation~\ref{eq:o_tgt_sample}~\refofpaper{}.

Once $s_{\Tgt}$ is determined, we generate candidate query poses $s_{\Qry}$ by applying actions that move the agent slightly away from the target view. Specifically, we randomly sample 10 actions, parameterized as in Section~\ref{sec:problem_formulation}~\refofpaper{}, each consisting of a heading rotation in $[-45^\circ, 45^\circ]$, a forward translation in $[50, 150]$ (centimeter), and a view rotation in $[-45^\circ,-15^\circ] \cup [15^\circ,45^\circ]$, where the forward translation is applied in the opposite direction so that the agent moves backward from the target viewpoint. Among the resulting candidates, we randomly select one query pose that satisfies the rule described in Equation~\ref{eq:o_qry_sample}~\refofpaper{}.

In Equations~\ref{eq:o_tgt_sample} and~\ref{eq:o_qry_sample}~\refofpaper{}, we set $\epsilon_{\text{vis}}^{\text{sup}} = 5{,}000$, $\epsilon_{\text{vis}}^{\text{obj}} = 10{,}000$, and $\epsilon_{\text{inv}}^{\text{obj}} = 30$, with a $90^\circ$ field of view, and the image resolution is $512\times512$. We use the default camera height of 90 centimeters and the default camera elevation angle of $15^\circ$ downward in ProcTHOR~\cite{Deitke:2022ProcTHOR}.

%, and a fixed $15^\circ$ camera elevation.

% For each modified scene $S'$, we obtain a \emph{ground-truth evidence view} by navigating toward the chosen receptacle and selecting viewpoints that maximize the projected area of the target object, subject to the thresholds above. We then roll back along a short inverse trajectory to obtain contextual views where the target is occluded, truncated, or out-of-frustum while the supporting object remains visible, producing $(s_{\Qry}, o_{\Qry})$ and $(s_{\Tgt}, o_{\Tgt})$.

% We retain a tuple only if three conditions are met:
% (i) \emph{Partiality:} $o_{\Qry}$ contains insufficient evidence, enforced by $N_p(o_{\Qry}, x_{\text{tgt}}) < \epsilon_{\text{inv}}^{\text{obj}}$;
% (ii) \emph{Solvability:} an oracle VLM (used as verifier) correctly answers the question from $o_{\Tgt}$;
% (iii) \emph{Scale:} the projected area of $x_{\text{tgt}}$ in $o_{\Tgt}$ exceeds a minimum threshold so the object is clearly visible: $N_p(o_\Tgt, x_\Tgt) > $.

% Once pairs of $(o_{\Tgt}, o_{\Qry})$ are sampled, we retain only the samples for which the VLM verifier answers the question correctly given the target view $o_{\Tgt}$. ~\dc{do we also check if the VLM answer is wrong given a query view?}

% Since we have access to both $s_{\Qry}$ and $s_{\Tgt}$, the ground-truth continuous action $a^\ast$ that moves the agent from $s_{\Qry}$ to $s_{\Tgt}$ is computed analytically using the action-space parameterization in Section~\ref{sec:problem_formulation}\refofpaper{} and is used as supervision for SFT.

\paragraph{Stage~3: Question–Answer Generation and Filtering.}
Given the selected target/supporting objects and their relation in the scene, we instantiate rule-based question templates for each question type.
% \begin{itemize}[leftmargin=1.0em,itemsep=2pt]
% \item \textbf{existence.} Questions ask whether a specific object is on the supporting receptacle.
% \item \textbf{counting.} Questions ask how many instances of a given object are on/near the supporting receptacle.
% \item \textbf{state.} Questions ask about the state of an actionable object (e.g., ``Is the faucet on?'').
% \end{itemize}

For every data sample tuple, we retain only the samples for which the VLM verifier answers the question correctly given the target view $o_{\Tgt}$.

% \paragraph{AVS Dataset vs. AVS-ProcTHOR.}
% Both the training AVS dataset and the AVS-ProcTHOR benchmark follow the pipeline above, but they differ in question design:
% \begin{itemize}[leftmargin=1.0em,itemsep=2pt]
% \item \textbf{AVS Dataset (training).} For training, we use only simple binary \emph{existence} questions where the answer is always ``yes’’ and the target object is guaranteed to be present in $o_{\Tgt}$. This, together with the oracle-verifier check on $o_{\Tgt}$, makes the reward $r^{\text{ver}}$ used in RL training nearly noise-free and monotonic in viewpoint quality, which is crucial for stable policy optimization in the continuous action space.
% \item \textbf{AVS-ProcTHOR (evaluation).} For evaluation, we apply the same curation procedure but use a richer set of question types: \emph{existence}, \emph{counting}, and \emph{state}. Existence questions are further cast into a multiple-choice format to reduce the chance of correct answers by random guessing. This benchmark thus tests both view-selection quality and downstream question answering under more diverse reasoning demands.
% \end{itemize}

\subsubsection{AVS-HM3D}
As discussed in Section~\ref{sec:vgavs_results}~\refofpaper{}, we construct an additional benchmark, AVS-HM3D, on real indoor environments from Habitat-Matterport3D~\cite{Ramakrishnan:2021HM3D} to assess the generalization of our method beyond the synthetic ProcTHOR scenes used for training. We reuse triplets of (question, answer, ground-truth view) from the validation split of the Fine-EQA dataset~\cite{Liu:2024EXPRESSBench}, where the ground-truth view is the human-annotated frame from which the question–answer pair is derived. Among its seven question types, we discard those that rely on external world knowledge or holistic layout priors and retain five locally grounded types: \emph{Attribute}, \emph{Counting}, \emph{Existence}, \emph{Object}, and \emph{State}. Given the ground-truth view, we then sample five candidate query views by applying small actions that move the agent away from the ground-truth view, following the same procedure as in the ProcTHOR setup. Due to the absence of scene metadata (e.g., object visibility and object relations) in real scenes, we manually select for each triplet, the query view that offers partial visual evidence---sufficient to guide the agent toward the ground-truth view, yet insufficient to directly answer the question. Using a GPT-5 model~\cite{OpenAI:2025GPT5}, we filter out low-quality items, such as cases where the question is not solvable from the ground-truth view or the ground-truth view has poor visual quality. The resulting curated benchmark comprises 208 samples.

% We further filter low-quality items using an automatic validator~\cite{OpenAI:2025GPT5} applied to each triplet (e.g., removing samples whose question is not solvable from the ground-truth view or whose ground-truth view is of poor quality). As in ProcTHOR, we sample five candidate query views by applying inverse actions from the goal view to induce partial observations and manually select, for each item, the query view that provides the strongest partial evidence. The resulting curated benchmark comprises 208 question–view pairs.

% \subsection{Experiment Results}
% \subsection{Training Details}
% We use Qwen2.5-VL-7B~\cite{Bai:2025Qwen2.5VL} as the backbone VLM.
% We train the model with a batch size of 32 for SFT and 128 for RL, running 7 epochs of SFT followed by 5 additional epochs of GRPO-based reinforcement learning.
% The total reward is a weighted sum of a format reward and a verifier reward, with weights 0.3 and 1.0, respectively. During GRPO training, we use a separate, frozen Qwen2.5-VL-7B~\cite{Bai:2025Qwen2.5VL} as the VLM verifier for computing verifiable reward $r^{\text{ver}}$, with a group size of 16. The learning rate is $2e-5$ for SFT and $1e-6$ for GRPO training.

\subsection{Prompts}
\label{sec:supp_prompts}
We present our input prompts in Figure~\ref{fig:prompt}, including the action prompt that enforces the VLM to predict action parameters and the format prompt that specifies the desired output structure. In the action prompt, \texttt{VQA\_question} is a placeholder that is replaced by the actual question about the scene. When training purely with RL, we adopt a ``think-then-act'' format, whereas during SFT the model is only supervised to directly predict the action parameters without any reasoning traces. Directly switching an SFT-trained model to the think-then-act format often causes a degenerate behavior where the model outputs the action first and then hallucinates a post-hoc ``thinking'' process. To mitigate this mismatch, we redesign the RL output format so that the model first makes an initial action guess, then reasons to refine it, and finally outputs a revised action, which encourages a smooth transition from direct prediction to genuine reason-before-acting behavior.

\newtcolorbox{promptbox}[1][]{%
  enhanced,
  % breakable,
  colback=white,
  colframe=black!20,
  % colbacktitle=black!5,
  % coltitle=black,
  % fonttitle=\bfseries,
  % title={#2},
  boxrule=1pt,
  arc=3pt,
  left=6pt,right=6pt,top=6pt,bottom=6pt,
  fontupper=\ttfamily\small, % <-- 전체 내용을 \texttt 폰트로
  #1
}

\newcommand{\ThinkTagColor}[1]{\textcolor{Blue}{\textbf{#1}}}
\newcommand{\ActionTagColor}[1]{\textcolor{Brown}{\textbf{#1}}}

\begin{figure*}
    \centering
    \begin{promptbox}
    \textbf{\textcolor{black}{Action Prompt:}}\\
    You are an embodied agent navigating a 3D scene from an egocentric camera.
    Given the current image and a question about the scene, predict the optimal NEXT action parameters to reach a better viewpoint.
    \\
    Action parameters (return integers only): \\
    1) Heading rotation (deg) in (-180, 180]: Azimuth yaw about your vertical axis BEFORE moving.
       Positive = clockwise/right, negative = counterclockwise/left, 0 = no rotation. \\
    2) Forward distance (cm) >= 0: Move forward in the NEW facing direction after the rotation. 0 = no move. \\
    3) View rotation (deg) in (-180, 180]: Final azimuth adjustment AFTER moving, relative to your post-move heading. Same sign convention as rotation. \\
    \\
    Goal: Choose a heading rotation angle, moving forward distance, final-viewing rotation angle that maximizes visibility of task-relevant objects and minimizes occlusion.
    Example: Rotating -90 degrees, moving forward 50 cm, then rotating 90 degrees is equivalent to translating 50 cm to your left while keeping the original heading. \\
    Question: \textcolor{cyan}{\textbf{"\{VQA\_question\}"}} \\
    DO NOT answer the question; ONLY predict the next action parameters. \\
    
    \textbf{RL Format Prompt:}\\
    First output the reasoning process in \ThinkTagColor{<think> </think>} tags. Then, output the final predictions in \ActionTagColor{<head> </head>, <fwd> </fwd>, <view> </view>} tags in order.\\
    The text between \ActionTagColor{<head>} and \ActionTagColor{</head>} must be the angle in degrees (-180, 180], \ActionTagColor{<fwd>} and \ActionTagColor{</fwd>} must be the nonnegative forward distance, and \ActionTagColor{<view>} and \ActionTagColor{</view>} must be the final viewing angle in degrees (-180, 180].\\
    Each must be exactly one integer number (no units, no extra text).\\
    In the reasoning process, explicitly reason about (1) how much to rotate to determine the moving direction, (2) how far to move forward to approach, (3) how much to further adjust your azimuth angle from your moving direction for the best view.\\

    \textbf{SFT-then-RL Format Prompt:}\\
    First, output your initial guess for the action parameter values.\\
    Then, think carefully to refine your initial guess for each action parameter.\\
    After output initial guess for the action parameters, output your reasoning process within \ThinkTagColor{<think> </think>} tags, and then provide the final guess within \ActionTagColor{<head> </head>}, \ActionTagColor{<fwd> </fwd>}, and \ActionTagColor{<view> </view>} tags, respectively.\\
    The text between \ActionTagColor{<head>} and \ActionTagColor{</head>} must be the rotation angle in degrees in the range (-180, 180]; the text between \ActionTagColor{<fwd>} and \ActionTagColor{</fwd>} must be the nonnegative forward distance; and the text between \ActionTagColor{<view>} and \ActionTagColor{</view>} must be the final viewing angle in degrees in the range (-180, 180]. \\
    Each must be exactly one integer (no units, no extra text).
    In the reasoning process, explicitly reason about (1) how much to rotate to determine the moving direction, (2) how far to move forward to approach, (3) how much to further adjust your azimuth angle from your moving direction for the best view.\\
    \\
    For example:\\
    \ActionTagColor{<head>} INITIAL\_GUESS \ActionTagColor{</head>} \ActionTagColor{<fwd>} INITIAL\_GUESS \ActionTagColor{</fwd>} \ActionTagColor{<view>} INITIAL\_GUESS \ActionTagColor{</view>} \\
    \ThinkTagColor{<think>} REASONING PROCESS \ThinkTagColor{</think>}\\
    \ActionTagColor{<head>} FINAL\_GUESS \ActionTagColor{</head>} \ActionTagColor{<fwd>} FINAL\_GUESS \ActionTagColor{</fwd>} \ActionTagColor{<view>} FINAL\_GUESS \ActionTagColor{</view>}
    \end{promptbox}
\caption{\textbf{Input System Prompts.} The action prompt provides the task context and action parameterization explanation, while the format prompt specifies the required reasoning trace and output tag structure.}
\label{fig:prompt}
\end{figure*}

\subsection{More Qualitative Comparisons}
\label{sec:supp_more_qualitative_comparisons}
% Preamble (once)
% \usepackage{graphicx,booktabs,tabularx,array,xcolor}
% \newcolumntype{Y}{>{\centering\arraybackslash}X}
% \setlength{\tabcolsep}{1pt}
% preamble

%\definecolor{wrongblue}{RGB}{38,82,170}
%\definecolor{correctred}{RGB}{192,57,43}

% \definecolor{correctgreen}{RGB}{105,182,28}
% \definecolor{wrongred}{RGB}{243,43,12}
% \definecolor{correctgreen}{RGB}{132,153,79}
% \definecolor{wrongred}{RGB}{243,1,3}

\renewcommand{\imgframe}[2]{
\begingroup \setlength{\fboxsep}{0pt}
\setlength{\fboxrule}{1.3pt}% 테두리 두께 (원하면 조절) 
\fcolorbox{#1}{white}{\includegraphics[width=1.0\linewidth,keepaspectratio]{#2}}% 
\endgroup }

\renewcommand{\hcell}[1]{\makecell{\scriptsize #1}}
\begin{figure*}[t]
\scriptsize
\centering
\renewcommand{\arraystretch}{1.0}
\setlength{\tabcolsep}{1pt}
\arrayrulecolor{black}
\setlength{\arrayrulewidth}{0.2pt}

\renewcommand{\wrongimg}[2][]{%
  \begin{overpic}[#1]{#2}
    \put(3,75){\color{wrongred}\fontsize{15}{15}\selectfont\bfseries\xmark}
  \end{overpic}
}

\renewcommand{\correctimg}[2][]{%
  \begin{overpic}[#1]{#2}
    \put(3,75){\color{correctgreen}\fontsize{15}{15}\selectfont\bfseries\cmark}
  \end{overpic}
}

\begin{tabularx}{\linewidth}{YYYYYYYYY}
\toprule
% ===== Category header row =====
& \multicolumn{1}{c|}{} & \hcell{EQA\\Framework} & \hcell{Backbone\\Model} & \hcell{Proprietary\\Model} & \hcell{Spatial\\VLM} & \multicolumn{3}{|c}{\hcell{AVS Framework}} \\
\midrule
% ===== Model-name header row (no bold, smaller font) =====
Query View & \multicolumn{1}{c|}{Target View} & \hcell{Fine-EQA~\cite{Liu:2024EXPRESSBench}} & \hcell{Qwen-2.5-VL~\cite{Bai:2025Qwen2.5VL}} & \hcell{GPT-5~\cite{OpenAI:2025GPT5}} & \hcell{Spatial\\Reasoner~\cite{Ma:2025SpatialReasoner}} & \multicolumn{1}{|c}{\hcell{SFT}} & \hcell{RL} & \hcell{\textbf{SFT+RL (Ours)}} \\
\midrule
\multicolumn{9}{c}{\footnotesize (Counting) ``\textit{How many\targetobj{\textbf{pans}} near the\receptacle{\textbf{bed}}?}''}\\[4pt]
\includegraphics[width=\linewidth,keepaspectratio]{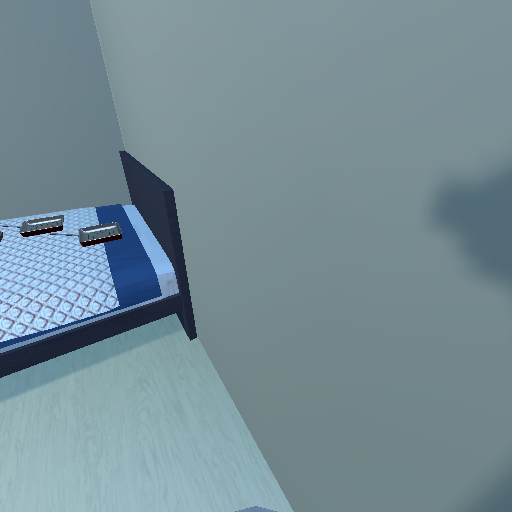} &
\includegraphics[width=\linewidth,keepaspectratio]{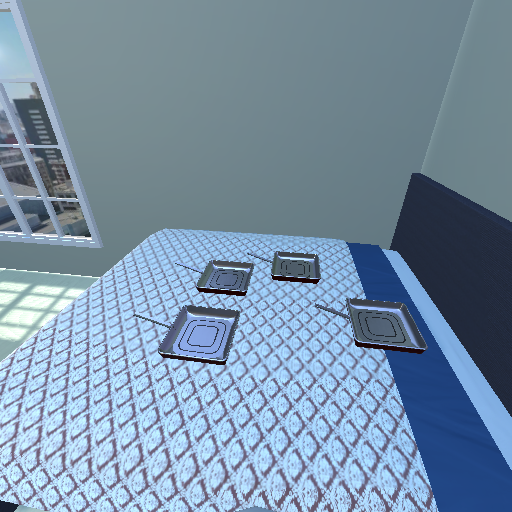} &
\wrongimg[width=\linewidth,keepaspectratio]{{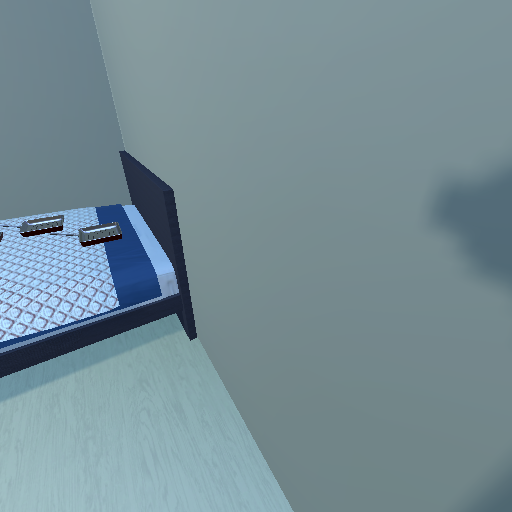}} &
\wrongimg[width=\linewidth,keepaspectratio]{{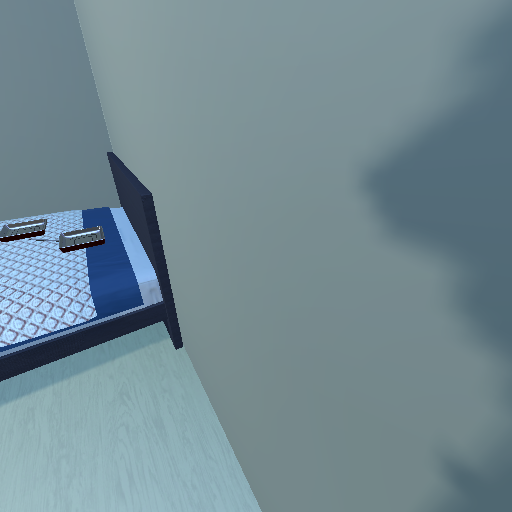}} &
\wrongimg[width=\linewidth,keepaspectratio]{{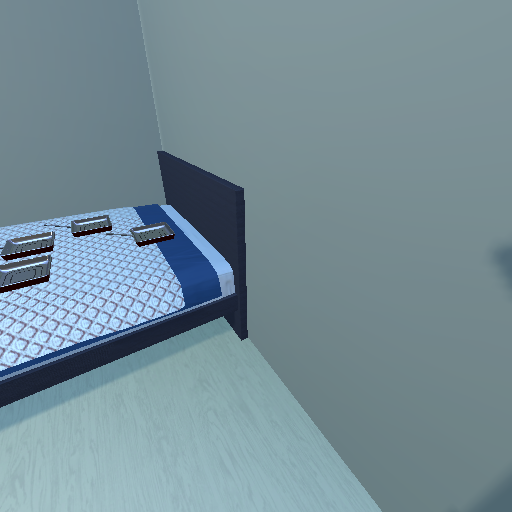}} &
\wrongimg[width=\linewidth,keepaspectratio]{{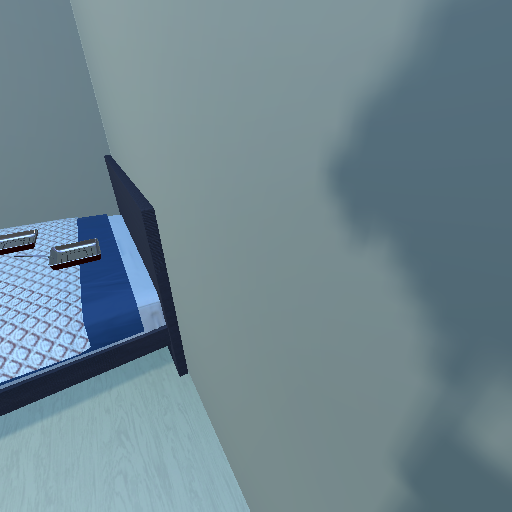}} &
\correctimg[width=\linewidth,keepaspectratio]{{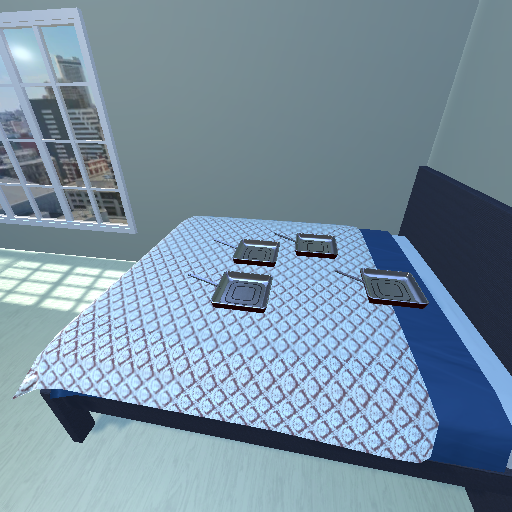}} &
\correctimg[width=\linewidth,keepaspectratio]{{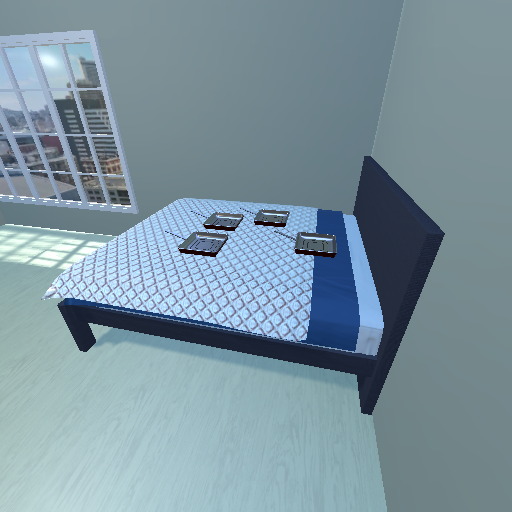}} &
\correctimg[width=\linewidth,keepaspectratio]{{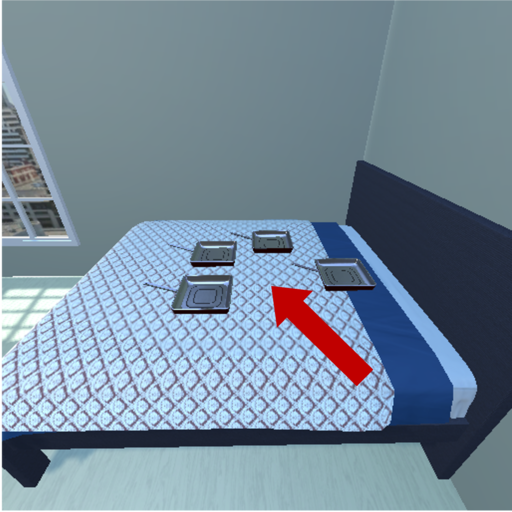}} \\

\midrule

\multicolumn{9}{c}{\footnotesize (Counting) ``\textit{How many\targetobj{\textbf{pots}} are there on the\receptacle{\textbf{desk}}?}''}\\[4pt]
\includegraphics[width=\linewidth,keepaspectratio]{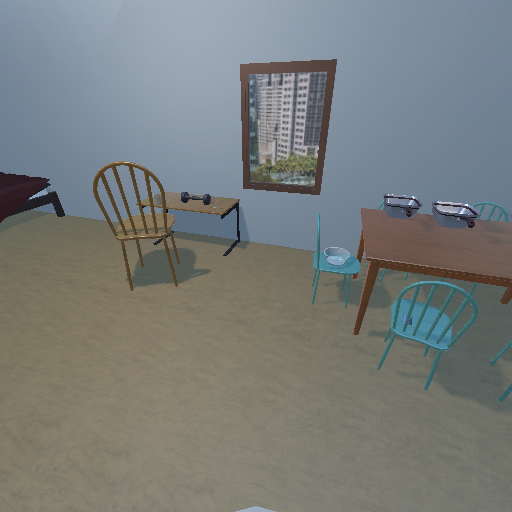} &
\includegraphics[width=\linewidth,keepaspectratio]{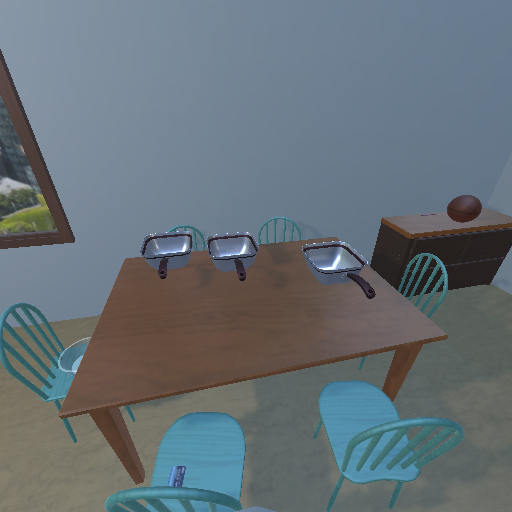} &
\wrongimg[width=\linewidth,keepaspectratio]{{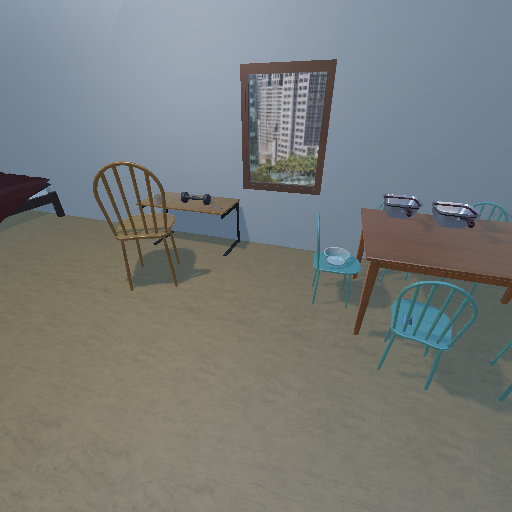}} &
\wrongimg[width=\linewidth,keepaspectratio]{{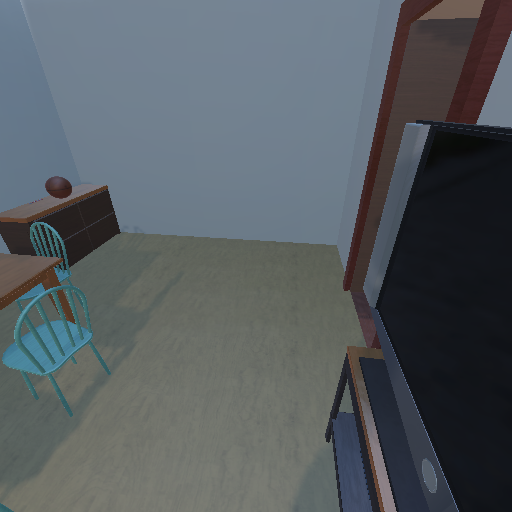}} &
\wrongimg[width=\linewidth,keepaspectratio]{{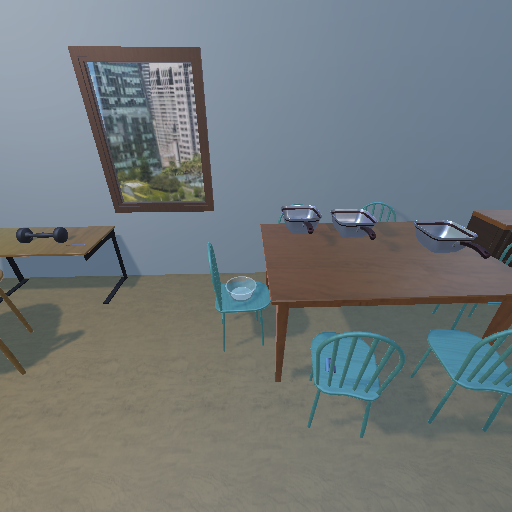}} &
\wrongimg[width=\linewidth,keepaspectratio]{{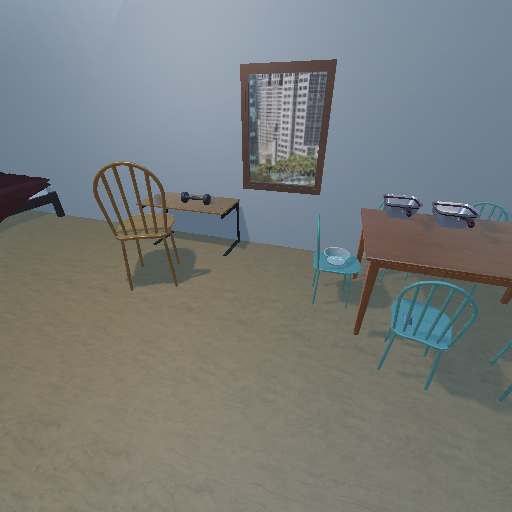}} &
\correctimg[width=\linewidth,keepaspectratio]{{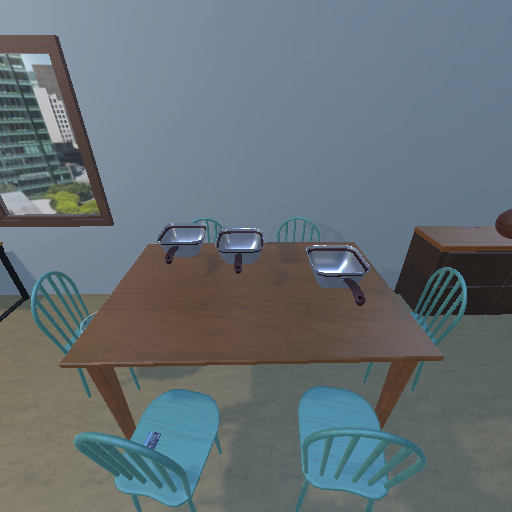}} &
\correctimg[width=\linewidth,keepaspectratio]{{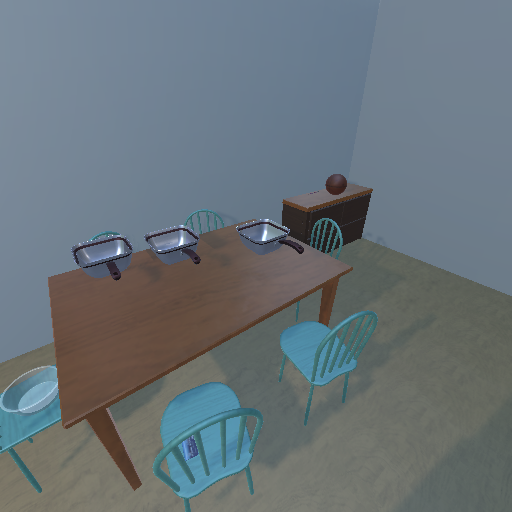}} & 
\correctimg[width=\linewidth,keepaspectratio]{{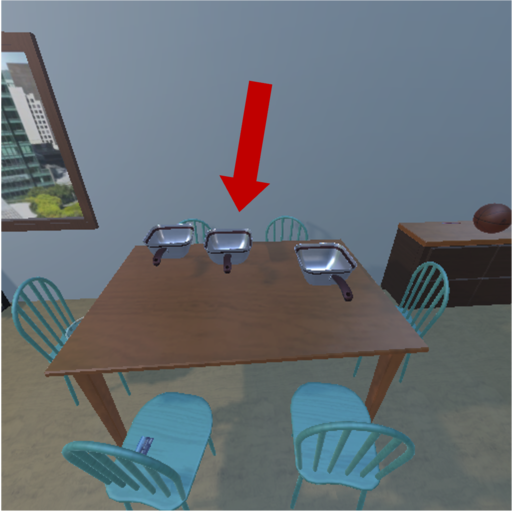}} \\

\midrule
%\multicolumn{9}{c}{\Benchmark{}-HM3D}
%\midrule
%\RowNine{figures/quali_cropped/hm3d/gpt_vilasr_0165}{(State) ``\textit{Is the \textbf{lamp} in the bedroom next to the window \textbf{turned on}?}''}

% ===== Category header row for second block (Gemini / SpatialReasoner) =====
% ===== Model-name header row variant (Gemini / SpatialReasoner) =====
%\hcell{Query View} & \hcell{Target View} & \hcell{Fine-EQA~\cite{Liu:2024EXPRESSBench}} & \hcell{Qwen-2.5-VL~\cite{Bai:2025Qwen2.5VL}} & \hcell{Gemini-2.5\\Pro~\cite{Google:2025Gemini25}} & \hcell{Spatial\\ Reasoner~\cite{Ma:2025SpatialReasoner}} & \hcell{SFT} & \hcell{RL} & \hcell{\textbf{Ours}} \\
%\midrule 
% ---- Rows (Gemini / SpatialReasoner set) ----
%\RowNine{figures/quali_cropped/procthor/counting_gemini_spatialreasoner_0478}{(Counting) ``\textit{How many \textbf{mugs} near the \textbf{diningtable}?}''}

\multicolumn{9}{c}{\footnotesize (State) ``\textit{Choose the state of the\targetobj{\textbf{book}} on the\receptacle{\textbf{drawer}}. A: opened B: closed}''}\\[4pt]
\includegraphics[width=\linewidth,keepaspectratio]{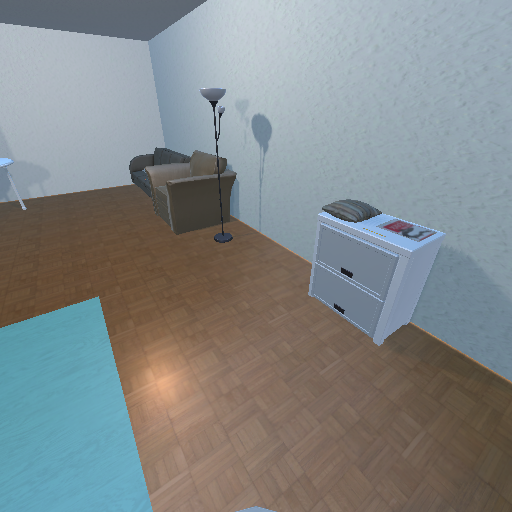} &
\includegraphics[width=\linewidth,keepaspectratio]{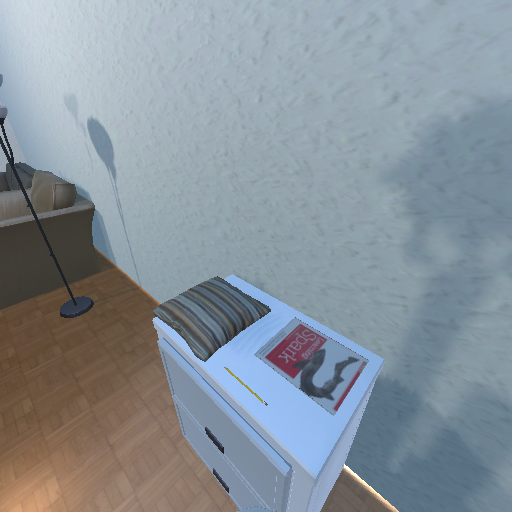} &
%\fcolorbox{blue}{white}
\wrongimg[width=\linewidth,keepaspectratio]{{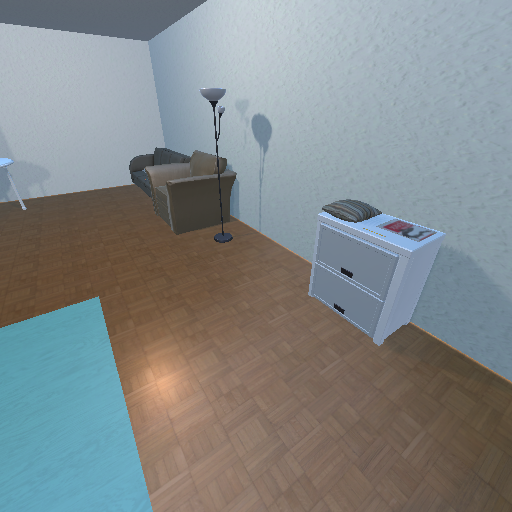}} &
\wrongimg[width=\linewidth,keepaspectratio]{{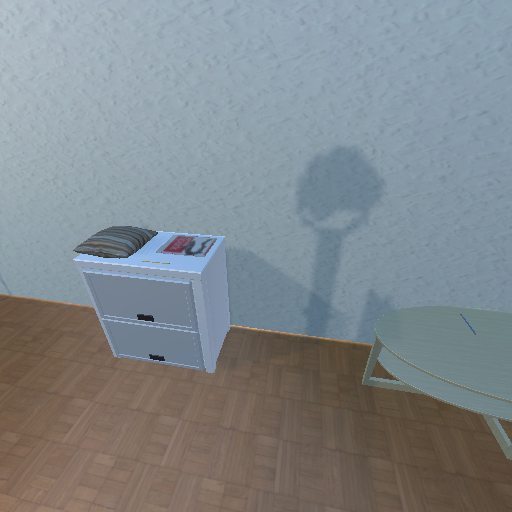}} &
\correctimg[width=\linewidth,keepaspectratio]{{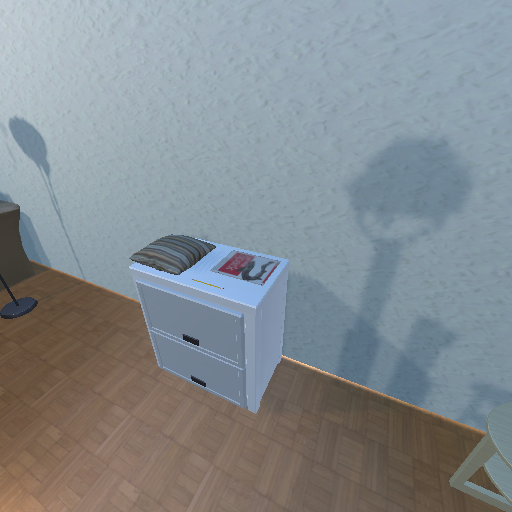}} &
\wrongimg[width=\linewidth,keepaspectratio]{{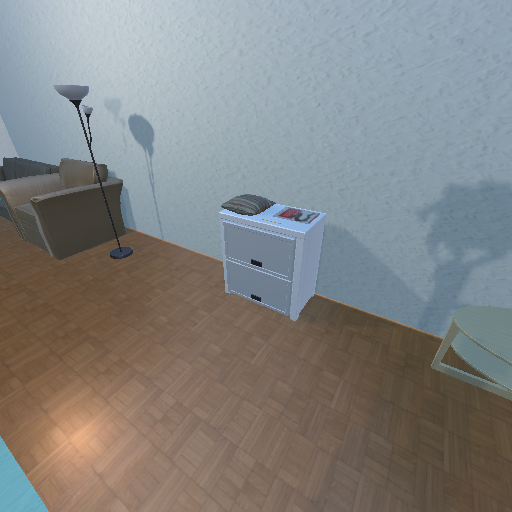}} &
\correctimg[width=\linewidth,keepaspectratio]{{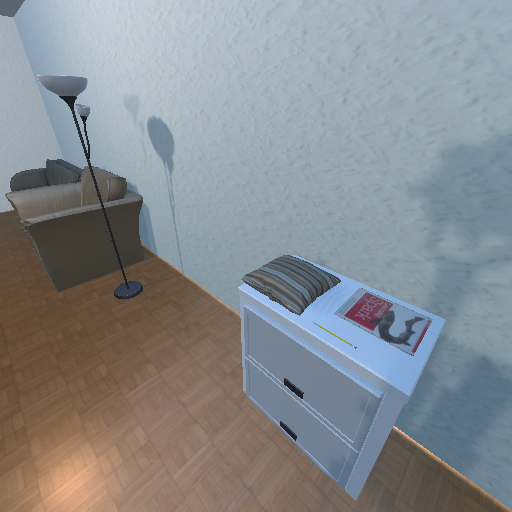}} &
\correctimg[width=\linewidth,keepaspectratio]{{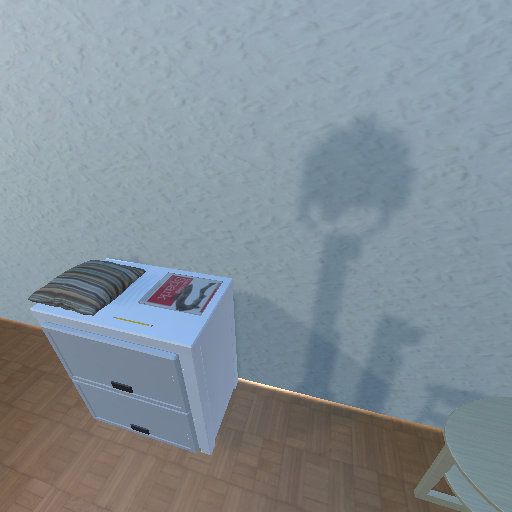}} &
\correctimg[width=\linewidth,keepaspectratio]{{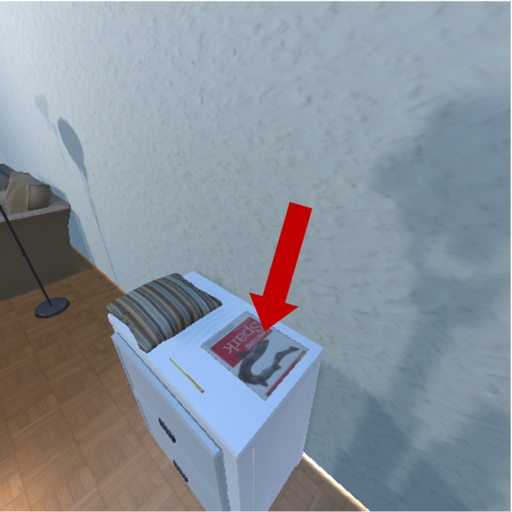}} \\

\midrule

\multicolumn{9}{c}{\footnotesize (State) ``\textit{Choose the state of the\targetobj{\textbf{box}} on the\receptacle{\textbf{dining table}}. A: closed B: opened}''}\\[4pt]
\includegraphics[width=\linewidth,keepaspectratio]{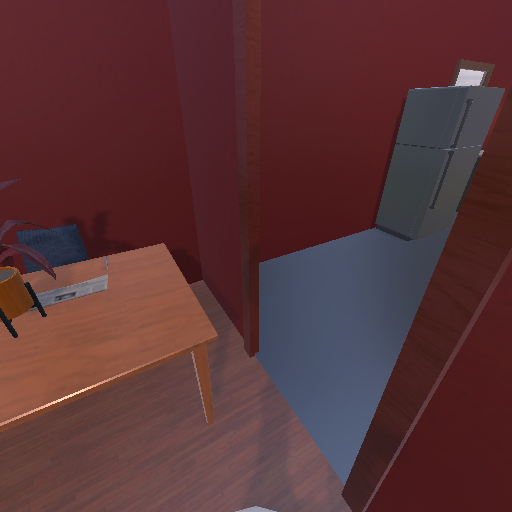} &
\includegraphics[width=\linewidth,keepaspectratio]{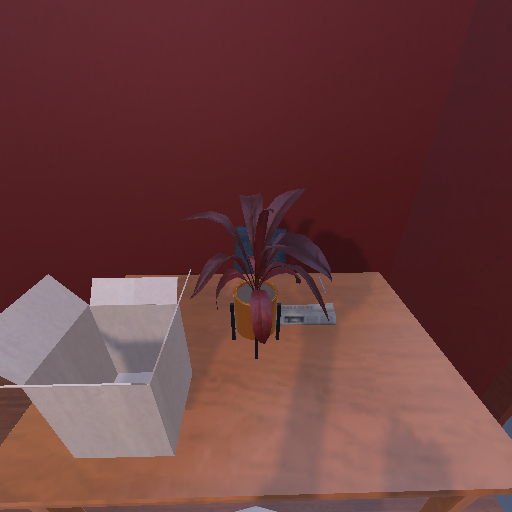} &
%\fcolorbox{blue}{white}
\wrongimg[width=\linewidth,keepaspectratio]{{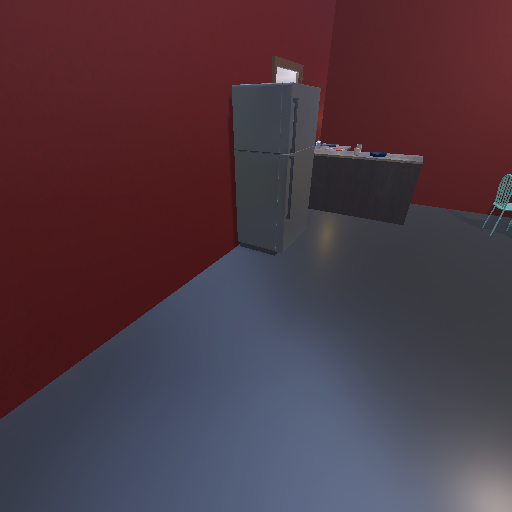}} &
\wrongimg[width=\linewidth,keepaspectratio]{{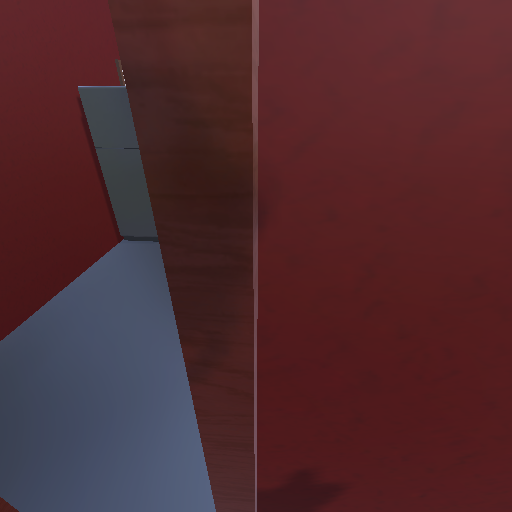}} &
\wrongimg[width=\linewidth,keepaspectratio]{{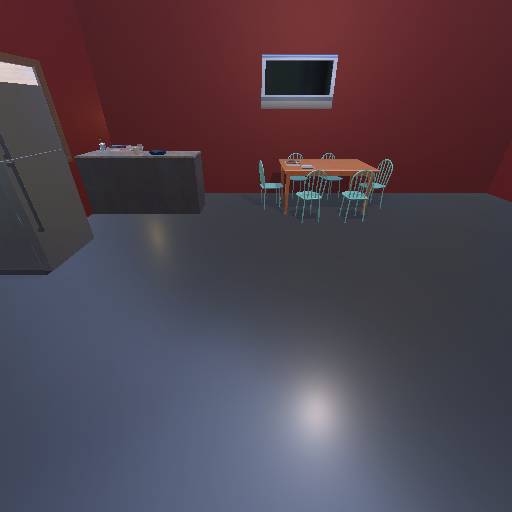}} &
\wrongimg[width=\linewidth,keepaspectratio]{{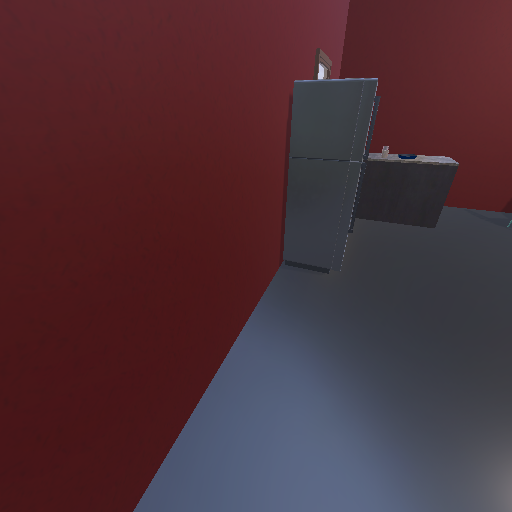}} &
\correctimg[width=\linewidth,keepaspectratio]{{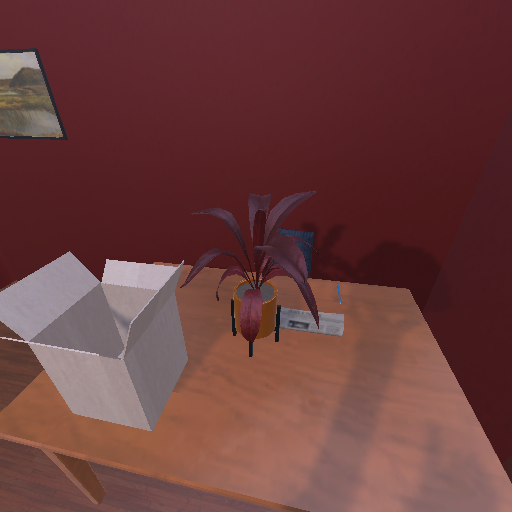}} &
\correctimg[width=\linewidth,keepaspectratio]{{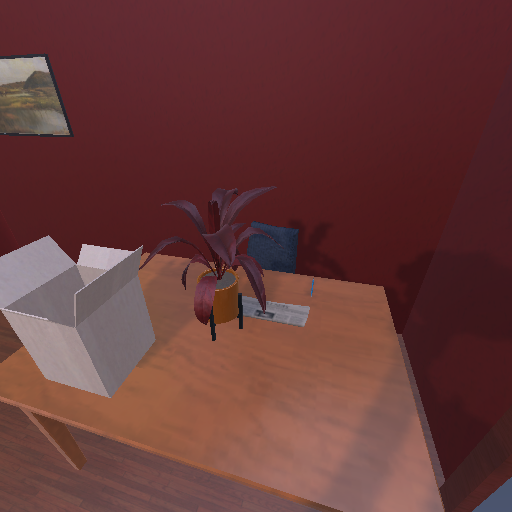}} &
\correctimg[width=\linewidth,keepaspectratio]{{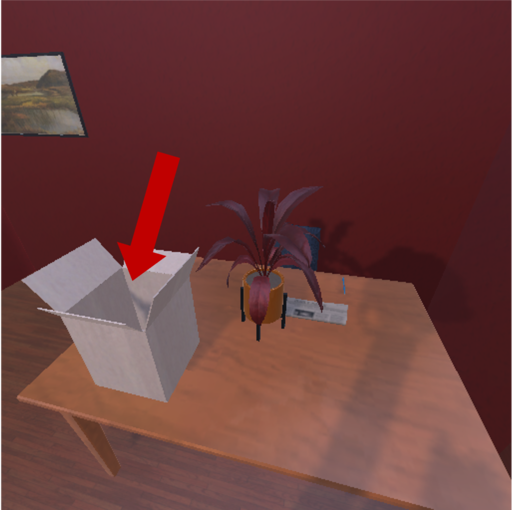}} \\

\midrule

\multicolumn{9}{c}{\footnotesize (Existence) ``\textit{Is there a\targetobj{\textbf{plant}} on the small\receptacle{\textbf{table}} in the sitting area in the study?}''}\\[4pt]
\includegraphics[width=\linewidth,keepaspectratio]{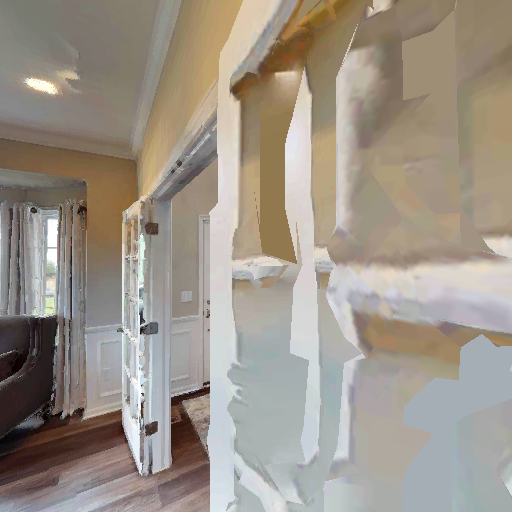} &
\includegraphics[width=\linewidth,keepaspectratio]{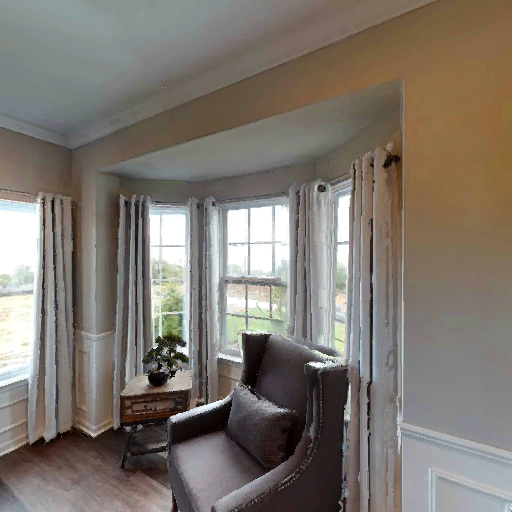} &
\wrongimg[width=\linewidth,keepaspectratio]{{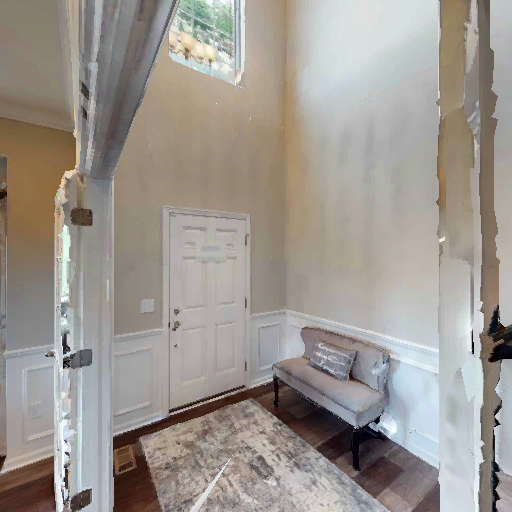}} &
\wrongimg[width=\linewidth,keepaspectratio]{{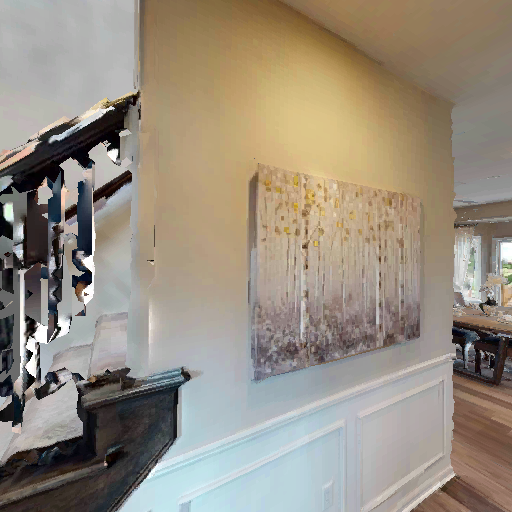}} &
\wrongimg[width=\linewidth,keepaspectratio]{{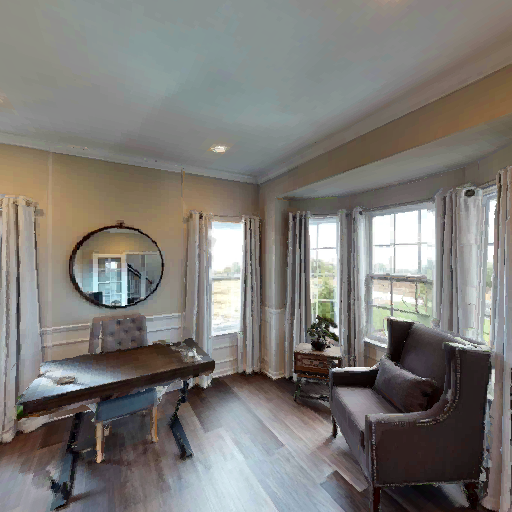}} &
\wrongimg[width=\linewidth,keepaspectratio]{{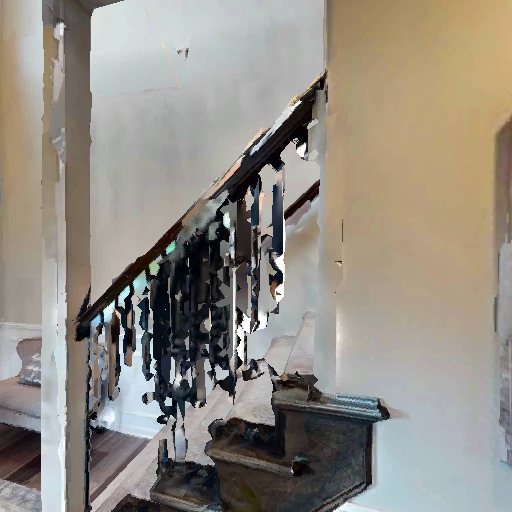}} &
\correctimg[width=\linewidth,keepaspectratio]{{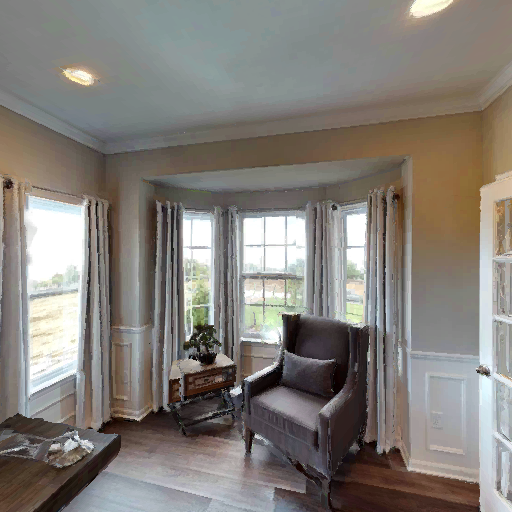}} &
\correctimg[width=\linewidth,keepaspectratio]{{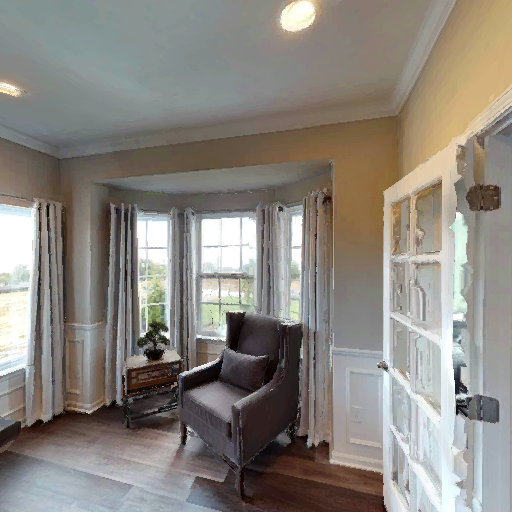}} &
\correctimg[width=\linewidth,keepaspectratio]{{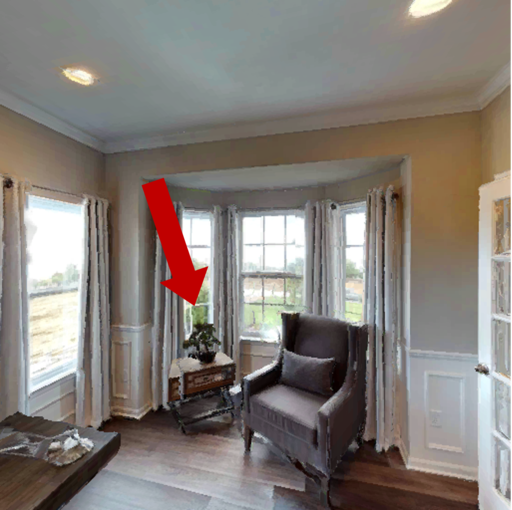}}
\\

\midrule

\multicolumn{9}{c}{\footnotesize (Counting) ``\textit{How many\targetobj{\textbf{paintings}} are hanging in the\receptacle{\textbf{entryway?}}}''}\\[4pt]
\includegraphics[width=\linewidth,keepaspectratio]{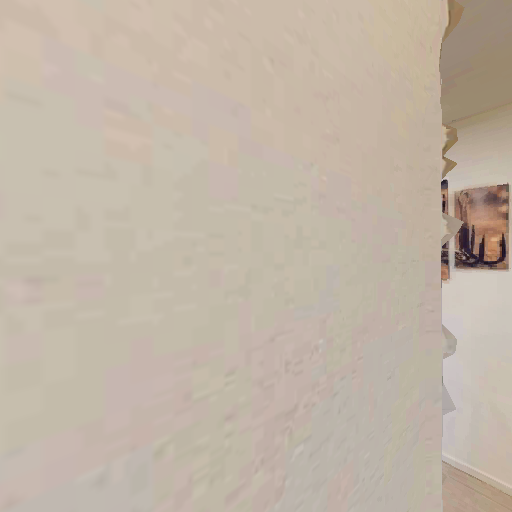} &
\includegraphics[width=\linewidth,keepaspectratio]{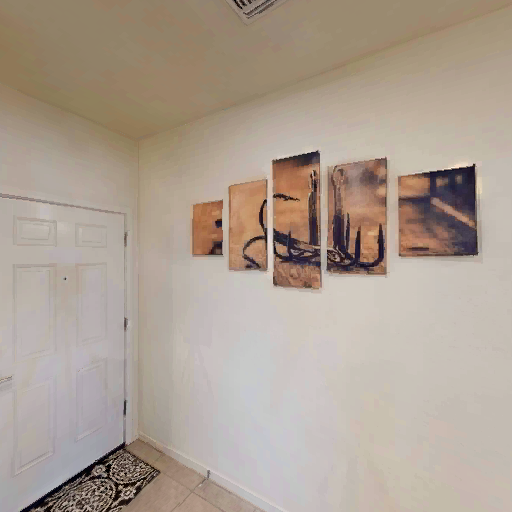} &
\wrongimg[width=\linewidth,keepaspectratio]{{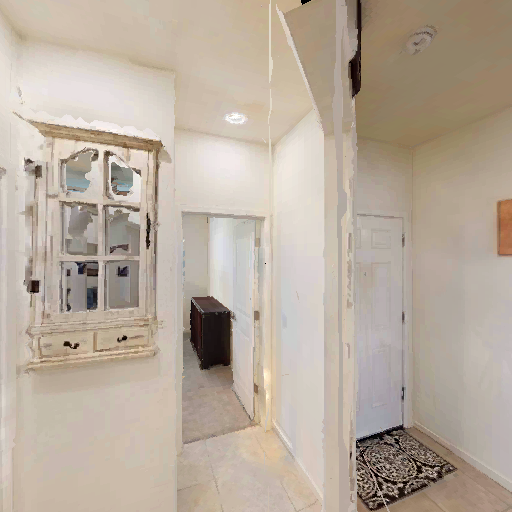}} &
\wrongimg[width=\linewidth,keepaspectratio]{{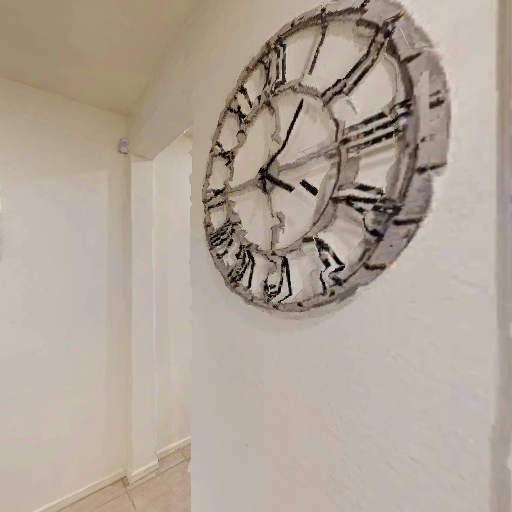}} &
\wrongimg[width=\linewidth,keepaspectratio]{{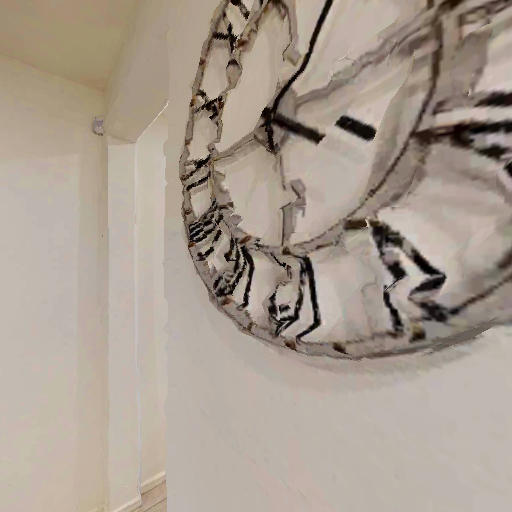}} &
\wrongimg[width=\linewidth,keepaspectratio]{{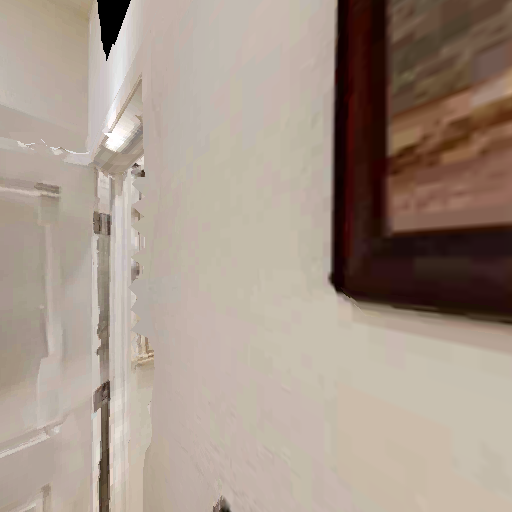}} &
\correctimg[width=\linewidth,keepaspectratio]{{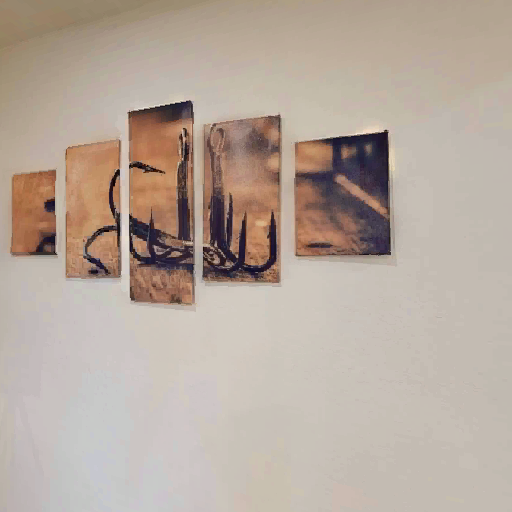}} &
\correctimg[width=\linewidth,keepaspectratio]{{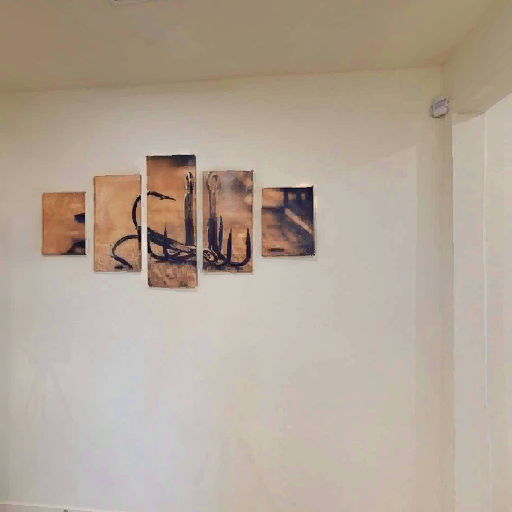}} &
\correctimg[width=\linewidth,keepaspectratio]{{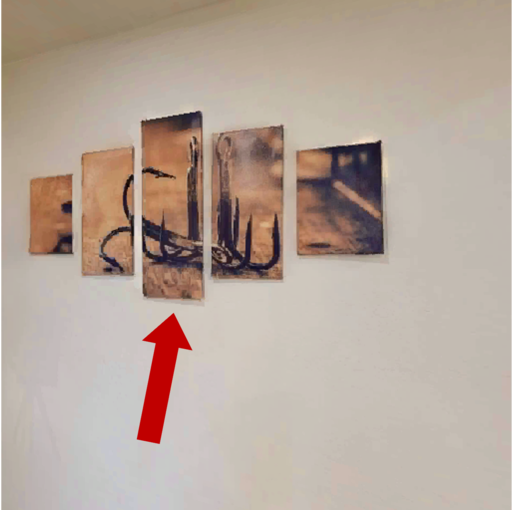}} \\

\midrule

\multicolumn{9}{c}{\footnotesize (State) ``\textit{Is the\targetobj{\textbf{light}} in the bedroom currently on?}''}\\[4pt]
\includegraphics[width=\linewidth,keepaspectratio]{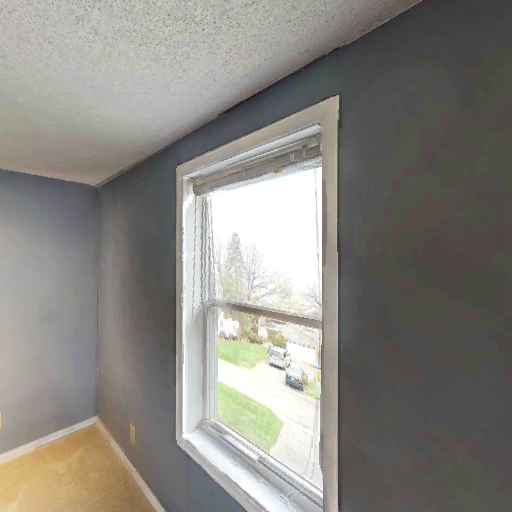} &
\includegraphics[width=\linewidth,keepaspectratio]{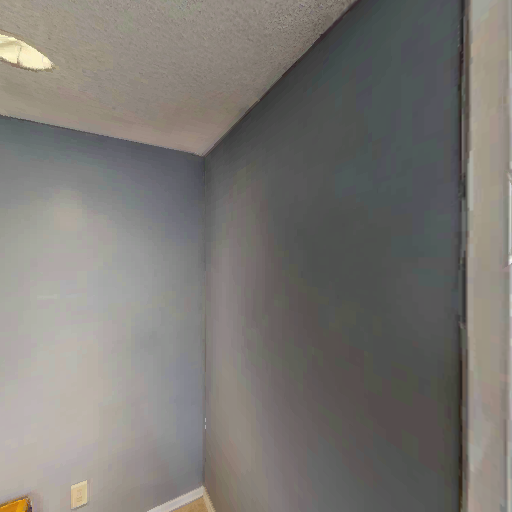} &
\wrongimg[width=\linewidth,keepaspectratio]{{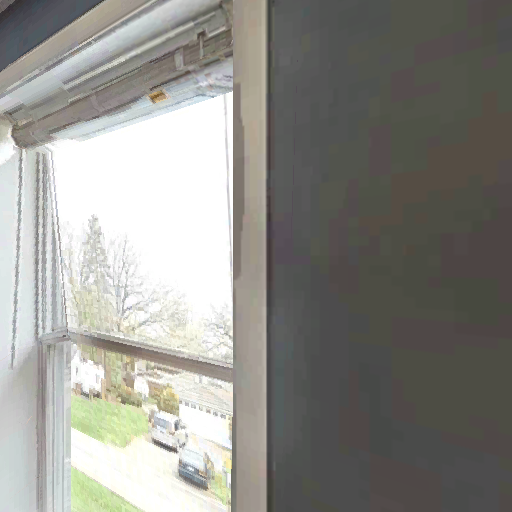}} &
\wrongimg[width=\linewidth,keepaspectratio]{{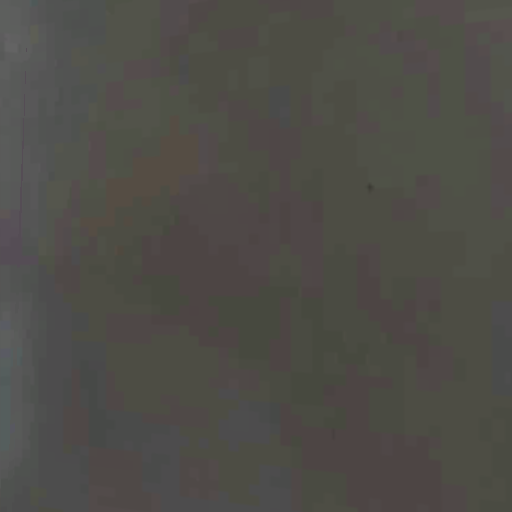}} &
\wrongimg[width=\linewidth,keepaspectratio]{{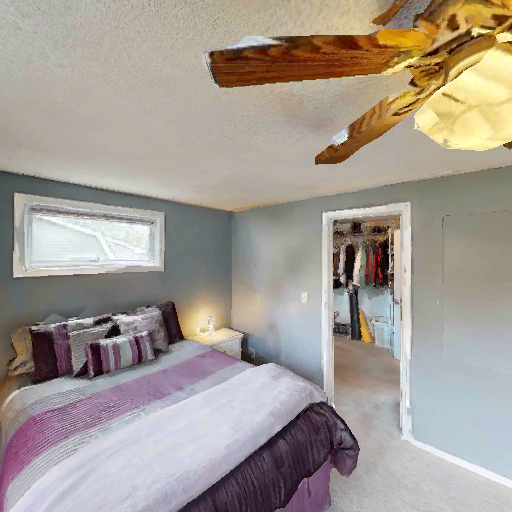}} &
\wrongimg[width=\linewidth,keepaspectratio]{{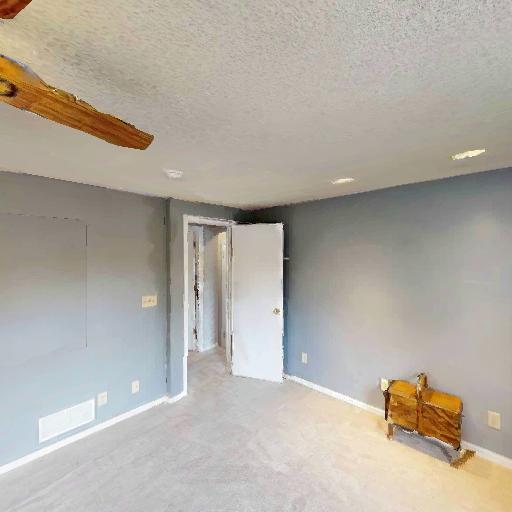}} &
\correctimg[width=\linewidth,keepaspectratio]{{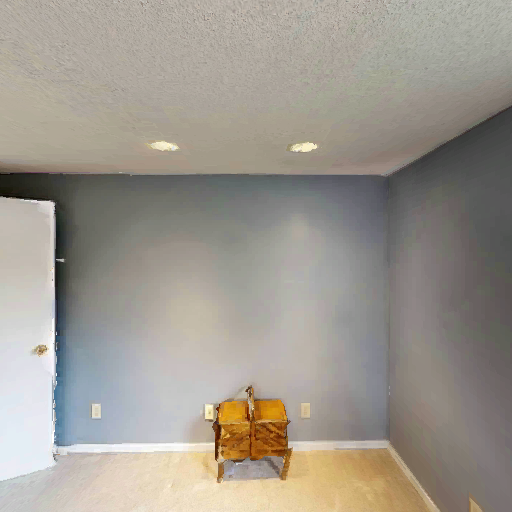}} &
\correctimg[width=\linewidth,keepaspectratio]{{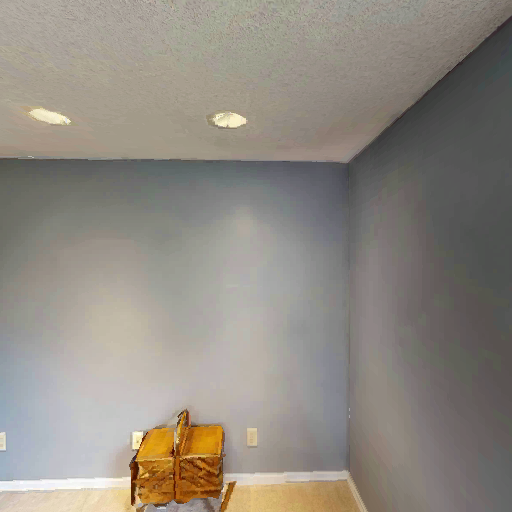}} &
\correctimg[width=\linewidth,keepaspectratio]{{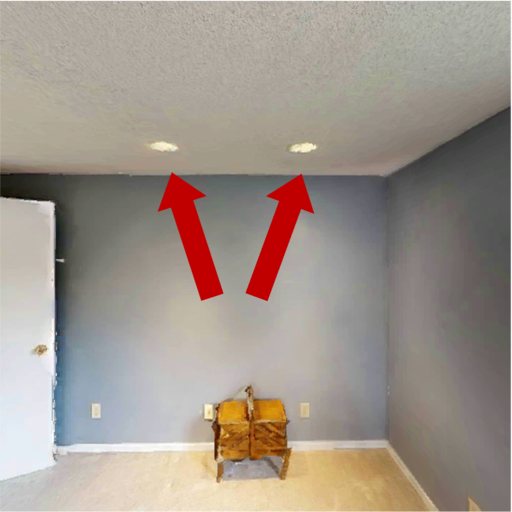}} \\

\midrule

\multicolumn{9}{c}{\footnotesize (State) ``\textit{Did I hung up the\targetobj{\textbf{paintings}} in the\receptacle{\textbf{hallway}}?}''}\\[4pt]
\includegraphics[width=\linewidth,keepaspectratio]{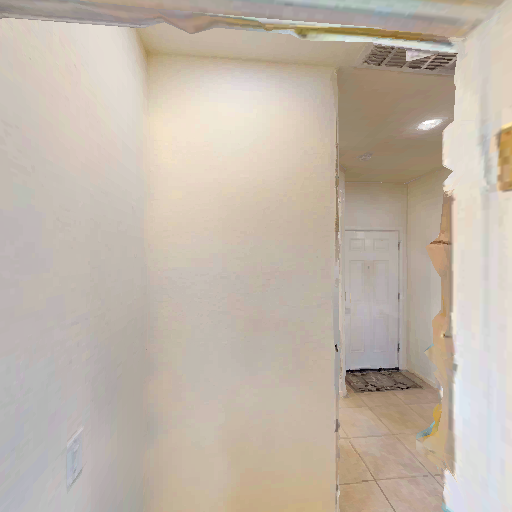} &
\includegraphics[width=\linewidth,keepaspectratio]{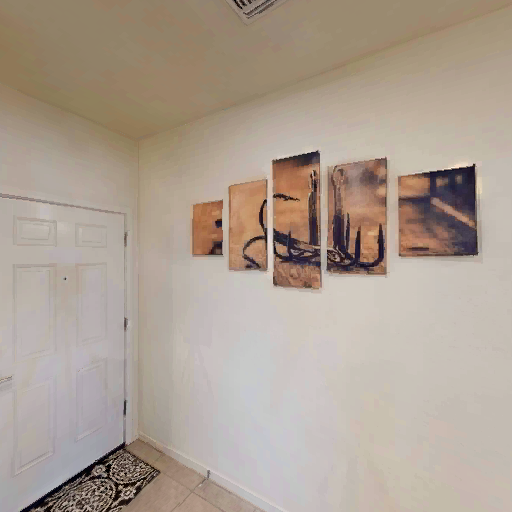} &
\wrongimg[width=\linewidth,keepaspectratio]{{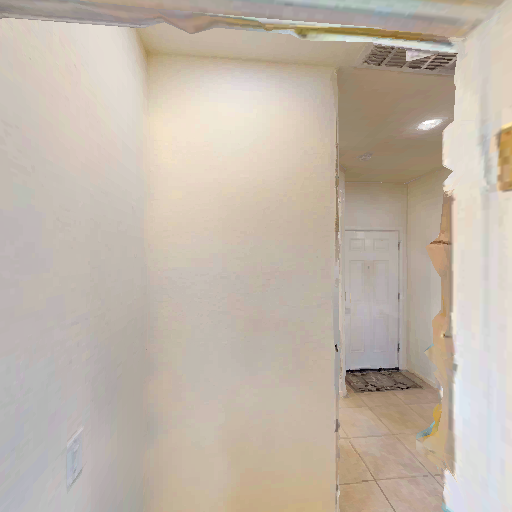}} &
\wrongimg[width=\linewidth,keepaspectratio]{{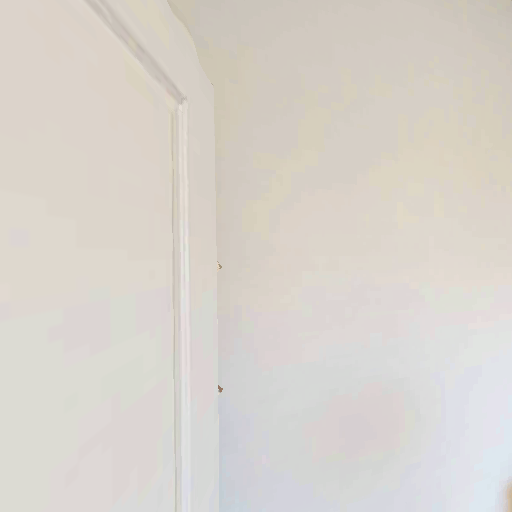}} &
\wrongimg[width=\linewidth,keepaspectratio]{{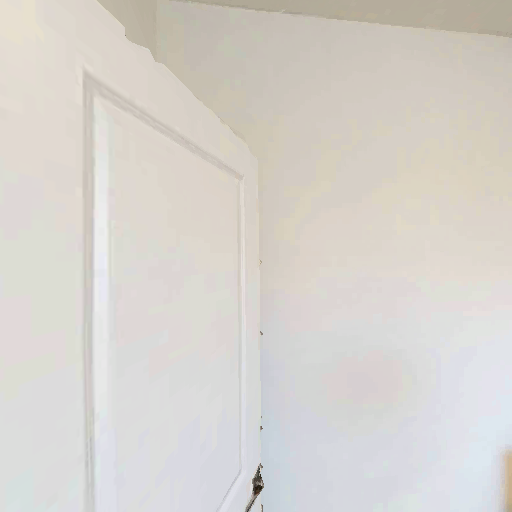}} &
\wrongimg[width=\linewidth,keepaspectratio]{{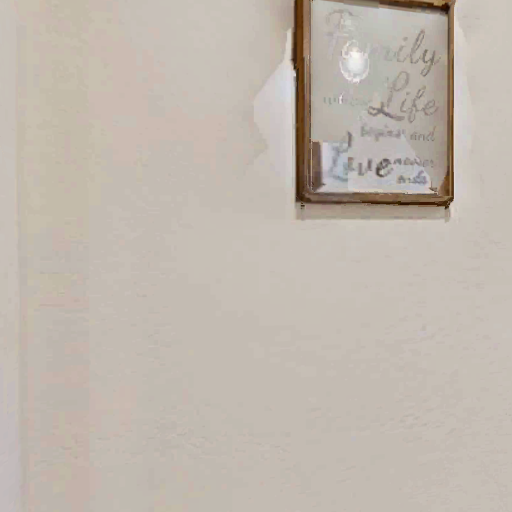}} &
\correctimg[width=\linewidth,keepaspectratio]{{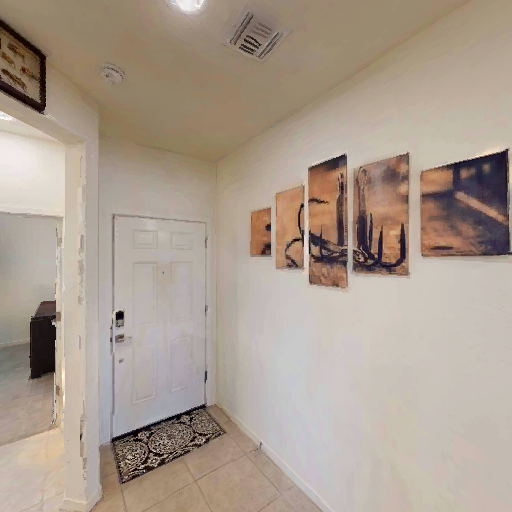}} &
\correctimg[width=\linewidth,keepaspectratio]{{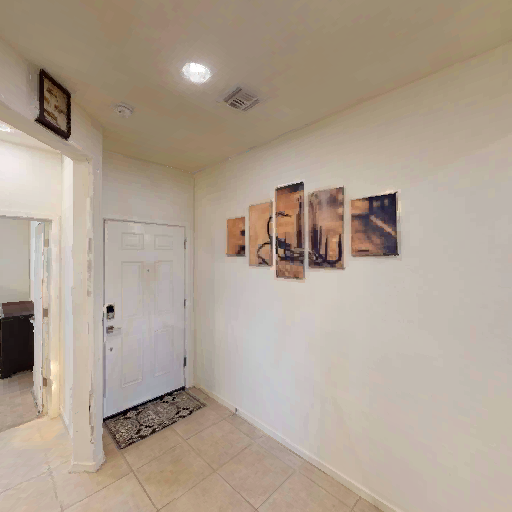}} &
\correctimg[width=\linewidth,keepaspectratio]{{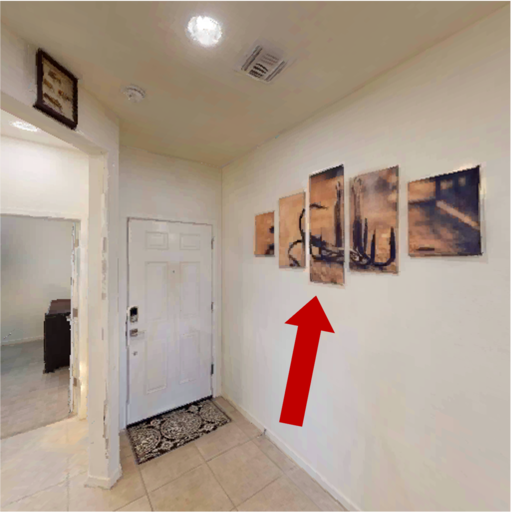}} \\
% \midrule
\bottomrule
\end{tabularx}
\vspace{-\baselineskip}
\caption{\textbf{More qualitative results on \Benchmark{}-ProcTHOR (top four rows) and \Benchmark{}-HM3D (bottom four rows).} \targetobj{Blue} and \receptacle{gray} mark the object of interest and  surrounding cue objects, respectively. \correctmark denotes correct answers (LLM-Match $=5$), \wrongmark incorrect ones (LLM-Match $\leq 2$). 
% The red arrow in the last column highlights the region of interest.
}
\label{fig:quali_additional}
\vspace{-1.5\baselineskip}
\end{figure*}

We provide more qualitative comparison results of Visually-Grounded Active View Selection in Figure~\ref{fig:quali_additional}. 

\clearpage
\newpage
\onecolumn
\subsection{Qualitative Examples of AVS Framework Reasoning}
\label{sec:supp_qualitative_examples}
In the following, we present qualitative examples illustrating how our AVS framework reasons from the query view to predict the desired action parameters.

\hfill
% \begin{center}
% \textbf{Qualitative examples are\\in the following pages.}
% \end{center}

% \clearpage
% \newpage

%%%%%%%%%%%%%%%%%%%%%%%%%%%%%%%%
\newcommand{\AllComparisonImages}{\ImagePair{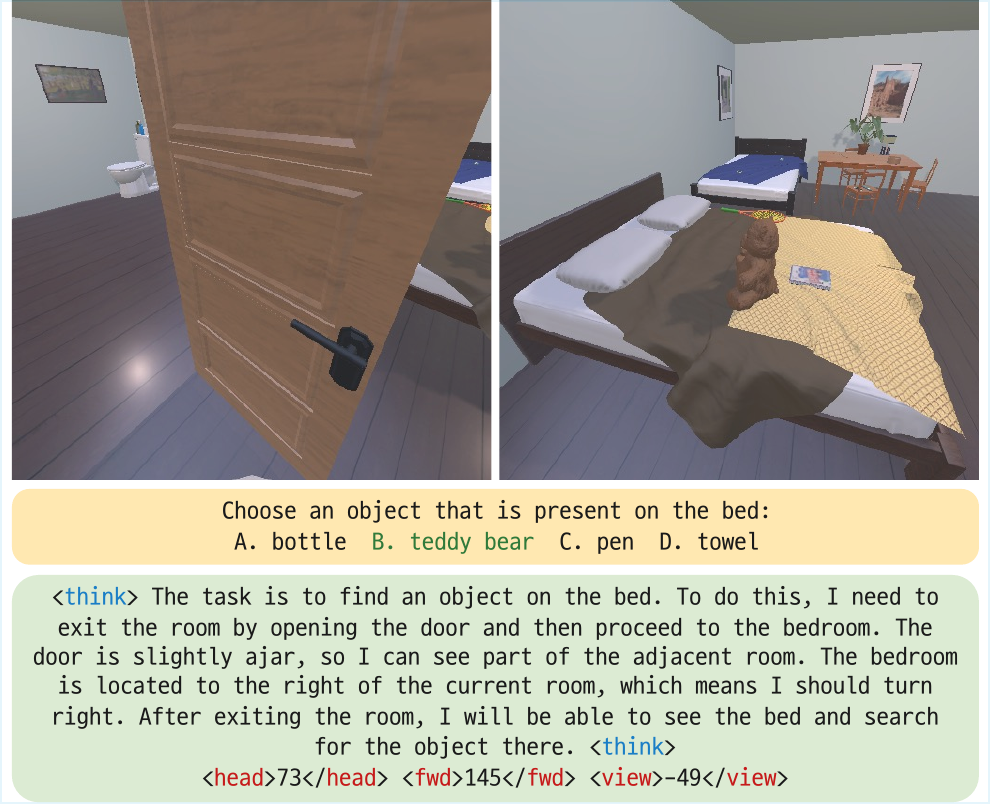}{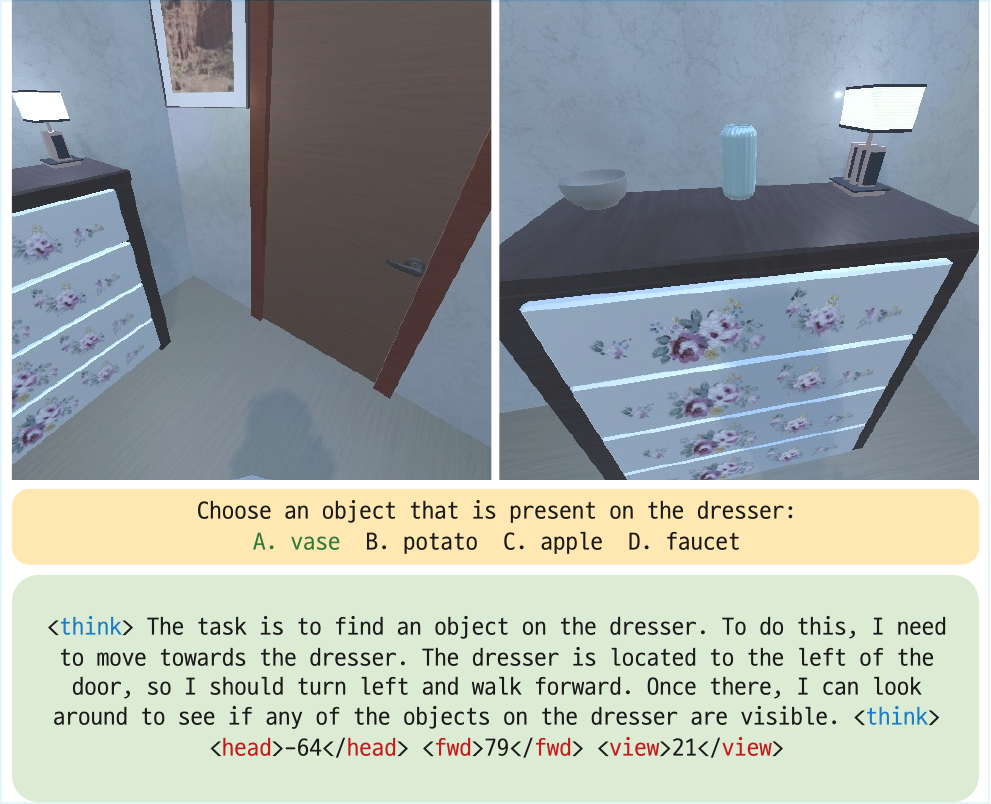}
\ImagePairLast{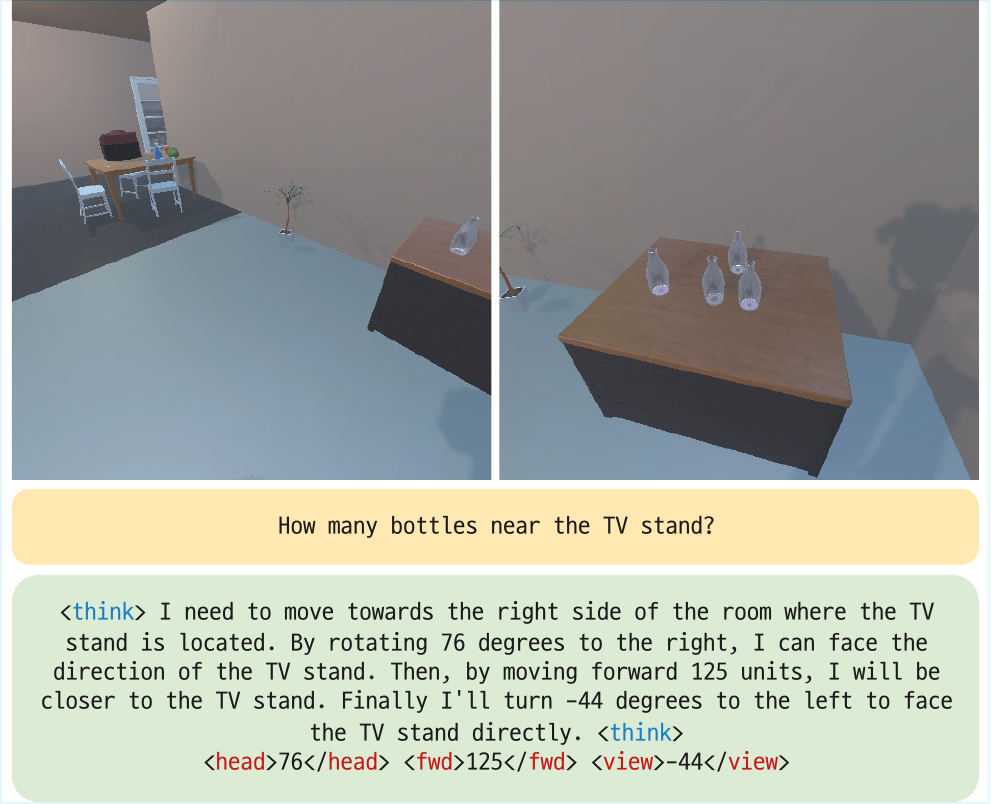}{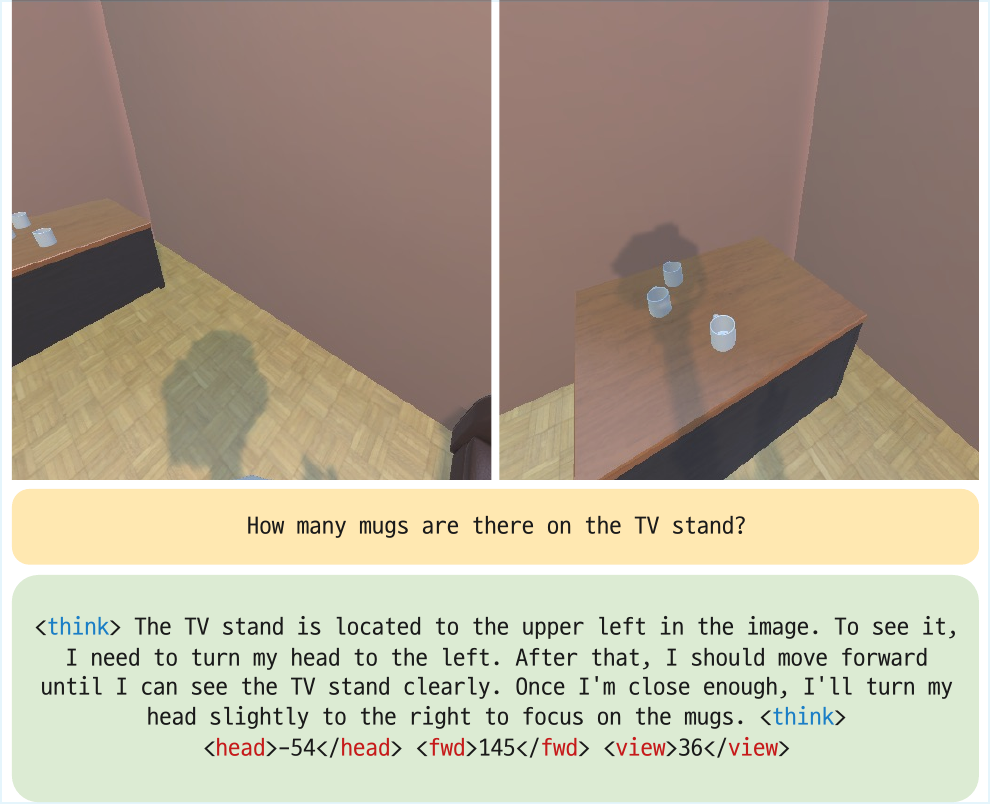}
\ImagePair{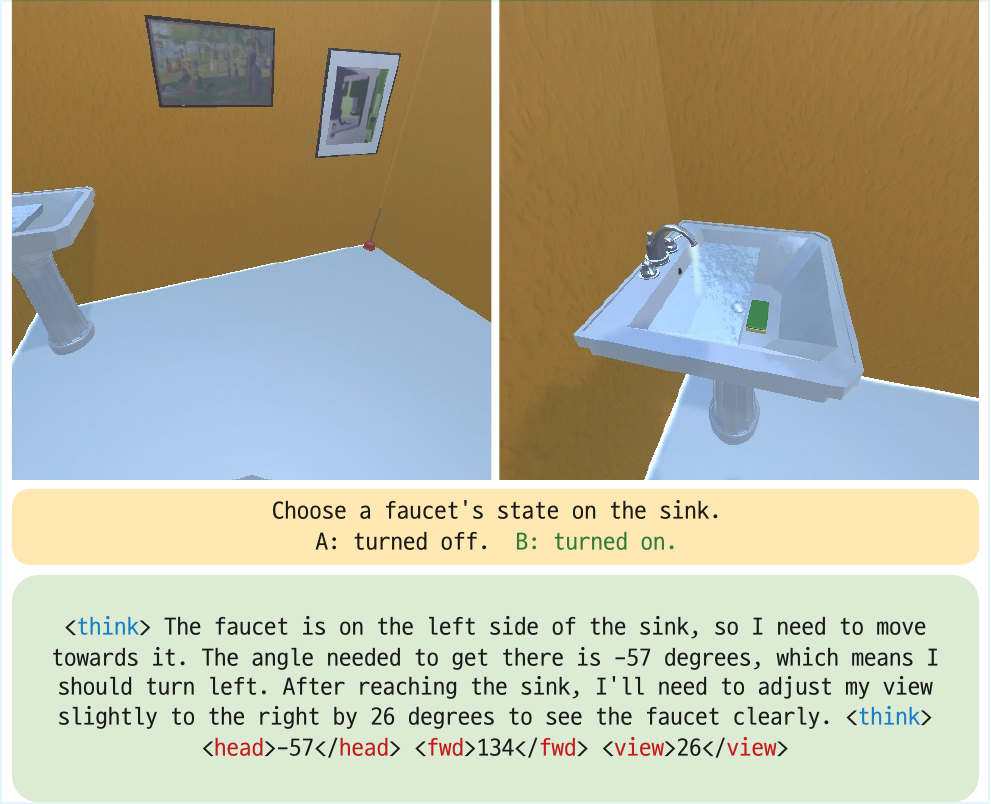}{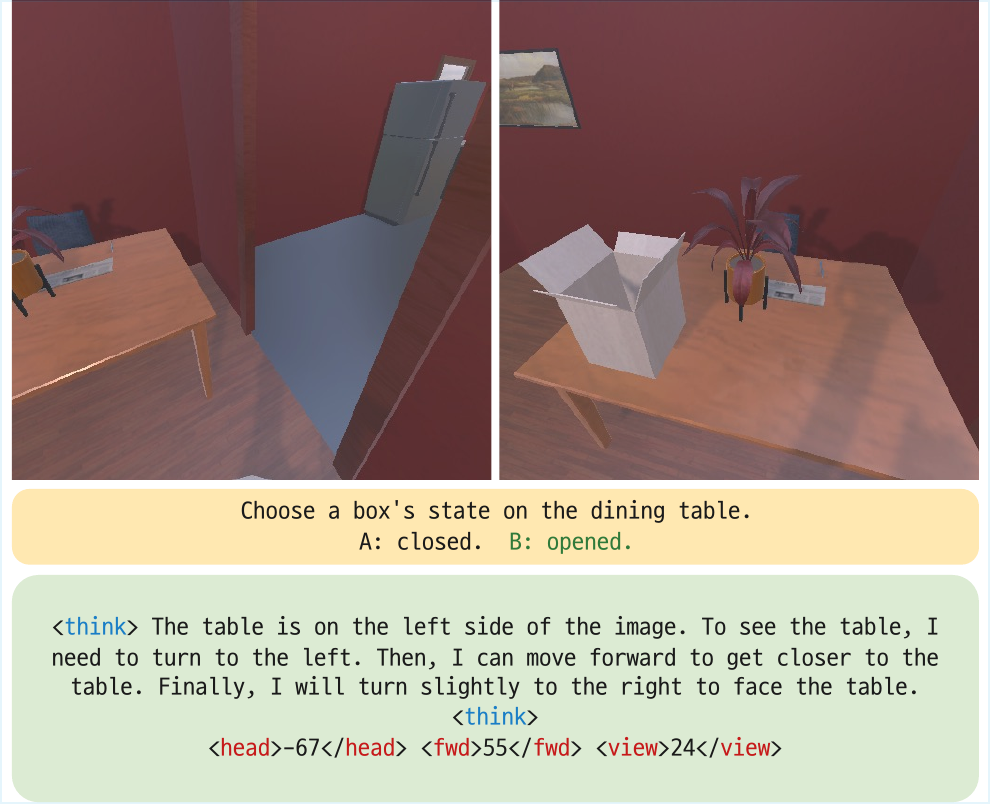}
\ImagePair{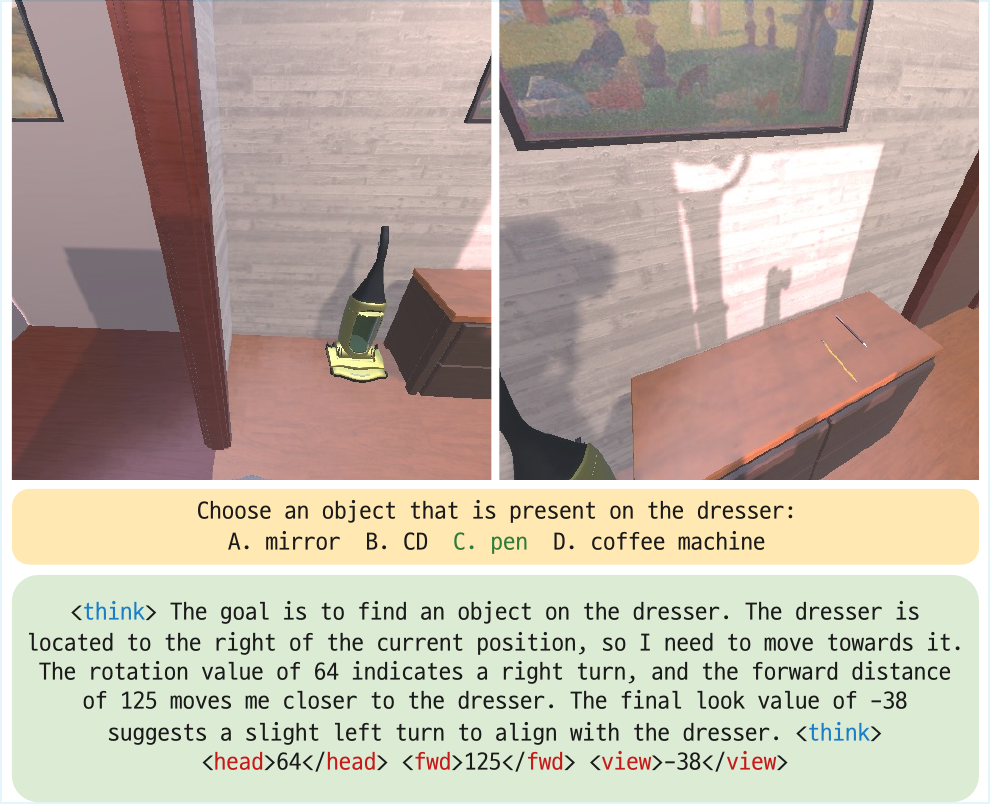}{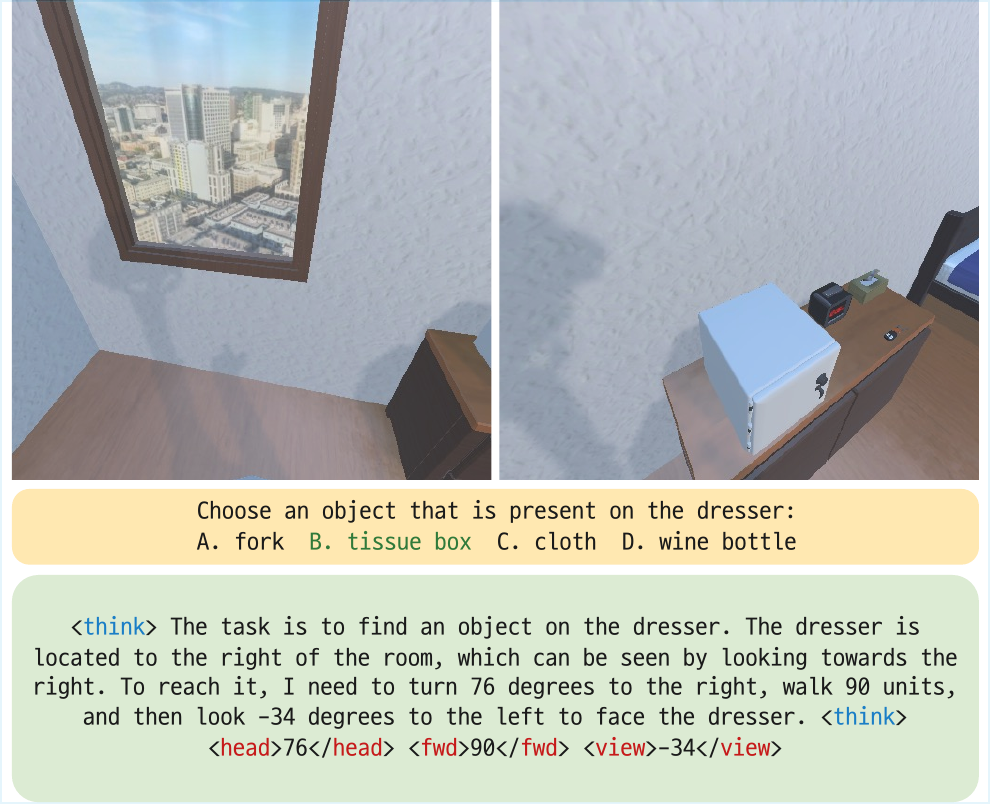}
\ImagePairLast{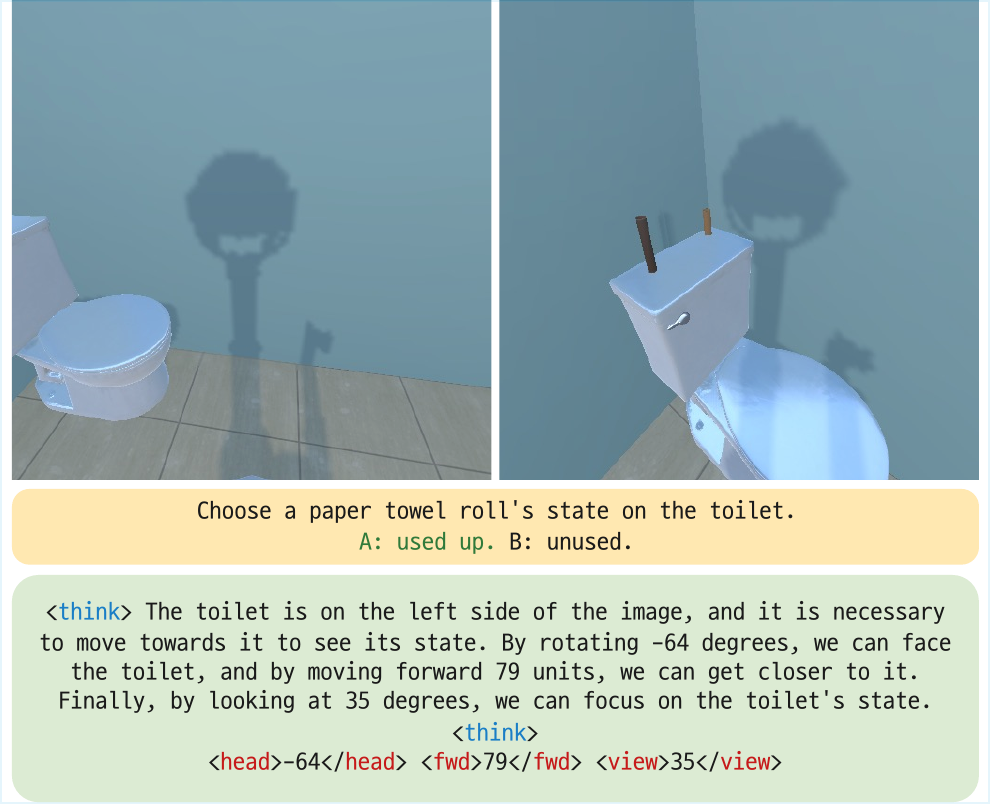}{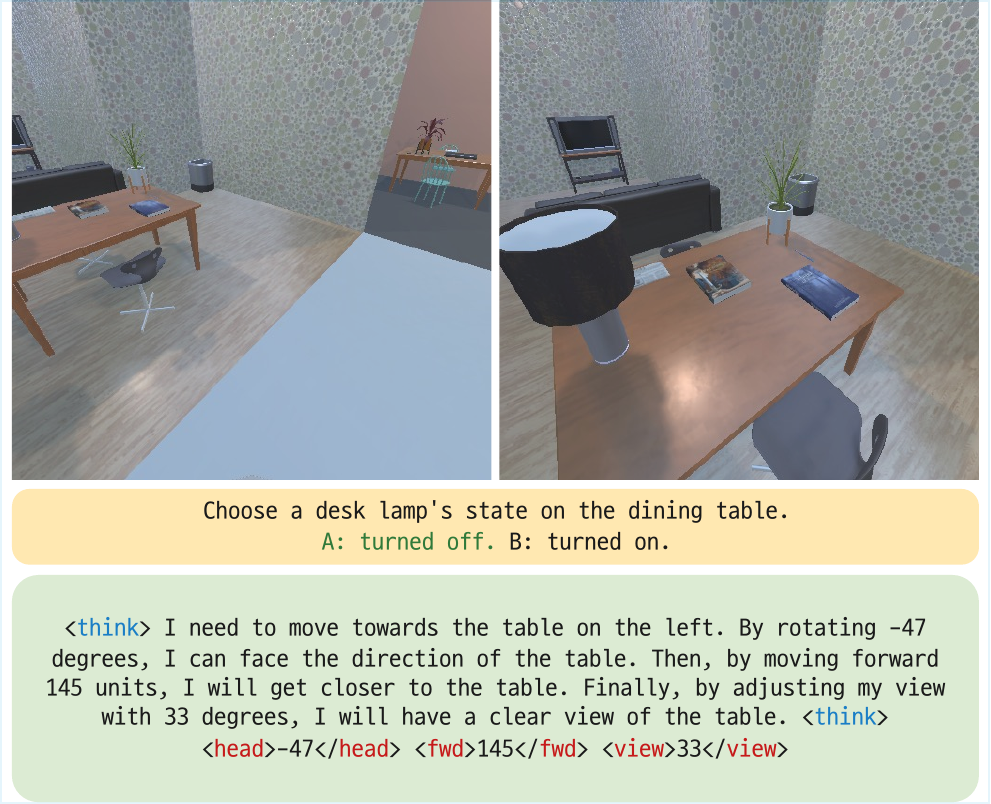}

}
\graphicspath{{figures/thinking_images/}}

\setlength{\tabcolsep}{0.0em}
\def\arraystretch{0.0}

{
\scriptsize
\begin{longtable}{Z  Z | Z  Z}
\caption{\textbf{Qualitative example illustrating the model’s reasoning process and executed actions in \Benchmark{}-ProcTHOR.} Given the input view (left), the model identifies partially observable visual cues and infers the appropriate action needed to complete the missing context. The predicted action sequence (\textless head\textgreater, \textless fwd\textgreater, \textless view\textgreater) is then executed, producing the output view (right), from which the verifier successfully answers.}\label{fig:procthor_thinking} \\
Input View & Output View & Input View & Output View \\
\midrule
\endfirsthead
Input View & Output View & Input View & Output View \\
\midrule
\endhead

\bottomrule
\endfoot

\ImagePair{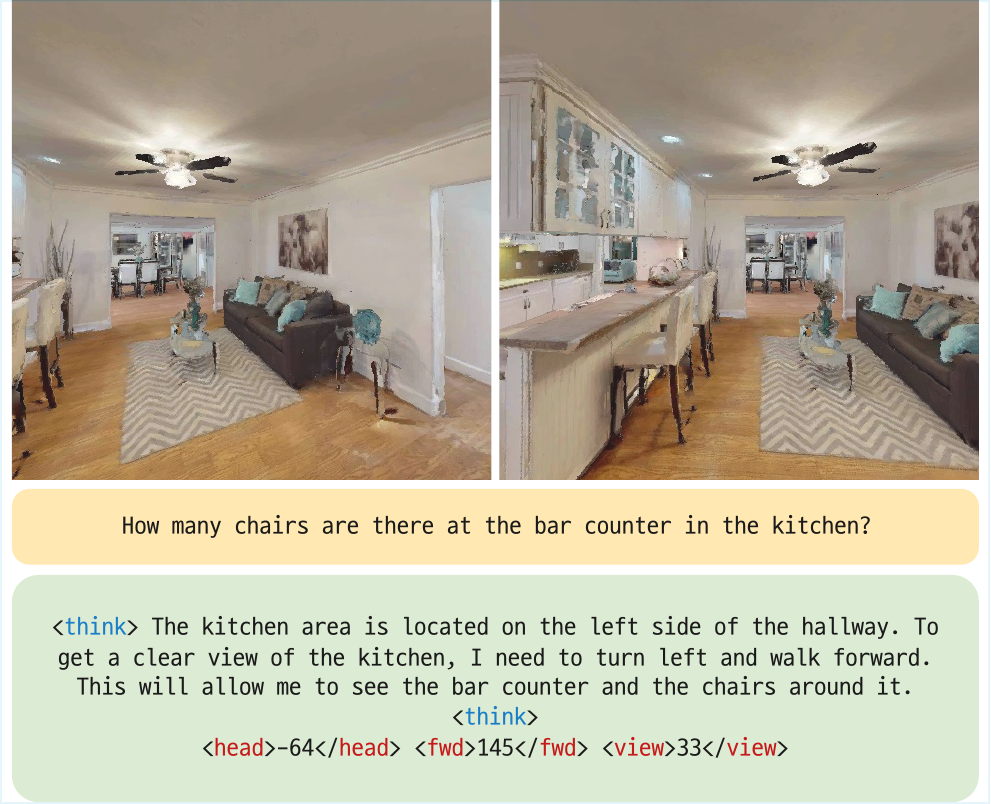}{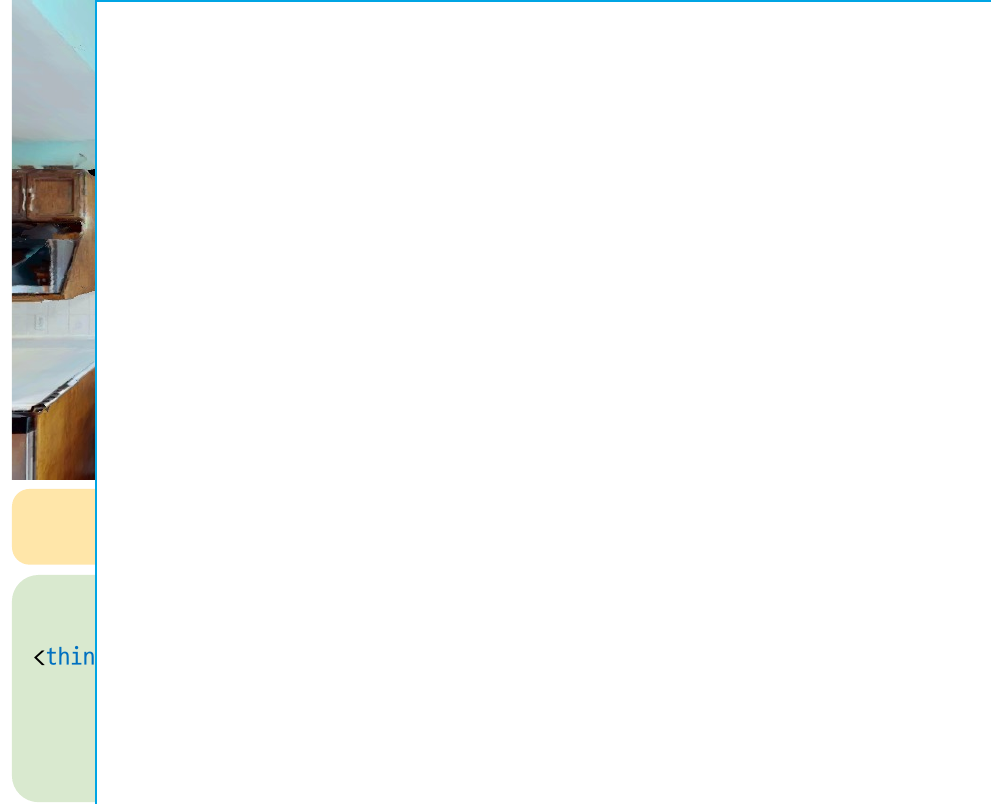}
\ImagePairLast{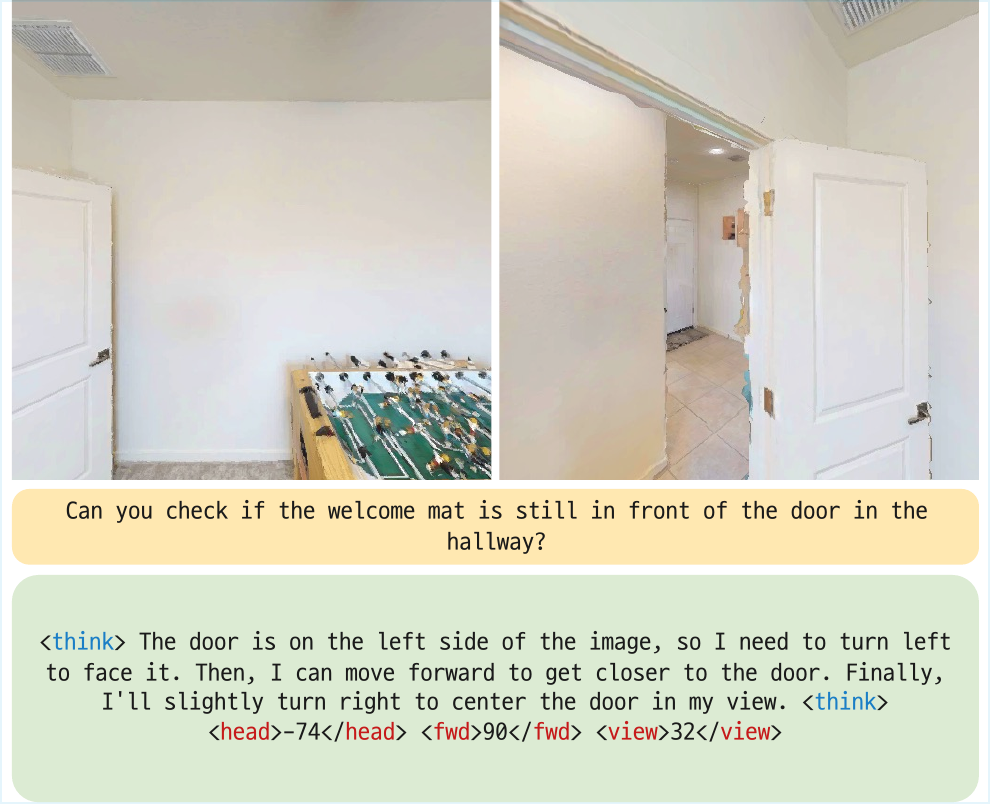}{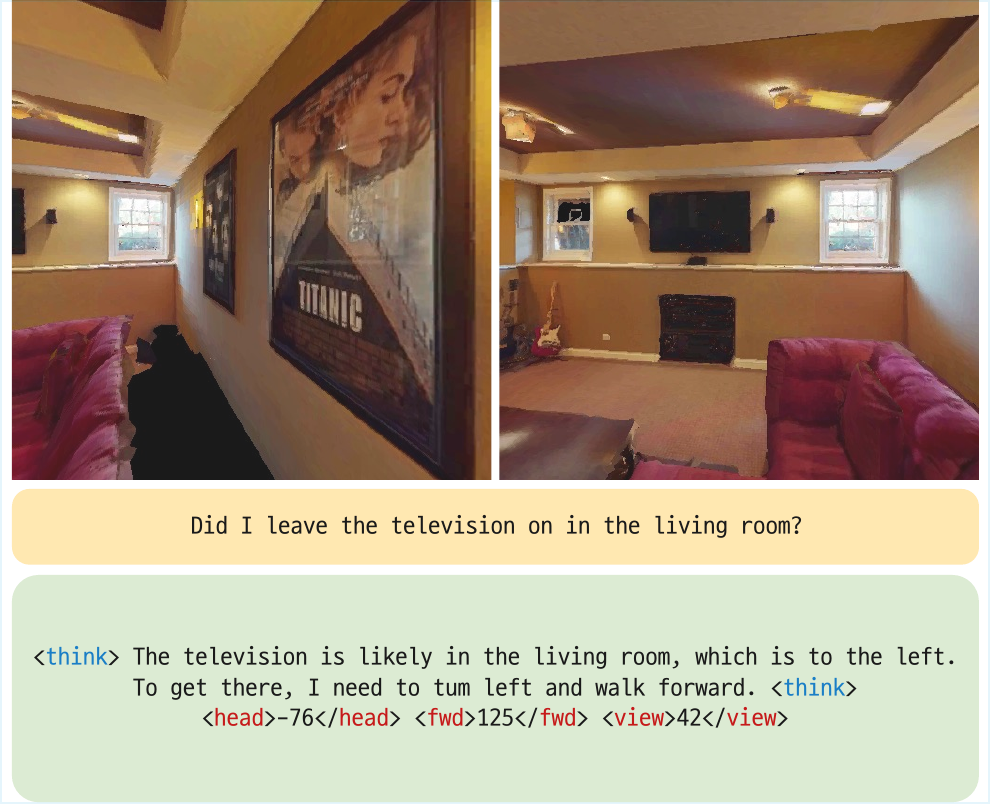}

\end{longtable}
}

%%%%%%%%%%%%%%%%%%%%%%%%%%%%%%%%
\newcommand{\AllComparisonImages}{\ImagePair{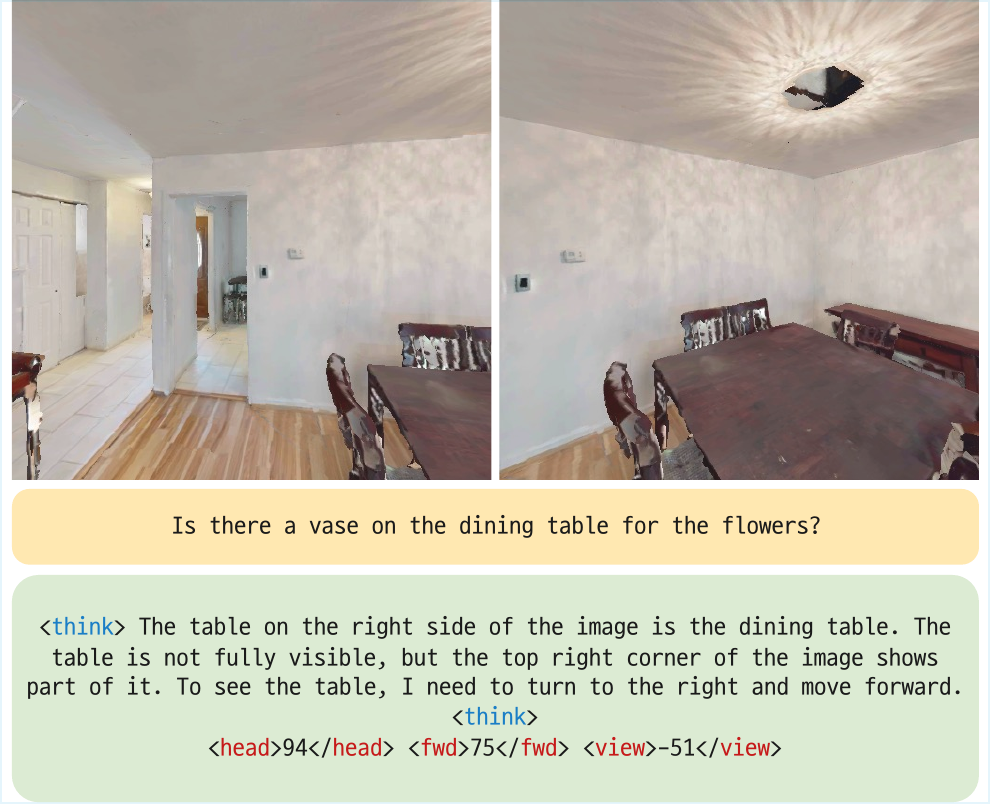}{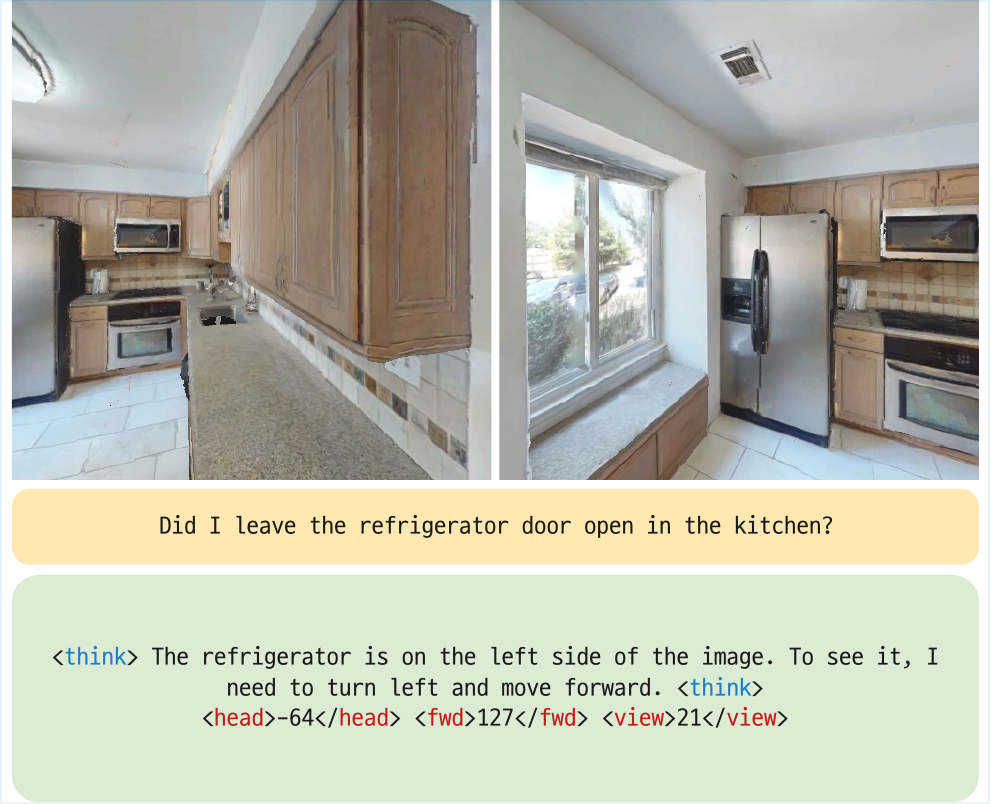}
\ImagePair{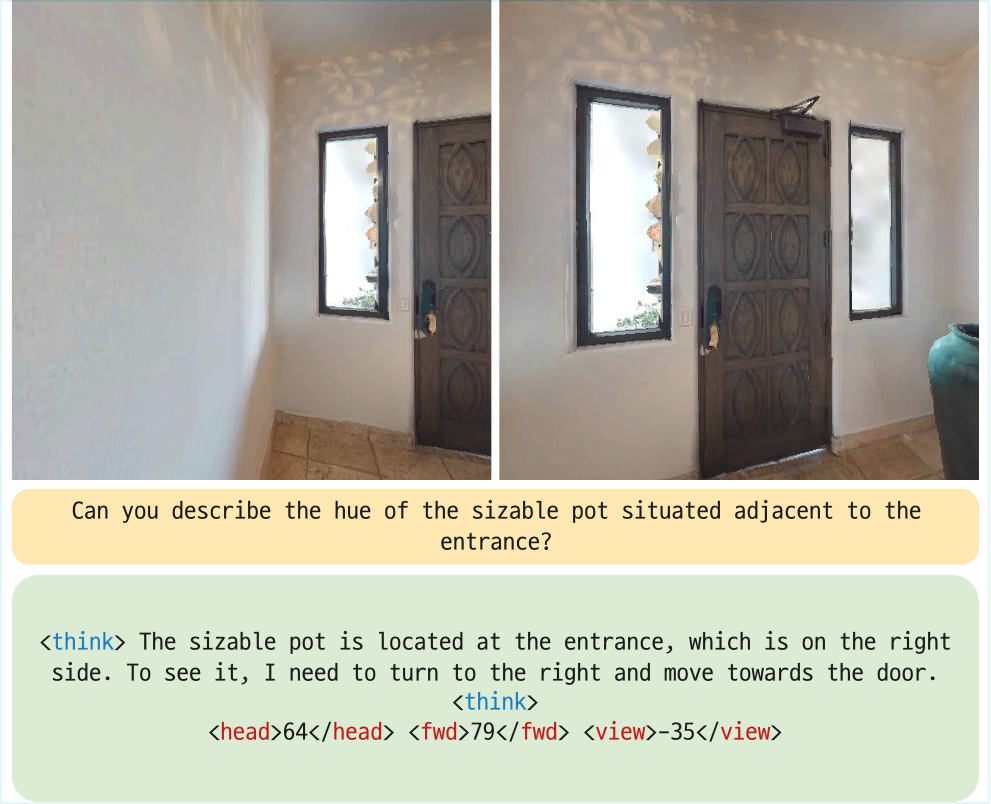}{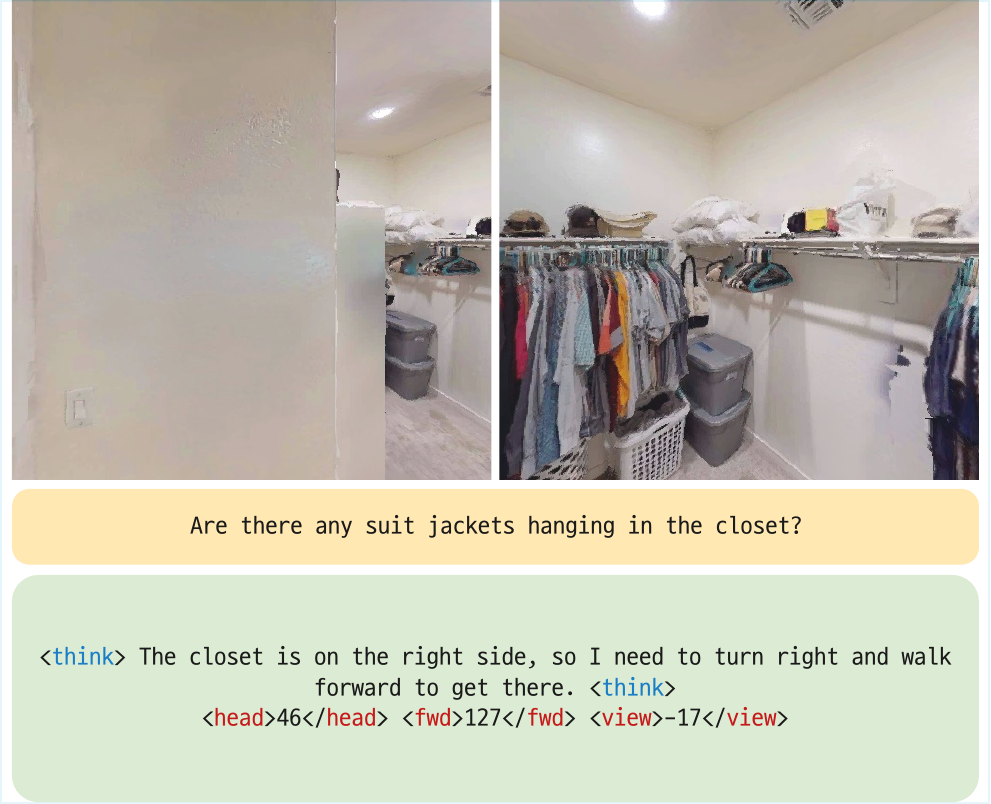}
\ImagePairLast{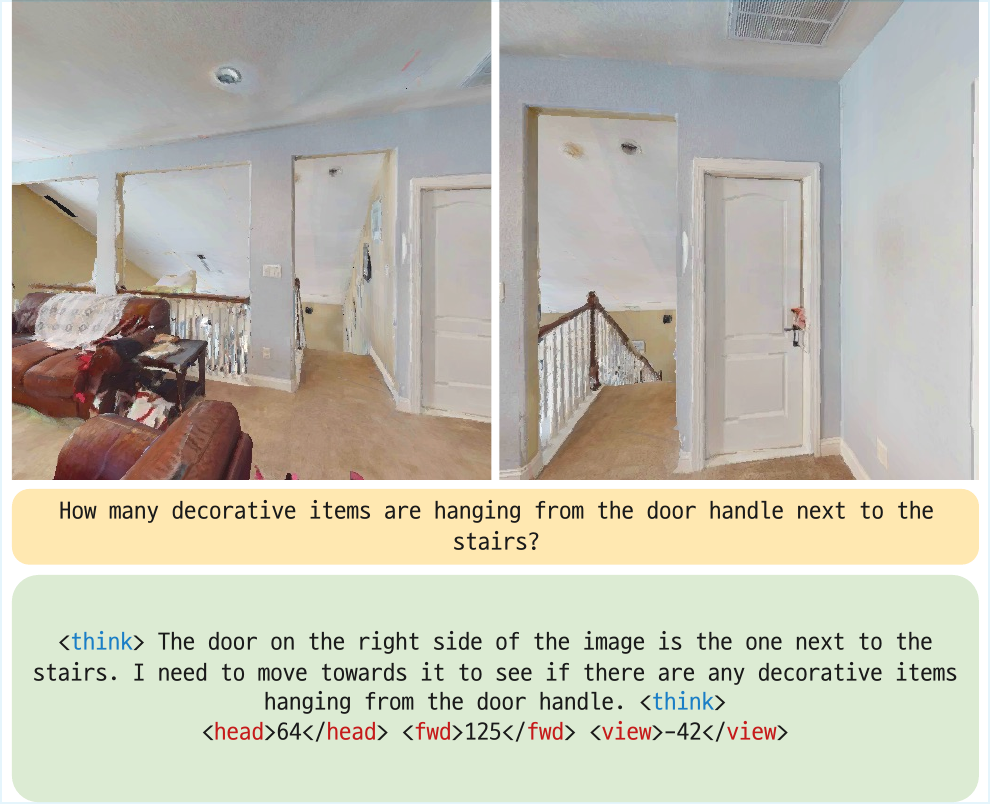}{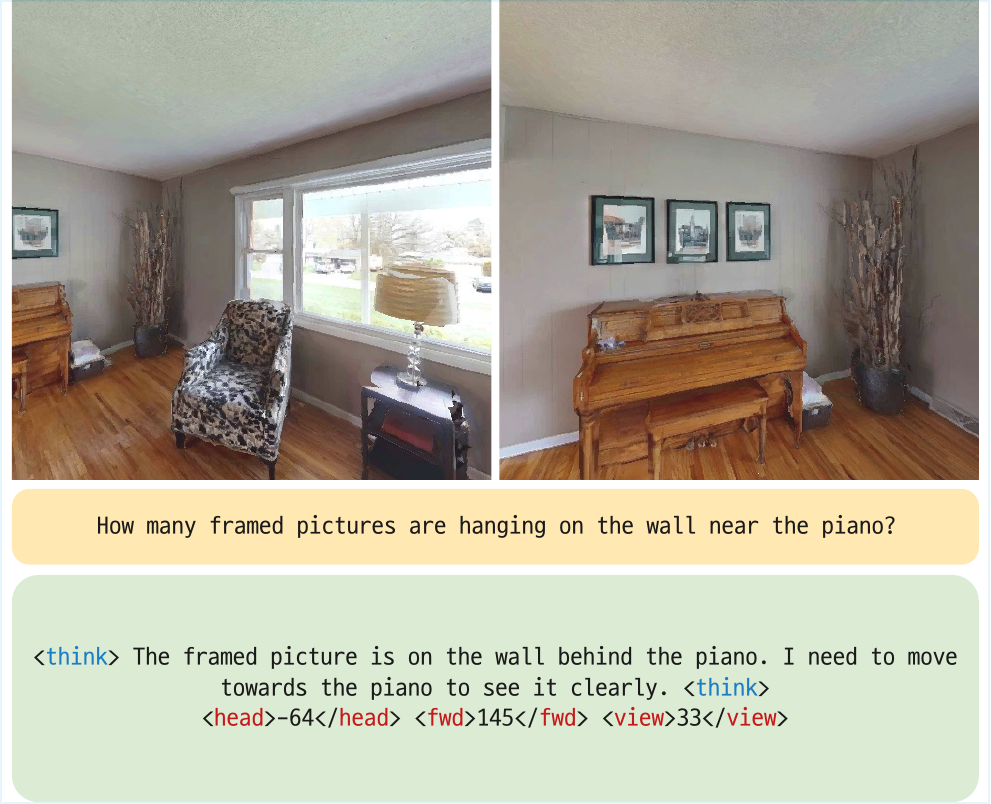}
\ImagePair{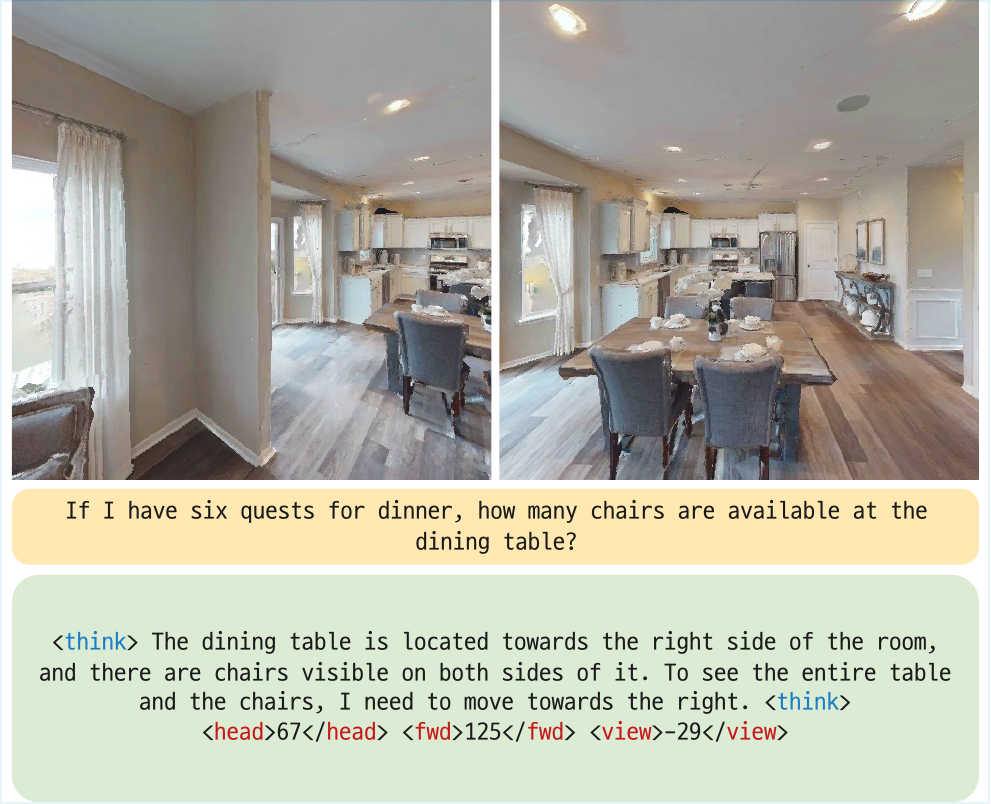}{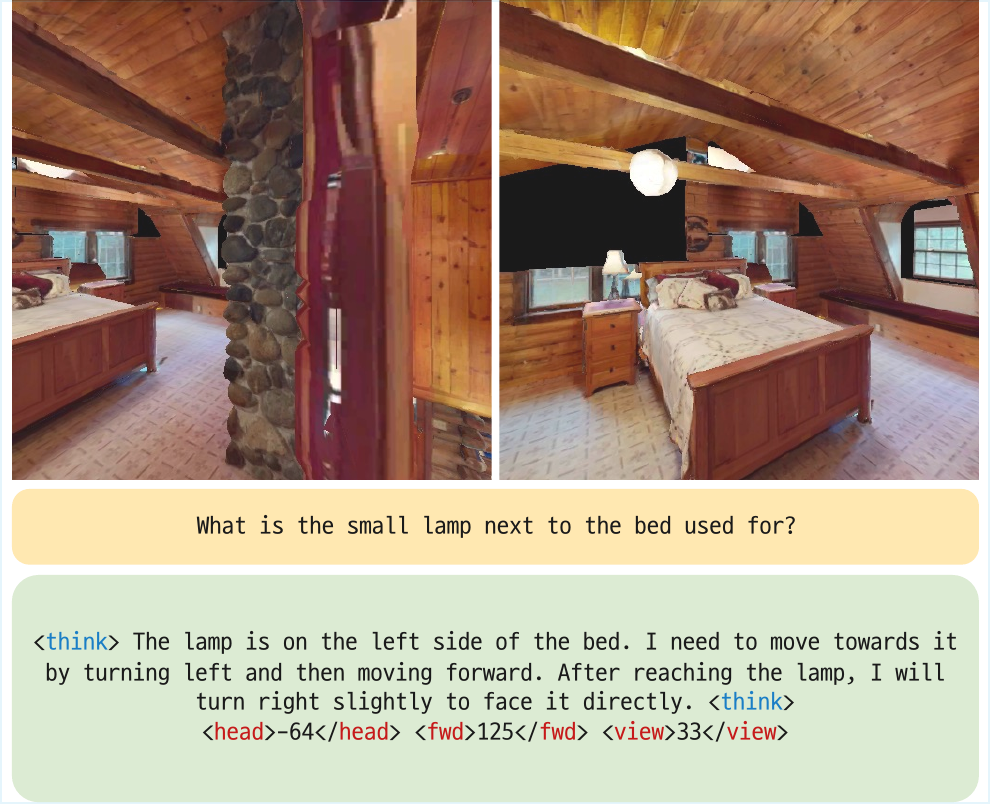}
\ImagePairLast{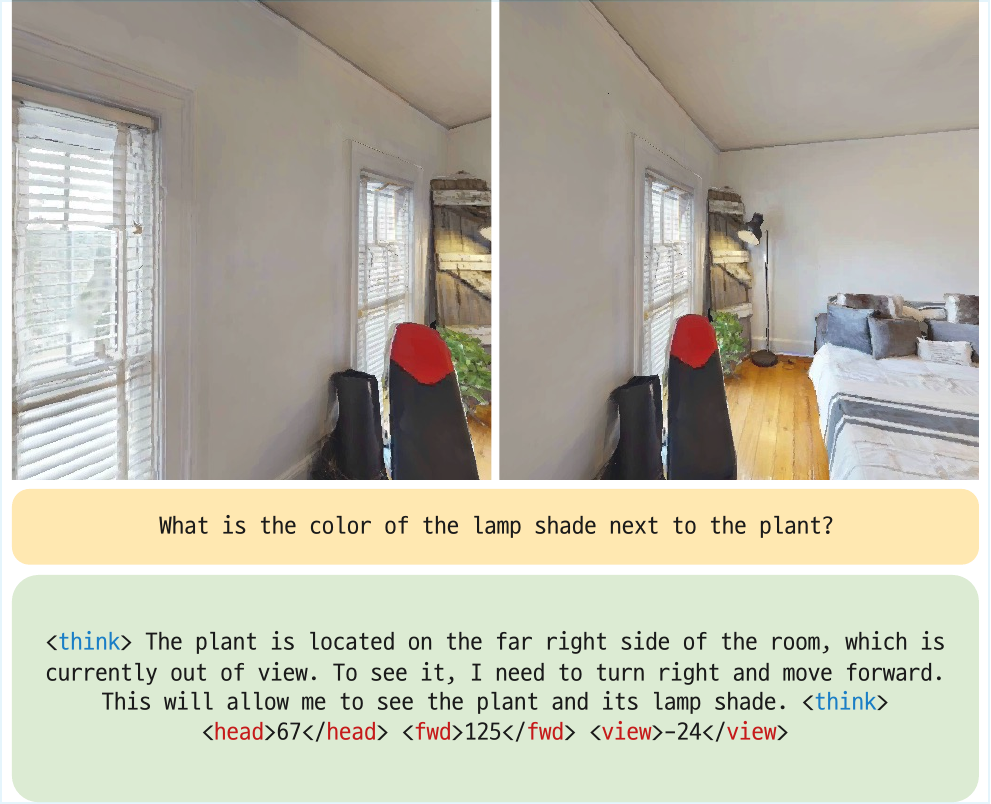}{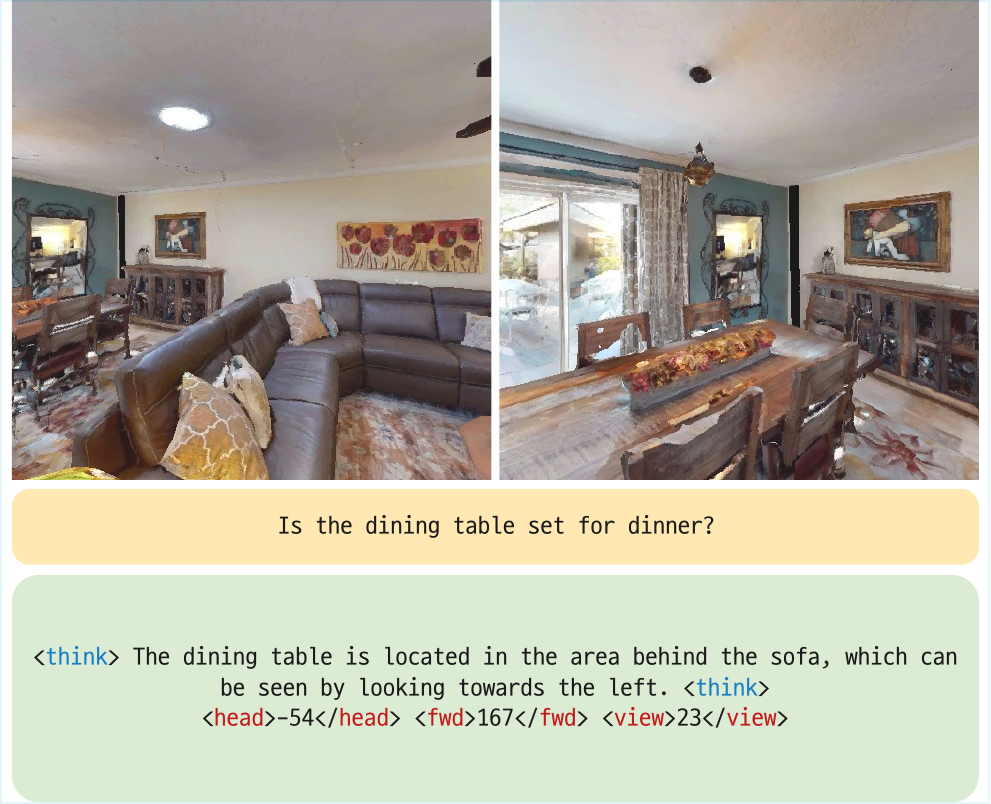}

}
\graphicspath{{figures/thinking_images/}}

\setlength{\tabcolsep}{0.0em}
\def\arraystretch{0.0}

{
\scriptsize
\begin{longtable}{Z  Z | Z  Z}
\caption{\textbf{Qualitative example illustrating the model’s reasoning process and executed actions in \Benchmark{}-HM3D.} Given the input view (left), the model identifies partially observable visual cues and infers the appropriate action needed to complete the missing context. The predicted action sequence (\textless head\textgreater, \textless fwd\textgreater, \textless view\textgreater) is then executed, producing the output view (right), from which the verifier successfully answers.}\label{fig:hm3d_thinking} \\
Input View & Output View & Input View & Output View \\
\midrule
\endfirsthead
Input View & Output View & Input View & Output View \\
\midrule
\endhead

\bottomrule
\endfoot

\ImagePair{figures/thinking_images/fine-eqa_1.pdf}{figures/thinking_images/fine-eqa_2.pdf}
\ImagePairLast{figures/thinking_images/fine-eqa_3.pdf}{figures/thinking_images/fine-eqa_4.pdf}

\end{longtable}
}

\hfill

%%%%%%%%%%%%%%%%%%%%%%%%%%%%%%%%
\newcommand{\AllComparisonImages}{\ImagePair{figures/thinking_images/fine-eqa_1.pdf}{figures/thinking_images/fine-eqa_2.pdf}
\ImagePairLast{figures/thinking_images/fine-eqa_3.pdf}{figures/thinking_images/fine-eqa_4.pdf}

}
\graphicspath{{figures/thinking_images/}}

\setlength{\tabcolsep}{0.0em}
\def\arraystretch{0.0}

{
\scriptsize
\begin{longtable}{Z  Z | Z  Z}
\caption{\textbf{Qualitative example illustrating the model’s reasoning process and executed actions in Fine-EQA~\cite{Liu:2024EXPRESSBench}.} 
Fine-EQA first performs its own exploration and provides an observation (left). Based on it, our model identifies visual cues that are insufficient for answering and reasons that an additional action is required to complete the missing context. The model then predicts an action sequence (\textless head\textgreater, \textless fwd\textgreater, \textless view\textgreater), executes it, and obtains the updated view (right), from which the verifier successfully answers the question.}
\label{fig:fine-eqa_thinking} \\
Final View of Fine-EQA~\cite{Liu:2024EXPRESSBench} & Refined View by Ours & Final View of Fine-EQA~\cite{Liu:2024EXPRESSBench} & Refined View by Ours \\
\midrule
\endfirsthead
Input View & Output View & Input View & Output View \\
\midrule
\endhead

\bottomrule
\endfoot

\ImagePair{figures/thinking_images/fine-eqa_1.pdf}{figures/thinking_images/fine-eqa_2.pdf}
\ImagePairLast{figures/thinking_images/fine-eqa_3.pdf}{figures/thinking_images/fine-eqa_4.pdf}

\end{longtable}
}

\clearpage
\newpage
\twocolumn
\else
\fi

\end{document}